\documentclass[twocolumn,twoside,letter]{IEEEtran}
\usepackage{psfrag}
\usepackage{amsmath}
\usepackage{amssymb}
\usepackage{mathabx}
\usepackage{graphicx}
\usepackage{graphbox}
\usepackage{subfigure}
\usepackage{verbatim}
\usepackage{latexsym}
\usepackage[dvipsnames]{xcolor}
\usepackage{cite}
\usepackage{stmaryrd}
\usepackage{multirow}
\usepackage{subequation}
\usepackage{textcomp}
\usepackage[ruled]{algorithm2e}

\usepackage[normalem]{ulem}
\usepackage{tikz}
\usetikzlibrary{fit,calc}

\usepackage{tikz}
\usepackage{textcomp}
\usepackage{hyperref}
\usepackage{lipsum}

\newcommand\copyrighttext{%
	\footnotesize \textcopyright 2024 IEEE. Personal use is permitted, but republication/redistribution requires IEEE permission. See https://www.ieee.org/publications/rights/index.html for more information.
	This article has been accepted for publication in IEEE Transactions on Robotics. Citation information: \href{https://ieeexplore.ieee.org/document/10506627}{DOI 10.1109/TRO.2024.3392077.}}
\newcommand\copyrightnotice{%
	\begin{tikzpicture}[remember picture,overlay]
		\node[anchor=south,yshift=11pt] at (current page.south) {\fbox{\parbox{\dimexpr\textwidth-\fboxsep-\fboxrule\relax}{\copyrighttext}}};
	\end{tikzpicture}%
}

\colorlet{pink}{red!40}

\makeatletter
\newcounter{phase}
\newlength{\phaserulewidth}
\newcommand{\setphaserulewidth}{\setlength{\phaserulewidth}}

\makeatother
\setphaserulewidth{.7pt}

\newbox\tempbox

\DeclareMathAlphabet{\mathpzc}{OT1}{pzc}{m}{it}

\newcommand{\bs}[1]{\ensuremath{{\boldsymbol{#1}}}}

\def\s{\mathop{\rm s}\nolimits}
\def\c{\mathop{\rm c}\nolimits}

\newtheorem{lem}{Lemma}

\newtheorem{prop}{Proposition}

\newcommand{\qed}{\hfill \mbox{~\rule[-1pt]{6pt}{6pt}}}
\def\atan2{\mathrm{atan2}}
\newenvironment{proof-sketch}{\noindent\hspace*{1em}{\it Sketch of Proof:}}{\qed}

\begin{document}

\bibliographystyle{IEEEtran}

\title{Complete and Near-Optimal Robotic Crack Coverage and Filling in Civil Infrastructure~\thanks{The work was supported in part by the US NSF under award IIS-1426828 (J. Yi). An earlier version of this paper was presented in part at the 2019 IEEE International Conference on Robotics and Automation, Montreal, Canada, May 20–24, 2019 [10.1109/ICRA.2019.8794407]. ({\em Corresponding author: Kaiyan Yu and Jingang Yi}).}}

\markboth{IEEE TRANSACTIONS ON ROBOTICS,~Vol.~XX,~No.~YY, 2024}{Veeraraghavan~\lowercase{\textit{et al.}}: Complete and Near-Optimal Robotic Crack Coverage and Filling in Civil Infrastructure}

\author{Vishnu Veeraraghavan,\thanks{The first two authors equally contributed to the work.} Kyle Hunte,\thanks{V. Veeraraghavan and K. Yu are with the Department of Mechanical Engineering, Binghamton University, Binghamton, NY 13902 USA (email: {vveerar1@binghamton.edu}; {kyu@binghamton.edu}).} Jingang Yi,\thanks{K. Hunte and J. Yi are with the Department of Mechanical and Aerospace Engineering, Rutgers University, Piscataway, NJ 08854 USA (email: {kdh95@scarletmail.rutgers.edu}; {jgyi@rutgers.edu}).} and Kaiyan Yu~\thanks{Videos of the experiments and source codes are available at  \href{https://github.com/Binghamton-ACSR-Lab/Crack-Filling-Robot}{https://github.com/Binghamton-ACSR-Lab/Crack-Filling-Robot.}}\\}

\maketitle

\copyrightnotice
\IEEEaftertitletext{\vspace{-1\baselineskip}}
\begin{abstract}
We present a simultaneous sensor-based inspection and footprint coverage (SIFC) planning and control design with applications to autonomous robotic crack mapping and filling. The main challenge of the SIFC problem lies in the coupling of complete sensing (for mapping) and robotic footprint (for filling) coverage tasks. Initially, we assume known target information (e.g., cracks) and employ classic cell decomposition methods to achieve complete sensing coverage of the workspace and complete robotic footprint coverage using the least-cost route. Subsequently, we generalize the algorithm to handle unknown target information, allowing the robot to scan and incrementally construct the target map online while conducting robotic footprint coverage. The online polynomial-time SIFC planning algorithm minimizes the total robot traveling distance, guarantees complete sensing coverage of the entire workspace, and achieves near-optimal robotic footprint coverage, as demonstrated through experiments. For the demonstrated application, we design coordinated nozzle motion control with the planned robot trajectory to efficiently fill all cracks within the robot's footprint. Experimental results illustrate the algorithm's design, performance, and comparisons. The SIFC algorithm offers a high-efficiency motion planning solution for various robotic applications requiring simultaneous sensing and actuation coverage.

\end{abstract}
  

\vspace{-0mm}
\section{Introduction}

\IEEEPARstart{S}{urface} cracks commonly exist in civil infrastructure, such as road and bridge deck surfaces, parking lots, airport runways, etc. To prevent crack growth and mitigate further deterioration, it is necessary to fill these cracks with appropriate materials in the early stages of crack appearance. Repairing the abundant existence of cracks in civil infrastructure by human workers is labor-intensive, time-consuming, and expensive. Robotics and automation technologies provide a promising tool to enable cost-effective, safe, and high-efficiency civil infrastructure maintenance. Mobile robot- or vehicle-based inspection systems were used for crack detection and maintenance on highways (e.g.,~\cite{zhang2023automated,eskandari2020robotic}). However, these systems were not fully automated, and human inspectors were still involved in operations. The autonomous robots developed in~\cite{LaTMech2013,LaCASE2013,GucunskiIJIRA2017} mainly focused on nondestructive bridge deck inspection, and although crack mapping and repair were discussed, they were not the primary focus.
Crack mapping and filling can be viewed and generalized as a simultaneous robotic sensor-based inspection and footprint coverage (SIFC) problem. The onboard sensing system (e.g., camera) needs to detect all unknown targets (i.e., cracks) in the workspace, and at the same time, the robot must physically cover all the detected targets within its footprint to conduct the repair action. The complexity of the SIFC problem lies in simultaneously achieving the above-mentioned two complete coverage tasks of unknown targets using one robotic platform. The onboard target detection sensing range and the robot footprint have different sizes and geometric shapes. Additionally, the robot footprint coverage task (e.g., filling cracks) might involve motion dynamics and control constraints, which are different from the coupled (passive) sensing coverage task.
Although inspired by autonomous robotic crack mapping and filling, the SIFC is indeed a fundamental robot motion planning problem in other applications, such as sensing and cleaning dirty surfaces, finding and collecting mines, etc.

The SIFC planning is related to the covering salesman problem, a variant of the traveling salesman problem where an agent must travel the shortest distance to visit all specified neighborhoods in each city. However, unlike traditional scenarios where the agent has prior knowledge of each city's location, in the SIFC problem, the agent must inspect every point in the environment to detect unknown targets. Other related problems include the art gallery problem and the watchman tour problem, but these do not involve the footprint coverage of targets. It's worth noting that all these problems are NP-hard, making obtaining optimal solutions feasible only for very limited problem domains~\cite{Bormann2018}.
Robotic exploration techniques can be used to detect unknown targets~\cite{kan2020online,wu2019energy}. However, in time-critical applications, many exploration paradigms suffer from inefficiencies due to backtracking, where the robot may revisit the same location more than once~\cite{Palacios2017}.

Coverage path planning explores environments by determining an optimal path that covers all points of interest while avoiding  obstacles~\cite{galceran2013survey,tan2021comprehensive}. These methods are classified into offline or online methods based on whether prior environmental information is known. In offline approaches, known information about the environment is used to produce the shortest or fastest path~\cite{mannadiar2010optimal}, while in online approaches, sensor information is used to plan coverage motion point-by-point. Many online strategies used heuristics to navigate to the nearest cell~\cite{rekleitis2008efficient}, the cell with the highest potential~\cite{Gupta2018,song2020care}, or the cell with the lowest cost~\cite{elgibreen2019dynamic}. However, as they did not optimize the coverage path from a global perspective, these online planning approaches might fall into local extrema. To prevent the robot from getting stuck in local extrema, online approaches employed heuristic strategies such as backtracking procedures~\cite{Bustacara2005} or potential surfaces~\cite{Gupta2018} to passively identify the next waypoint. These heuristic strategies prevent the robot from getting stuck in local extrema; however, they may cause the robot to repeatedly visit the covered area. Alternatively, coverage path planning approaches in~\cite{Zhang2021ICRA,mannadiar2010optimal} avoided local extrema by using spanning trees, guiding the robot to circumvent the virtual tree. However, these algorithms did not account for partially occupied cells or special tree nodes, resulting in incomplete or repeated coverage~\cite{song2020care}.

Morse-based cellular decomposition (MCD)~\cite{acar2002morse} and generalized Boustrophedon decomposition~\cite{choset1998coverage} guarantee complete coverage of an unknown environment. MCD methods ensure encountering all critical points of the decomposition online~\cite{acar2002sensor,acar2006sensor,Acar}. These algorithms perform online decomposition such that the areas are covered completely by back-and-forth motions. However, no optimality is claimed for the planned paths. The work in~\cite{mannadiar2010optimal} proposed an optimal coverage of a known environment. Grid-based approaches for planning a complete coverage path were also studied~\cite{gabriely2002spiral}. However, these approaches restrict space and robot motion to grids rather than arbitrary points in the workspace. Readers can refer to~\cite{galceran2013survey,choset2001coverage} for extensive surveys on coverage planning strategies.

Online re-planning of the robot's path is necessary when new information is obtained by onboard sensors~\cite{LaValle2006}. A graph-based Simplex method was presented to solve the re-planning problem~\cite{ersson2001path}. The anytime dynamic A* algorithm is a generic graph-based re-planning scheme that generates bounded sub-optimal solutions when the map changes~\cite{likhachev2005anytime}. Navigation and coverage planning for autonomous underwater vehicles is a closely related problem (e.g.,~\cite{paull2013sensor}). However, existing work focuses solely on the exploration task of unknown environments and does not consider simultaneously performing other tasks such as robot footprint coverage. Another closely related piece of work is efficient autonomous robotic cleaning or vacuuming~\cite{hess2013poisson,hess2014probabilistic}, where a learned dirt map is used for planning the robot to clean a set of cells. This differs from the SIFC problem because the initial learning process may be costly or infeasible for applications such as the online construction of a crack map for repair. The probabilistic planner in~\cite{rottmann2021probabilistic} utilized Monte Carlo localization for complete coverage path planning but suffered from accumulating localization errors. While such probabilistic planners can perform online planning in dynamic environments, they often entail higher computational complexity, rely on specific modeling assumptions (e.g., distributions of obstacles and uncertainty), and provide solutions that are only probabilistically optimal.

In this paper, we present a set of motion planning algorithms for the SIFC problem with the application of robotic crack-filling in civil infrastructure. Instead of directly solving the SIFC problem, we first discuss the motion planning of a mobile robot to physically cover the targets (i.e., cracks) for a given known target map; that is, no complete sensing coverage is considered. We decouple the two coverage planning tasks using classic cell decomposition methods to achieve complete sensing and robotic footprint coverage of the targets. A near-optimal complete footprint coverage plan is proposed to guide the robot, but with an offline-constructed target graph. The least-cost route is selected to traverse the constructed target graph. Finally, we propose a complete online motion planning solution for the SIFC problem, called the online sensor-based complete coverage (${\tt oSCC}$) algorithm. The ${\tt oSCC}$ detects unknown targets using onboard sensors when the robot traverses the targets to conduct the filling actuation at the shortest distance. In experiments, the crack-filling robot is equipped with four omni-directional wheels to perform arbitrary direction motion, and an $XY$-table mechanism is used to drive a fluid nozzle for the filling action. Motion control of omni-directional-wheel robots was reported in~\cite{liu2008omni,indiveri2009swedish}. We formulate motion coordination between the mobile robot and the nozzle movements into a nonlinear model predictive control (MPC). Extensive experiments validate and demonstrate the proposed planning and control algorithms.

The main contribution of this work is the development of a new, complete, and empirically near-optimal motion planning and control approach for the SIFC problem in robotic crack mapping and filling applications. The proposed planning algorithms offer two key attractive features. First, the novel ${\tt oSCC}$ algorithm guarantees complete sensing coverage of the free space in the entire workspace while simultaneously achieving complete robotic footprint coverage of the detected targets. This algorithm aims to minimize the total distance travelled by the robot and achieves a near-optimal path in polynomial time. Second, the motion control of the robotic filling mechanism is coordinated with the planned mobile robot trajectory, enabling efficient execution of the robot's footprint task. The coupling between the sensor-based inspection and the onboard footprint coverage actuation is formulated and resolved by a new coordinated robot control design.
Compared to the presented conference publications~\cite{yu2019icra,guo2017optimal}, this paper introduces additional analyses for the motion planning and control design. It also presents extensive new experiments and detailed discussions using an upgraded crack-filling robot platform for improved motion planning and control. These developments provide valuable insights and a comprehensive understanding of the proposed approach in various scenarios, enhancing the applicability and reliability of the system.

The rest of the paper is organized as follows. In Section~\ref{overview}, we outline the problem statement and provide an overview of the planning and control algorithms. Section~\ref{coverage} discusses basic footprint coverage planning with known target information, while Section~\ref{sensorcoverage} extends to sensor-based online coverage planning. Robotic control for crack-filling actuation is covered in Section~\ref{control}. Experimental setup and results are presented in Sections~\ref{experiment} and~\ref{results}, respectively. Finally, concluding remarks are summarized in Section~\ref{sec_concl}.

\section{Problem Statement and Algorithms Overview}
\label{overview}

\subsection{Problem Statement}
\label{sec_design}

\begin{figure}[h!]
	\vspace{0mm}
	\centering
	\includegraphics[width=3.1in]{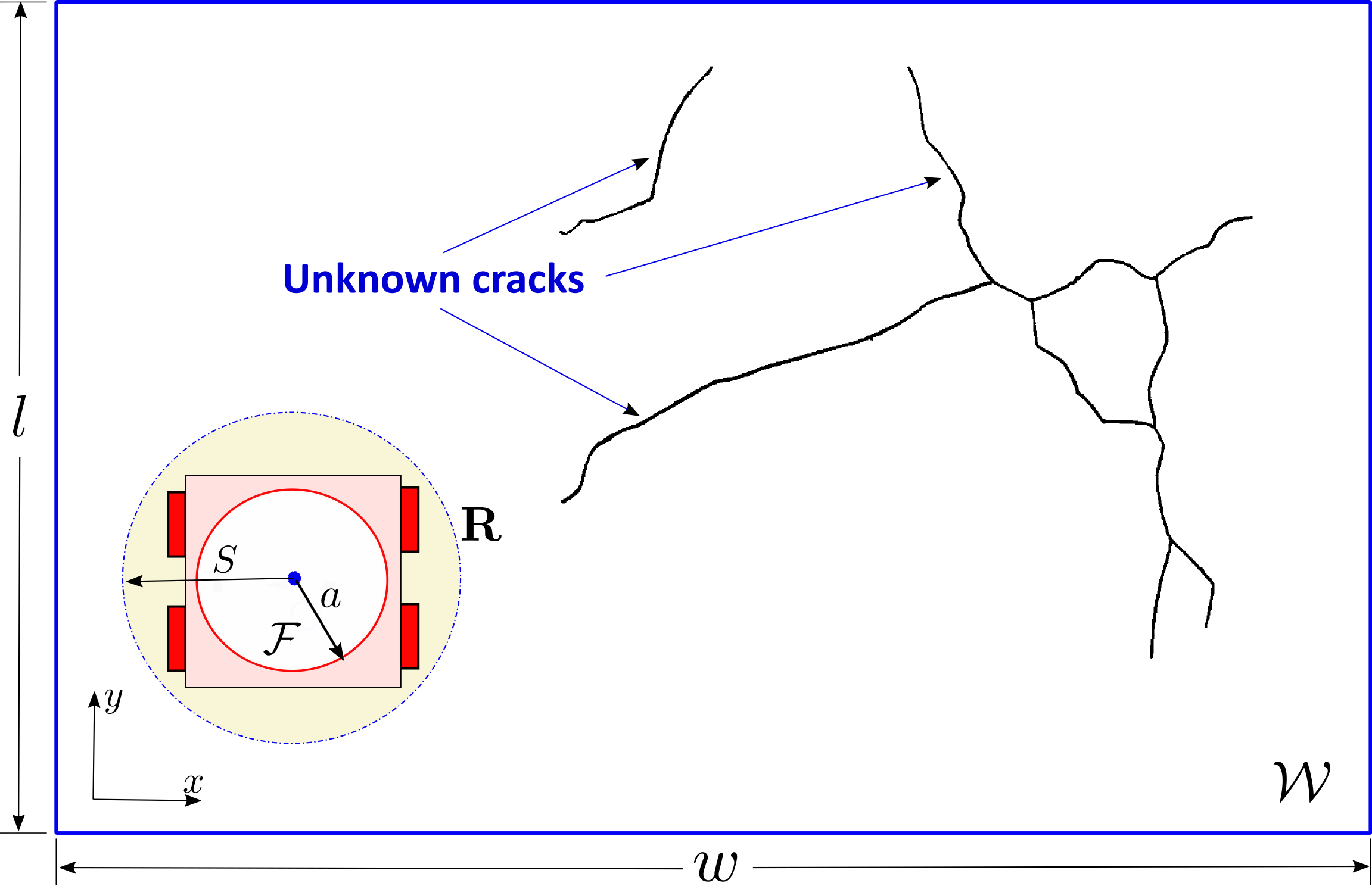}
	\caption{The illustration of robotic crack inspection and filling setup with unknown crack information in a rectangular workspace $\mathcal{W}$.}
	\label{conf}
\end{figure}

Fig.~\ref{conf} shows an illustrative SIFC setup for the robotic crack-filling application. We consider the coverage planning and motion control for a robot, denoted as $\mathbf{R}$, which is equipped with a crack detection sensor and filling actuator. $\mathbf{R}$ is in a compact free workspace $\mathcal{W} \subset \mathbb{R}^2$. { In general, the free space can be in any shape with a finite number of obstacles. For simplicity, we consider a known rectangular obstacle-free workspace with a size of $l \times w$, and the results are extendable to any other free space with different shape (readers can refer to~\cite{acar2002morse,acar2002sensor} for more details). 
Robot $\mathbf{R}$ needs to completely cover $\mathcal{W}$ to detect all unknown cracks (as targets) and simultaneously fill the detected cracks. $\mathbf{R}$ is assumed to freely move in any arbitrary direction in $\mathcal{W}$. The robot's footprint, denoted as $\mathcal{F}$, is a circular area around its geometric center with a radius of $a$. Any cracks within $\mathcal{F}$ can be repaired by the filling actuator. The onboard target detection sensor (e.g., a panoramic camera) can identify any cracks within a circular area of a radius of $S$ around the robot's geometric center. The target detection range is larger than the robot footprint size, that is, $S\geq a$.

To focus on the coverage planning problem, robot $\mathbf{R}$ is assumed to know its location in $\mathcal{W}$. Crack widths are assumed to be constant, and $\mathbf{R}$ only needs to consider the crack length for footprint coverage.
Additionally, we assume that the nozzle-filling motion and the robot's motion are coordinated at all times, ensuring that any targets under the robot's footprint can be reached by the nozzle. Therefore, our primary goal for the planner is to minimize the robot's total traveling distance.

{\em Problem Statement}: Given the unknown targets (e.g., cracks) in $\mathcal{W}$, the goal of the motion planner and controller for robot $\mathbf{R}$, with an onboard sensing range $S$ and a robot footprint size $a$, is to completely detect and footprint-cover all targets in $\mathcal{W}$ while minimizing the total distance traveled by the robot. The objective is to achieve near-optimal solutions, considering the uncertainties associated with the unknown targets.
The optimality of the planning algorithm involves finding an efficient path that minimizes the total distance traveled by the robot while covering all the cracks and scanning the entire free space. 

\subsection{Algorithm Design Overview}

To solve the above SIFC problem, we present a set of algorithmic developments. Fig.~\ref{overdiag} illustrates an overview of the planning algorithms and their relationships. First, we describe a robotic coverage planning algorithm, called graph-based coverage (${\tt GCC}$), to drive robot $\mathbf{R}$ to cover all targets by its footprint under a given target map. Note that ${\tt GCC}$ does not cover the entire workspace $\mathcal{W}$ using onboard detection sensors, but it serves as a basic planning module to cover the given target map by robotic footprint $\mathcal{F}$.

\begin{figure}[t!]
	\vspace{2mm}
	\centering
	\includegraphics[width=3.4in]{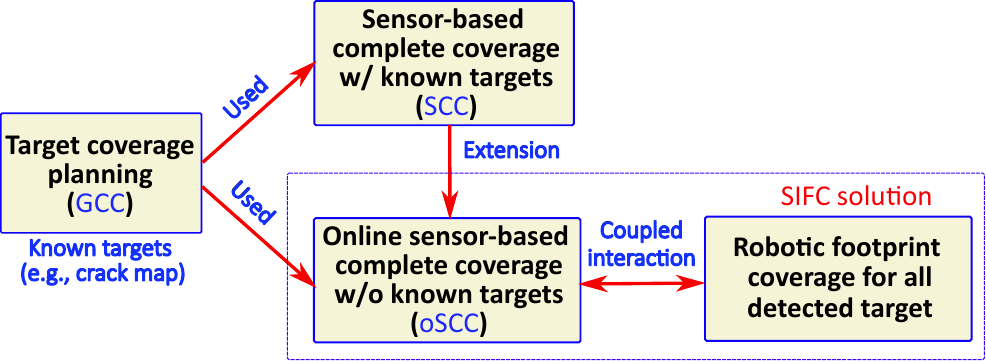}
	\caption{The overview of the SIFC planning and control algorithms.}
	\label{overdiag}
\vspace{-3mm}
\end{figure}

We then present a sensor-based complete coverage of workspace $\mathcal{W}$ assuming known target maps, which is denoted as the $\tt SCC$ algorithm. The $\tt SCC$ algorithm initially constructs a graph map from the provided target maps and subsequently utilizes the $\tt GCC$ to explore the entire workspace $\mathcal{W}$ and guide robot $\bf{R}$ to achieve footprint coverage of all targets while minimizing travel distance. Serving as the foundation for addressing the SIFC problem, the $\tt SCC$ algorithm plays a crucial role. By leveraging the insights gained from the edge connections observed in the $\tt SCC$ algorithm and its properties of completeness and near-optimality, we extend this approach to handle scenarios with unknown target information. This extension leads to the development of an online $\tt SCC$ algorithm, denoted as $\tt oSCC$, which is derived from $\tt SCC$ by relaxing the assumption of given target maps to achieve near-optimal sensor-based target detection and real-time robot footprint coverage to completely cover $\mathcal{W}$. Finally, crack filling control is designed to drive the onboard actuation mechanism to fill the detected targets with coordinated robot motion given by $\tt oSCC$. Both $\tt SCC$ and $\tt oSCC$ use ${\tt GCC}$ as part of the algorithmic module. The crack-filling control has to follow the dynamic constraints of the actuator and mechanical systems, considering the coupled robot motion. 
The sensor-based detection coverage offered by the $\tt oSCC$ algorithm and the robotic footprint coverage design collectively provide the SIFC solution.
In the next two sections, we present the $\tt GCC$, $\tt SCC$, and $\tt oSCC$ algorithms in detail.

\section{Crack Coverage Planning}
\label{coverage}

In this section, we present the $\tt GCC$ planner for robot footprint coverage of the given target (i.e., crack) map. 

\vspace{-2mm}
\subsection{Target Graph Construction}

We construct a target graph, denoted by $\mathbb{G}_c$, from crack images that are captured by the onboard camera sensors. { Because the $\tt GCC$ planner assumes known target maps, as shown in Figs.~\ref{conf} and~\ref{fig_GCC}, the images of the cracks were captured offline using the onboard camera and then stitched together to obtain the entire crack image $\mathcal{I}$.}  We first extract the crack skeletons from $\mathcal{I}$ and then dilate the skeletons by circular area with radius $a$, i.e., Minkowski sum with footprint size. For example, Fig.~\ref{fig_topology} illustrates the dilated skeletons of the cracks shown in Fig.~\ref{conf}. We define the union of all the dilated cracks as \textit{footprint region}, denoted by $\mathcal{M}_{\text{f}}$. The endpoints ($P_{\text{end}}$) and intersection points  ($P_{\text{int}}$)  of dilated cracks are used to build the nodes of $\mathbb{G}_c$, and the crack's extension directions are used to create the edges of $\mathbb{G}_c$. Multiple $P_{\text{end}}$ and $P_{\text{int}}$ are merged if their distances are within $a$, namely, those points are covered within $\mathcal{F}$. The edges of $\mathbb{G}_c$ are the shortest distance routes connecting the corresponding nodes inside $\mathcal{M}_{\text{f}}$. Fig.~\ref{fig_path} illustrates the example of the constructed graph $\mathbb{G}_c$. 

\begin{figure}[t!]
	\vspace{1mm}
	\hspace{-3mm}
	\subfigure[]{
		\label{fig_topology}
		\includegraphics[width=1.82in]{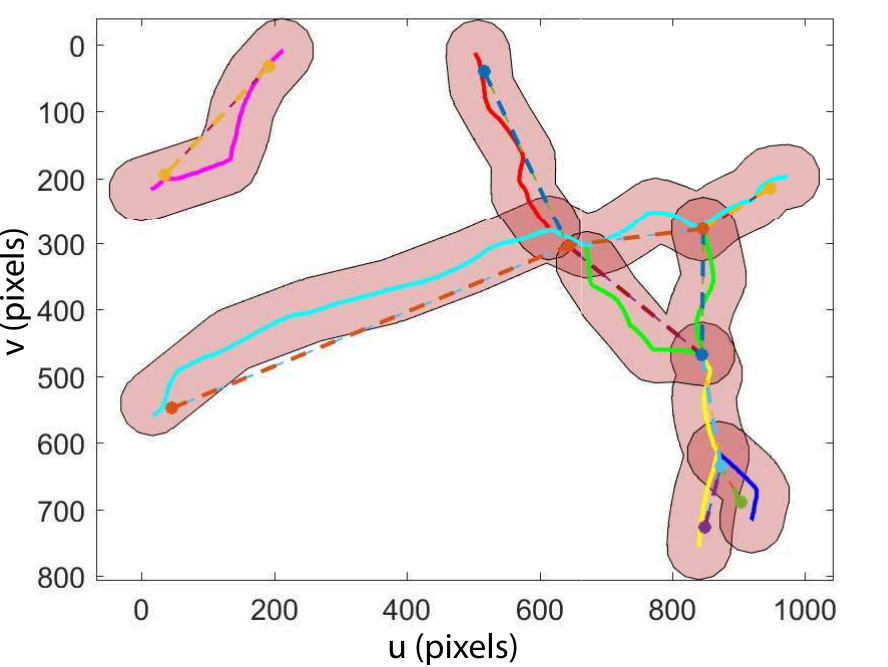}}
	\hspace{-6mm}
	\subfigure[]{
		\label{fig_path}
		\includegraphics[width=1.77in]{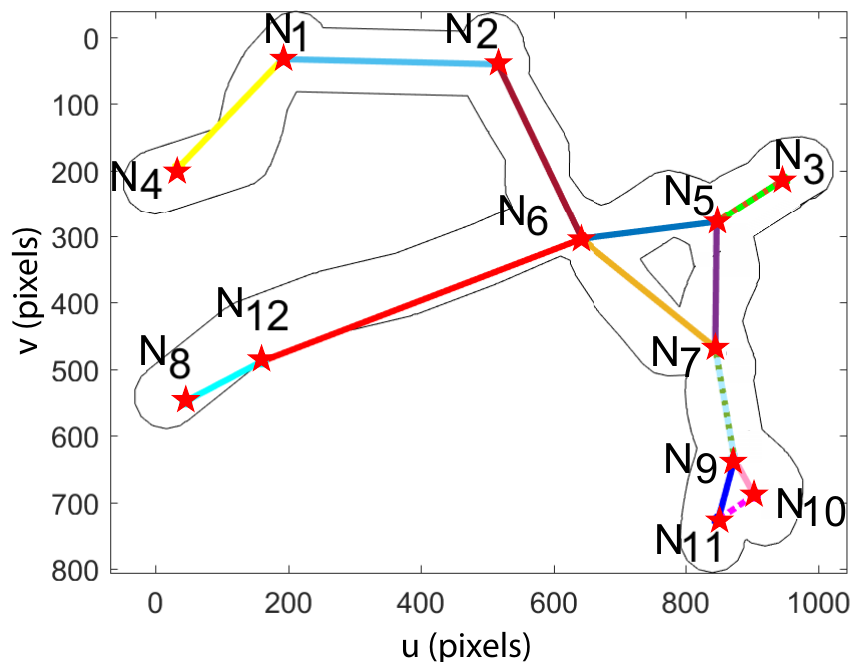}}
	\vspace{-3mm}
	\caption{(a) An example of the footprint region $\mathcal{M}_{\text{f}}$ (red shaded areas). The solid lines are the separated cracks. (b) The constructed $\mathbb{G}_c$. Points $N_1$ to $N_{12}$ shown in red stars are nodes of $\mathbb{G}_c$, and the solid lines are edges of $\mathbb{G}_c$. The dashed lines are the added edges in the ${\tt GCC}$ algorithm.}
	\label{fig_GCC}
	\vspace{-3mm}
\end{figure}

Algorithm~\ref{alg_motionplanning} describes the construction of graph $\mathbb{G}_c$. The input is the binary image $\mathcal{I}$, which is homography warped according to their actual shape. The first step is to use the skeleton method to find the topology $\mathcal{T}$ of  $\mathcal{I}$ (line 1). Endpoints of the cracks, $P_{\text{end}}$, are found from $\mathcal{T}$ by searching the 8 neighbors of each pixel. $\mathcal{T}$ is then separated into individual cracks, $\textit{crack}_\mathcal{I}$, by the endpoints and crack extension direction (line 2). Lines 3 and 4 find all $P_{\text{end}}$, with a distance larger than $a$ between each other to form end-nodes $N_{\text{end}}$. Footprint region $\mathcal{M}_\text{f}$ is obtained by the Minkowski sum (line~5). The intersected areas $A_{\text{int}}$ of all dilations are found, and their corresponding overlapping numbers $\textit{No}_{\text{over}}$, and centroid point $P_{\text{int}}$ are determined (lines 6-7). Among {$P_{\text{int}}$} and  $N_{\text{end}}$ being the final candidates for the graph nodes $N_{\text{cand}}$ (line 9), we select the nodes with a distance greater than $a$ with each other as the nodes $\bs{N}$ of $\mathbb{G}_c$ (line 11). The graph's edges $\bs{E}_c$ are obtained from its nodes and crack topology (line 12). Function ${\tt shortest\_path}$ is used to adjust the graph edges to guarantee their locations inside the Minkowski area by using visibility graph shortest path planning (line 13).


\begin{algorithm}[h!]
	\DontPrintSemicolon
	\label{alg_motionplanning}
	\caption {${\tt Crack\_Graph}$}
	\SetAlgoVlined
	\SetKwInOut{Input}{Input}
	\SetKwInOut{Output}{Output}
	\Input{$\mathcal{I}$}
	\Output{$\mathbb{G}_c$}
	\nl $\mathcal{T} \gets {\tt topology} (\mathcal{I})$, $P_{\text{end}} \gets {\tt get\_endpoint} (\mathcal{T})$ \;
	\nl $\textit{crack}_\mathcal{I} \gets {\tt get\_cracks} (\mathcal{T}, P_{\text{end}})$ \;
	\For{$ \text{each } P_{\text{end}} \rightarrow e_i $}{
		\nl $P_{\text{test}} \gets P_{\text{end}}/e_i$\;
		\lIf{ $ \text{all} ( {\tt distance} (e_i, P_{\text{test}})) > a$}\
		{\nl $N_{\text{end}} \gets {\tt add\_node}(e_i)$}
	}
	\nl $\mathcal{M}_{\text{f}} \gets \textit{crack}_\mathcal{I}\oplus a$ \;
	\nl $(A_{\text{int}}, \textit{No}_{\text{over}}) \gets {\tt get\_intersection}(\mathcal{M}_{\text{f}})$ \;
	\nl $P_{\text{int}} \gets {\tt centroid} (A_{\text{int}})$ \;	
	\nl $(\textit{crack}_\mathcal{I}, N_{\text{end}}) \gets {\tt shorten}(\textit{crack}_\mathcal{I},N_{\text{end}})$\;
	\nl $N_{\text{cand}} \gets {\tt combine} ( P_{\text{int}}, N_{\text{end}})$ \;
	\nl $N_{\text{cand}} \gets {\tt sort} (N_{\text{cand}}, \textit{No}_{\text{over}})$ \;
	\For{$\text{each } N_{\text{cand}} \rightarrow n_i$ }{
		\lIf{ ${\tt distance}(n_i, \bs{N}) > a$}\
		{\nl $\bs{N} \gets {\tt add\_node}(n_i)$}
	}
	\nl $\bs{E}_c \gets {\tt search\_edge}(\bs{N}, \mathcal{T})$\;
	\nl $\mathbb{G} \gets \{ \bs{N}, \bs{E}_c \}$, $\mathbb{G}_c \gets {\tt shortest\_path}(\mathbb{G},\mathcal{M}_{\text{f}})$\;
\end{algorithm}

We need to take special consideration for the different formed angles by cracks. Fig.~\ref{Acute_fig} illustrates a few examples of $\mathbb{G}_c$ with different formed angles. Fig.~\ref{Acute_fig:a} shows a general case where two vertex points (i.e., $N_1$ and $N_2$) are far away with a distance larger than $a$. When a vertex point on $\mathcal{T}$ forms an acute angle, it overlaps with its own Minkowski sum area, which makes the nodes visible to each other; see Fig.~\ref{Acute_fig:b}. In this case, the shortest path does not transverse the crack, which leads to unfilled cracks. To achieve the shortest crack-filling path, the vertex must be at least a distance of $a$ from the boundary of the Minkowski sum area. These cracks are pre-identified, and their respective Minkowski offset values are adjusted (line 5 in Algorithm~\ref{alg_motionplanning}). Fig.~\ref{Acute_fig:c} shows the Minkowski sum area with the adjusted Minkowski offset value. By doing so, it guarantees that the calculated shortest path covers the entire crack in ${\tt shortest\_path}$.

\begin{figure}[t!]
	  \vspace{-3mm}
	\centering
	\subfigure[]{
		\includegraphics[height=0.9in]{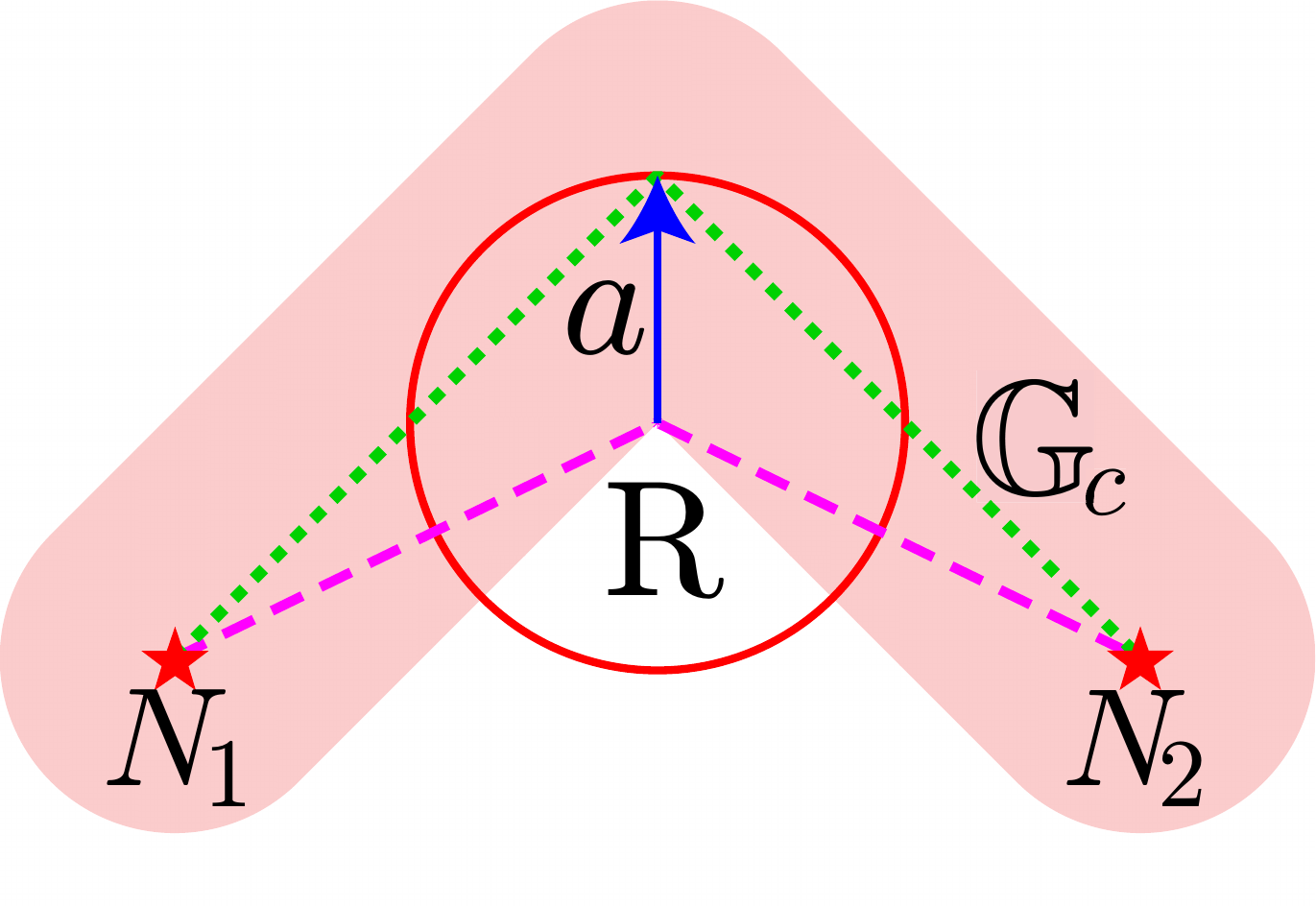}
		\label{Acute_fig:a}}
	\hspace{3mm}
	\subfigure[]{
		\includegraphics[height=0.9in]{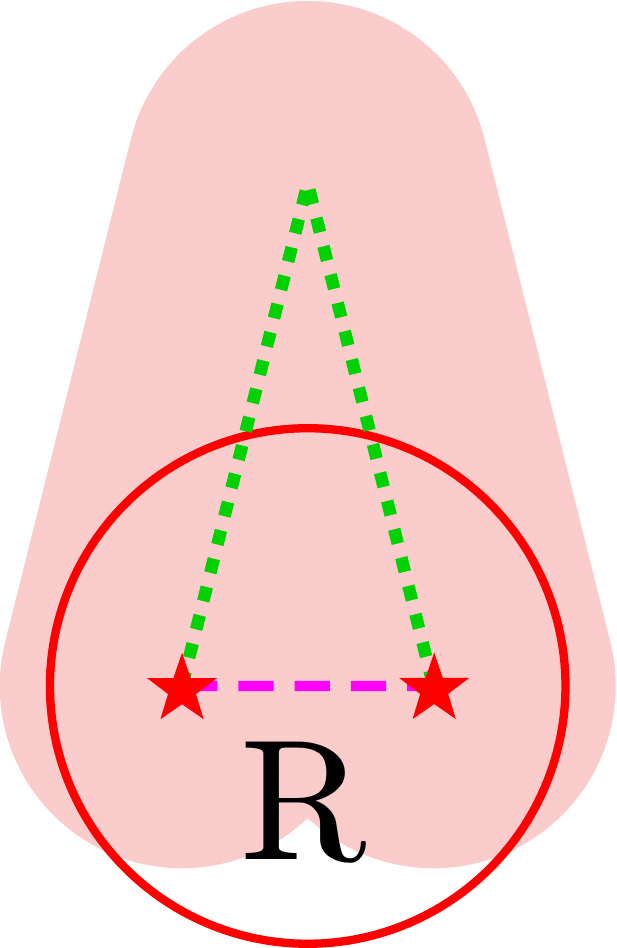}
		\label{Acute_fig:b}}
	\hspace{3mm}
	\subfigure[]{
		\includegraphics[height=0.9in]{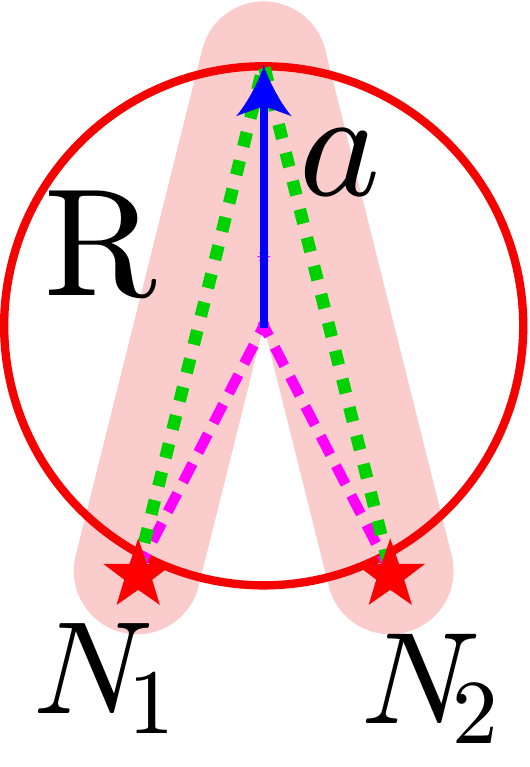}
		\label{Acute_fig:c}}
	\vspace{-2mm}   
	\caption{Illustration of crack graphs with different formed angles. Robot footprint $\mathcal{F}$ is illustrated by a red circle. The graph $\mathbb{G}_c$ is shown as dotted green lines, and the red stars $N_1$ and $N_2$ represent its nodes. The pink-shaded areas are the footprint region $\mathcal{M}_{\text{f}}$. The red dashed lines are the shortest paths to connect nodes $N_1$ and $N_2$ within $\mathcal{M}_{\text{f}}$. (a) Every point on $\mathbb{G}_c$ is covered by $\mathcal{F}$ as robot $\mathbf{R}$ travels along the shortest path. (b) With sharp crack angles, robot $\mathbf{R}$ cannot fully cover $\mathbb{G}_c$. (c) By adjusting the Minkowski sum area, the shortest path is achieved to ensure full crack coverage.}  
	\label{Acute_fig}
  \vspace{-3mm}
\end{figure}

\subsection{Target Coverage Planning}
\label{CCP}

With $\mathbb{G}_c$, we consider the required properties of the graph to guide the robot to traverse the cracks with minimum cost. All the graph vertices except the first and last in the route must have an even number of connected edges. Otherwise, no route exists in the graph to allow traveling along each edge exactly once, meaning the robot would get stuck at odd vertices. To prevent this, vertices with an odd number of connected edges are made even by adding edges. To achieve the shortest traveling distance, the robot motion planner must search for all the different ways to pair off the vertices with odd numbers of connected edges and choose the pair that adds the least total distance to the graph.

\begin{figure*}[h!]
	\hspace{-3mm}
	\centering
	\subfigure[]{
		\label{GraphConf:a}
		\includegraphics[width=2.33in]{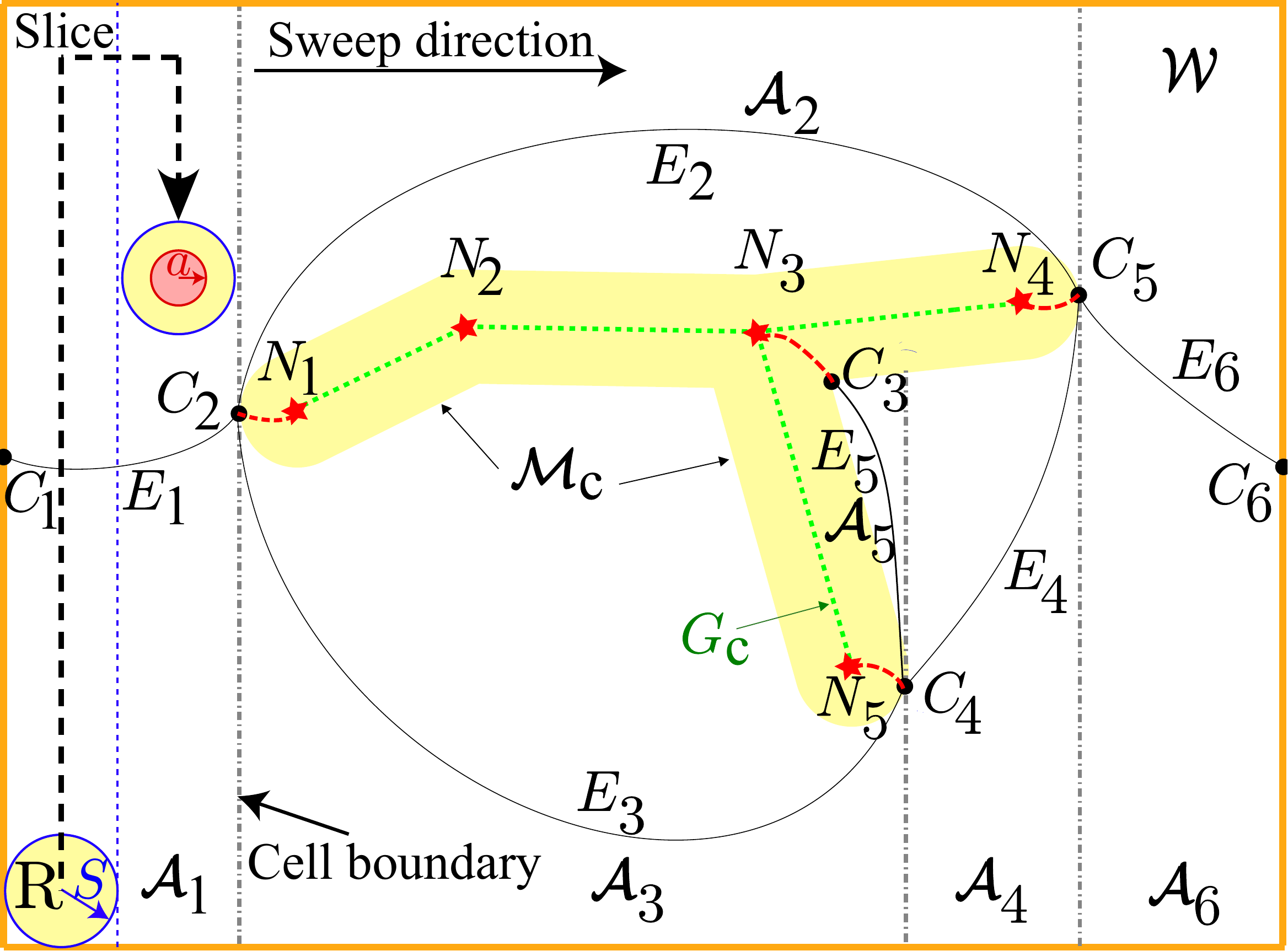}}
	\hspace{-1.5mm}
	\subfigure[]{
		\label{GraphConf:b}
		\includegraphics[width=2.33in]{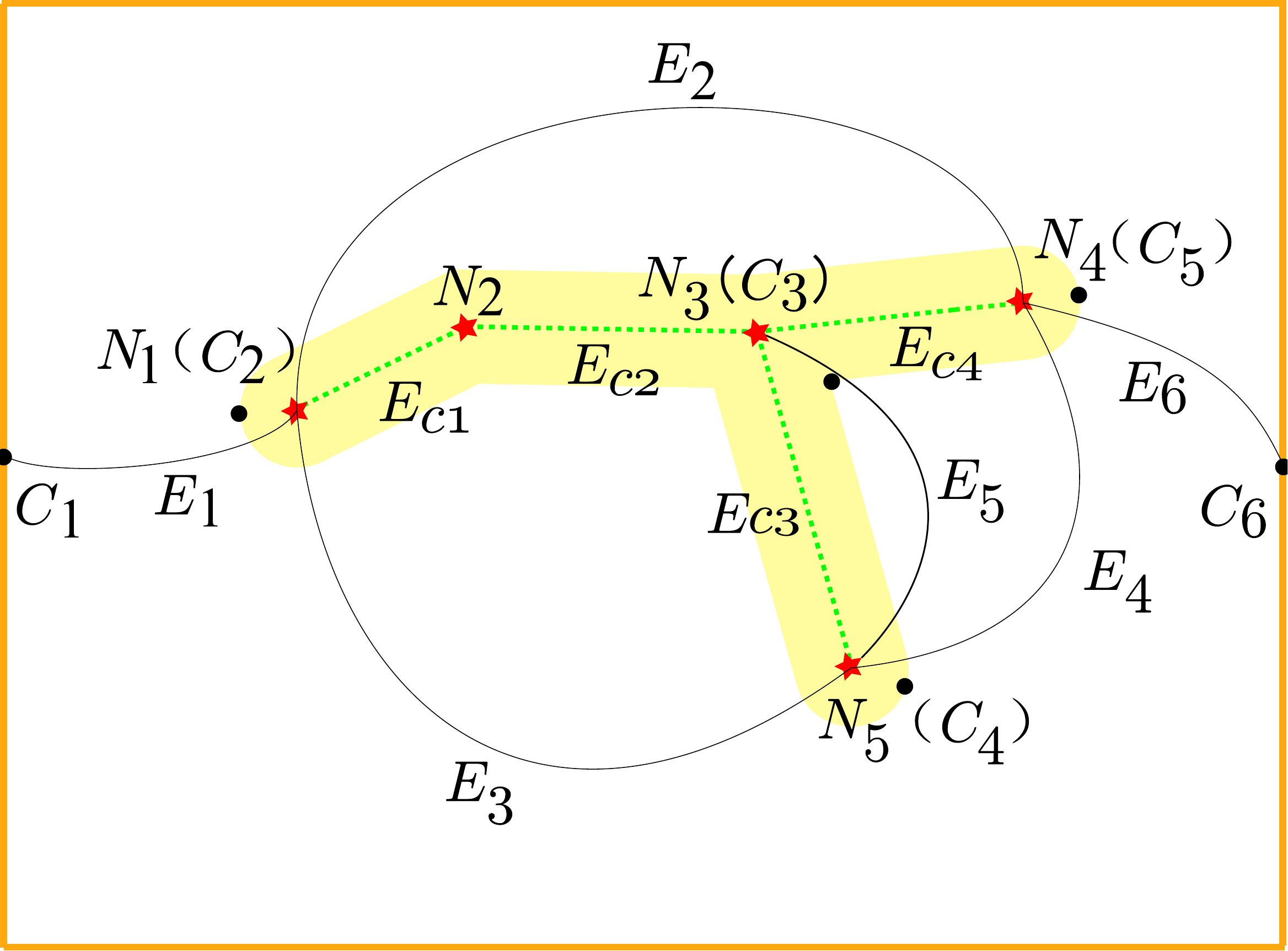}}
	\hspace{-1.5mm}
	\subfigure[]{
		\label{GraphConf:c}
		\includegraphics[width=2.33in]{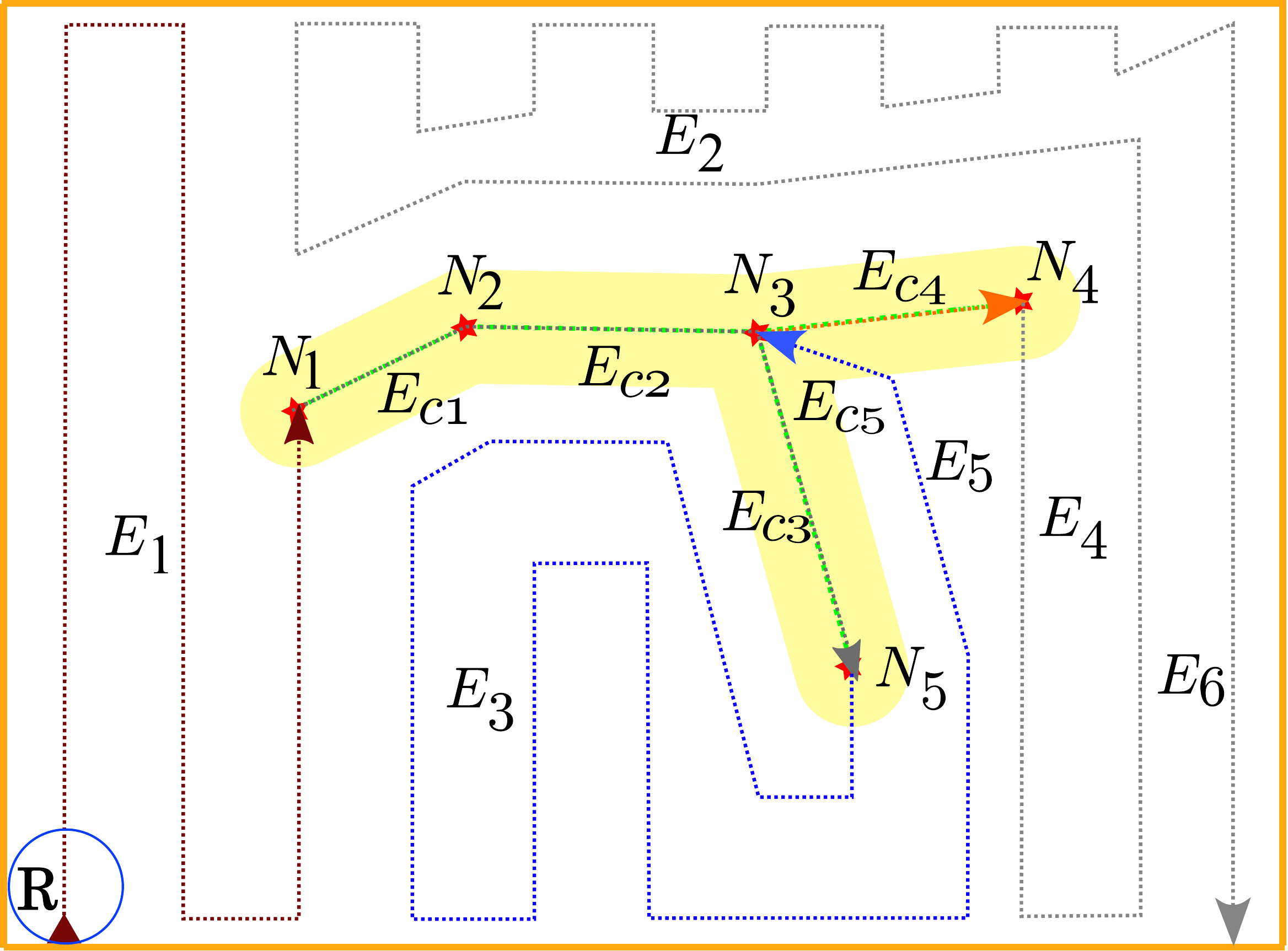}}
	\vspace{-3mm}
	\caption{Illustration of the SIFC planner. (a) The MCD with $\mathcal{M}_\text{c}$ (highlighted in yellow). $\mathcal{A}_i$, $C_i$, and $E_i$, $i=1,\cdots,6$, represent the Reeb graph's cells, nodes, and edges, respectively. The graph $\mathbb{G}_c$ is shown in a dotted green line, and red stars represent the nodes. The Reeb graph $\mathbb{G}_\text{w}$ of the MCD is connected with $\mathbb{G}_c$. The red-dash edges are added to the combined $\mathbb{G}_\text{w}$ and $\mathbb{G}_c$ to form an Euler tour. (b) The simplified $\mathbb{G}_\text{w}$ and $\mathbb{G}_c$. Each critical point on the boundary of the $\mathcal{M}_\text{c}$ is combined with its corresponding node in $\mathbb{G}_c$. (c) The robot path $\mathcal{P}_R$ is depicted in dotted lines, with arrows indicating the direction of travel.}
	\label{GraphConf}
	\vspace{-3mm}
\end{figure*}

The task of finding the shortest route covering all the graph edges is similar to the Chinese Postman Problem (CPP)~\cite{kwan1962graphic}. However, the crack edges might not form a connected graph and instead form subsets of graphs. As a result, when such situations arise, we adapt the rural postman problem (RPP) solution~\cite{christofides1981algorithm}. The RPP was demonstrated to be NP-complete, and heuristic solution procedures were proposed to approximate the solution~\cite{pearn1995algorithms}. Unlike in any postman problem where the postman typically needs to finish at the same location as the starting point, robot $\mathbf{R}$ can start and finish the job at different locations. Therefore, two nodes with an odd number of connected edges are left unpaired and selected as the starting and ending positions, respectively.

Algorithm~\ref{alg_GCC} illustrates the ${\tt GCC}$ planning algorithm. The algorithm formulates an RPP problem to guarantee that the robot covers all the cracks with the least number of revisits. The algorithm comprises three phases. In the first phase, the minimum spanning tree is computed over $\mathbb{G}_c$ to get a subset of the edges $\mathbb{E}_\text{t}$ that connects all the vertices with the minimum possible total edge weight (line 1). The union of $\mathbb{G}_c$ and $\mathbb{E}_\text{t}$ ensures a single connected network for the postman problem. In the second phase, using matching theory~\cite{edmonds1973matching}, we search for all possible ways to pair up the odd vertices $N_{\text{odd}}$~(line 2) by describing the solution as a linear programming~(LP) polyhedron. $\bs{E}_{\text{add}}$ represents the collection of all sets of added edges for every potential pairing, while $\mathbb{E}_\text{add}(i)$ indicates the added edge for the $i$th potential pairing. $C_{\text{total}}$, $C_{\text{max}}$, and $C_{\text{final}}$ denote the total, maximum, and final edge cost, respectively. The edge cost is formulated as the line distance.  $E_{\text{final}}$ is selected as the minimum cost of the total added edges minus the maximum-cost single added edge. The maximum-cost single edge is broken into the starting and ending nodes of the path (lines 3 to 6). The graph is updated by adding $E_{\text{final}}$ (line 7). In the final phase, the function ${\tt fleury}$~\cite{fleury} is used to obtain the optimal path $\mathcal{P}_c$ for $\mathbf{R}$. For example, as shown in Fig.~\ref{fig_path}, the optimal path is obtained as a sequence $\mathcal{P}_c: N_4 \rightarrow N_1 \rightarrow N_2 \rightarrow N_6 \rightarrow N_5 \rightarrow N_3 \rightarrow N_5 \rightarrow N_7 \rightarrow N_9 \rightarrow N_{10} \rightarrow N_{11} \rightarrow N_{9} \rightarrow N_{7} \rightarrow N_{6} \rightarrow N_{12} \rightarrow N_{8}$.  

\begin{algorithm}[ht!]
	\DontPrintSemicolon
	\label{alg_GCC}
	\caption {${\tt GCC}$}
	\SetAlgoVlined
	\SetKwInOut{Input}{Input}
	\SetKwInOut{Output}{Output}
	\Input{$\mathbb{G}_c$}
	\Output{$\mathcal{P}_c$}
	\nl $ \mathbb{E}_\text{t} \gets {\tt MST}(\mathbb{G}_c)$, $\mathbb{G}_c \gets \mathbb{G}_c\cup \mathbb{E}_\text{t}$ \;
	\nl $ N_{\text{odd}} \gets {\tt find\_oddNode}(\mathbb{G}_c)$, $\bs{E}_{\text{add}} \gets {\tt pair}(N_{\text{odd}}, \mathbb{G}_c)$ \;
	\For{\text{each} $\bs{E}_{\text{add}} \rightarrow {\mathbb{E}_\text{add}}(i) $}{
		\nl $C_{\text{max}} \gets {\tt max} ({\mathbb{E}_\text{add}}(i).\textit{edgeCost})$ \;
		\nl $C_{\text{total}}(i) \gets {\tt sum} ({\mathbb{E}_\text{add}}(i).\textit{edgeCost})-C_{\text{max}}$ \;
		\nl  $(C_{\text{final}}, \textit{Index}) \gets {\tt min} (C_{\text{total}})$ \;
	}
	\nl $E_{\text{final}} \gets {\mathbb{E}_\text{add}}(\textit{Index}).\textit{edge}$\; \nl $\mathbb{G}_c \gets {\tt add}(\mathbb{G}_c,E_{\text{final}})$,
	$\mathcal{P}_c  \gets {\tt fleury}(\mathbb{G}_c)$
	\vspace{1mm}
	\BlankLine
	\vspace{-1mm}
\end{algorithm}

Since the distance between any two nodes in $\mathbb{G}_c$ is guaranteed to be at least $a$, the resulting path $\mathcal{P}_c$ is always less than or equal to the complete coverage of the target map. If the target map is dense enough, then the graph degrades into the complete coverage problem, and the final cost is the same as that of the ``lawn mowing'' problem~\cite{LaValle2006}. 

\section{Sensor-Based Complete Coverage Planning}
\label{sensorcoverage}

In this section, we first discuss the ${\tt SCC}$ planner to cover the entire workspace $\mathcal{W}$ with known target information and then generalize the algorithm to ${\tt oSCC}$ with unknown targets. 
  \vspace{-3mm}
\subsection{${\tt SCC}$ Planner}

To explain our approach, we borrow the following definitions from~\cite{acar2002morse,acar2002sensor}. As shown in Fig.~\ref{GraphConf:a}, the \textit{slice} is effectively a vertical line sweeping from left to right along the \textit{sweep direction} in $\mathcal{W}$. A \textit{cell} is an area where slice connectivity does not change, and changes in the connectivity of the slice only occur at \textit{critical points}. A critical point is located on the boundary of an object whose surface normal is perpendicular to the sweep direction. Critical points are used to determine the \textit{cell boundaries}. A \textit{target region}, denoted by $\mathcal{M}_\text{c}$, is obtained by dilating $\mathbb{G}_c$ with a circular disk area with a radius of $S$. For example, $\mathcal{M}_\text{c}$ is the yellow area in Fig.~\ref{GraphConf:a}.

Unlike most MCD-based coverage path planning problems, the critical points defined here are not only on the boundary of obstacles but also on the boundary of the target regions. According to the MCD of the free space $\mathcal{W}$, a Reeb graph~\cite{acar2002sensor,acar2002morse} is constructed and denoted as $\mathbb{G}_\text{w}$. In Fig.~\ref{GraphConf:a}, $\mathbb{G}_\text{w}=(\bs{C},\bs{E})$ is shown as the black solid lines. The nodes of $\mathbb{G}_\text{w}$ are the critical points $\bs{C}=\{C_i\}$, and the edges $\bs{E}=\{E_i\}$ connect the neighboring critical points. The edge $E_i$ in the Reeb graph directly corresponds to the cell $\mathcal{A}_i$ in the free space, where $i=1,\cdots,6$ in the figure.

\begin{algorithm}[h!]
	\DontPrintSemicolon
	\label{alg_sGCC}
	\caption {${\tt SCC}$}
	\SetAlgoVlined
	\SetKwInOut{Input}{Input}
	\SetKwInOut{Output}{Output}
	\Input{$\mathcal{W}, \mathcal{I}, S$}
	\Output{$\mathcal{P}_R$}
	\nl $\mathbb{G}_c \gets {\tt Crack\_Graph}(\mathcal{I})$, $(\bs{C},\bs{E}) \gets {\tt MCD}(\mathcal{W} \setminus (\mathbb{G}_c \oplus S))$\;
	\nl $\mathbb{G}_{\text{w}} \gets{\tt Reeb\_graph}(\bs{C},\bs{E})$, $\mathcal{P}_\text{wc}  \gets{\tt GCC}(\mathbb{G}_{\text{w}}\cup \mathbb{G}_{c})$\;
	\nl  $\textit{conn} \gets {\tt cell\_connect}(\mathcal{P}_\text{wc},\bs{E})$\;
	\nl  $\mathcal{P}_R \gets {\tt  complete\_coverage} (\mathcal{P}_\text{wc},\bs{E},\textit{conn},S)$\;
\end{algorithm}
Algorithm~\ref{alg_sGCC} briefly describes the ${\tt SCC}$ planner. With $\mathbb{G}_\text{w}$ and $\mathbb{G}_c$, we search for the shortest cost route covering all the graph edges at least once; that is, we connect edges in both graphs to form an Euler tour with the least cost. In the case of an unspecified ending position, all of the graph nodes except for the initial and ending vertices must have an even number of connected edges (i.e., an even degree); otherwise, the robot could get stuck at vertices with odd degrees. Therefore, the degree of vertices is maintained even by adding edges. For example, as shown in Fig.~\ref{GraphConf:a}, $C_1$-$C_6$, $N_1$, and $N_3$-$N_5$ are the vertices with odd degrees, and the red dashed edges connecting \{$C_2$,$N_1$\}, \{$C_3$,$N_3$\}, \{$C_5$,$N_4$\}, and \{$C_4$,$N_5$\} are added to the graph to form the shortest route. The resulting Euler tour, denoted as $\mathcal{P}_\text{wc}$, provides an order of edges that the robot should visit (line 2). The connecting distance between each adjacent cell in the Euler tour is minimized in the function ${\tt cell\_connect}$. The variable \textit{conn} represents the connections of adjacent cells in the Euler tour (line 3). Finally, complete coverage is performed by following the sequence of edges in the Euler tour, resulting in the robot path $\mathcal{P}_R$ (line 4).

The back-and-forth motion path is generated to cover the interior of the cell sequences using the onboard sensor. Once the robot is joined with $\mathbb{G}_\text{c}$, it transitions to following the path $\mathcal{P}_\text{c}$ generated by the $\tt GCC$ to traverse the crack graph edges, where $\mathcal{P}_\text{c}$ is a part of $\mathcal{P}_\text{wc}$. By following the crack graph edges, the targets are covered by the robot footprint $\mathcal{F}$. The back-and-forth coverage motion is well documented in~\cite{choset2001coverage,acar2002morse}. If an edge of $\mathbb{G}_\text{w}$ is doubled in the resulting Euler tour, the corresponding cell is split in half. 
The robot has the ability to adjust the height of its coverage for each slice incrementally, thereby enabling it to control the exit point of each cell~\cite{mannadiar2010optimal}. Thus, by minimizing the distance from the exit to the entry points of the next cell, the connecting distance of each cell is minimized in function ${\tt cell\_connect}$. 

As the Reeb and crack graphs provide a complete model of $\mathcal{W}$ and each edge of the Euler tour is traversed exactly once, the proposed algorithm guarantees complete and near-optimal coverage of all the cracks and $\mathcal{W}$ with the minimized traveling distance. The proof of the completeness and near-optimality of the ${\tt SCC}$  algorithm follows the same approach as in references~\cite{xu2014efficient, mannadiar2010optimal}, drawing upon principles from cellular decomposition and Euler tour theory. The Reeb graph, which serves as a comprehensive model of the environment, ensures that all available free space is covered by traversing each edge exactly once. This guarantees complete coverage of all free space. The resulting Euler tour establishes a systematic order for visiting the cells of the Reeb graph without covering any area twice. While backtracking may be necessary to re-position the robot at reachable corners of the next cell to be covered, it typically adds at most one extra sweep line in length. Although some backtracking can be avoided in certain cell configurations by adjusting the order of traversal through two loops~\cite{mannadiar2010optimal}, even when unavoidable, the areas covered twice are usually much smaller compared to the total area. Additionally, as the environment size increases, the percentage of repeat coverage per cell decreases~\cite{xu2014efficient}. This ensures near-optimality. The optimal traversal ordering, equivalent to the Euler tour, can be efficiently computed in polynomial time~\cite{edmonds1973matching}.

\vspace{-2mm}
\subsection{${\tt oSCC}$ Planner}

When target information is unknown, robot $\mathbf{R}$ needs to detect all targets and the critical points online in $\mathcal{W}$. We first combine graphs $\mathbb{G}_\text{c}$ and $\mathbb{G}_\text{w}$ using the following lemma and proposition with proofs given in Appendices~\ref{proof1} and~\ref{proof2}, respectively. 
\begin{lem}
	\label{lem1}
	Each critical point generated by $\mathcal{M}_\text{c}$ corresponds to one node of $\mathbb{G}_\text{c}$. 
\end{lem}
\begin{prop}
\label{lem2}
Optimality is preserved with the choice of connecting critical points of $\mathcal{M}_\text{c}$ to corresponding nodes in~$\mathbb{G}_\text{c}$.
\end{prop}

According to Proposition~\ref{lem2}, we simplify $\mathbb{G}_\text{c}$ and $\mathbb{G}_\text{w}$ by combining the critical points of $\mathcal{M}_\text{c}$ with their corresponding nodes of $\mathbb{G}_\text{c}$. Fig.~\ref{GraphConf:b} shows an example of the simplified graph by such action. Based on the simplified graph, we propose the ${\tt oSCC}$ planning algorithm with unknown targets. The robot first treats the target regions $\mathcal{M}_\text{c}$ as obstacles in MCD. Then, robot $\bf{R}$ follows the coverage path $\mathcal{P}_R$ until it finds those combined critical points to enter $\mathcal{M}_\text{c}$ and follows $\mathbb{G}_\text{c}$ according to the current Euler tour. The ${\tt oSCC}$ algorithm is a practical extension of ${\tt SCC}$ where robot $\mathbf{R}$ stores and incrementally constructs the crack graph $\mathbb{G}_\text{c}$ online. As the robot navigates through the workspace $\mathcal{W}$, it continuously scans for new cracks and updates $\mathbb{G}_\text{c}$ whenever it encounters a node (such as end points or interaction points) of the crack graph. To construct the incremental crack map, the robot utilizes existing scanned information of the crack to update the crack map, focusing only on the nodes of the cracks as shown in Algorithm 1. Subsequently, the robot enters the target region and follows the constructed crack graph until one edge ends, while simultaneously conducting footprint coverage.

\begin{figure*}[ht!]
	\hspace{-5mm}
	\subfigure[]{
		\label{traj:b}
		\includegraphics[width=1.85in]{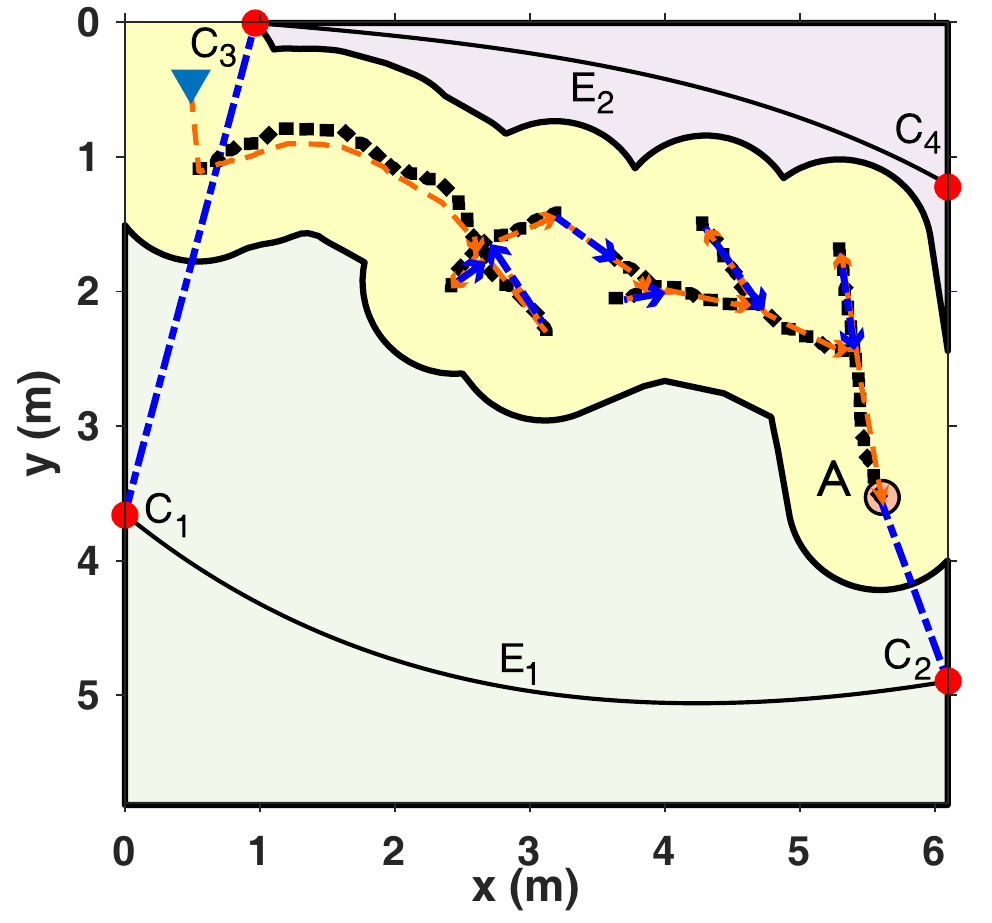}}
	\hspace{-4mm}
	\subfigure[]{
		\label{traj:c}
		\includegraphics[width=1.81in]{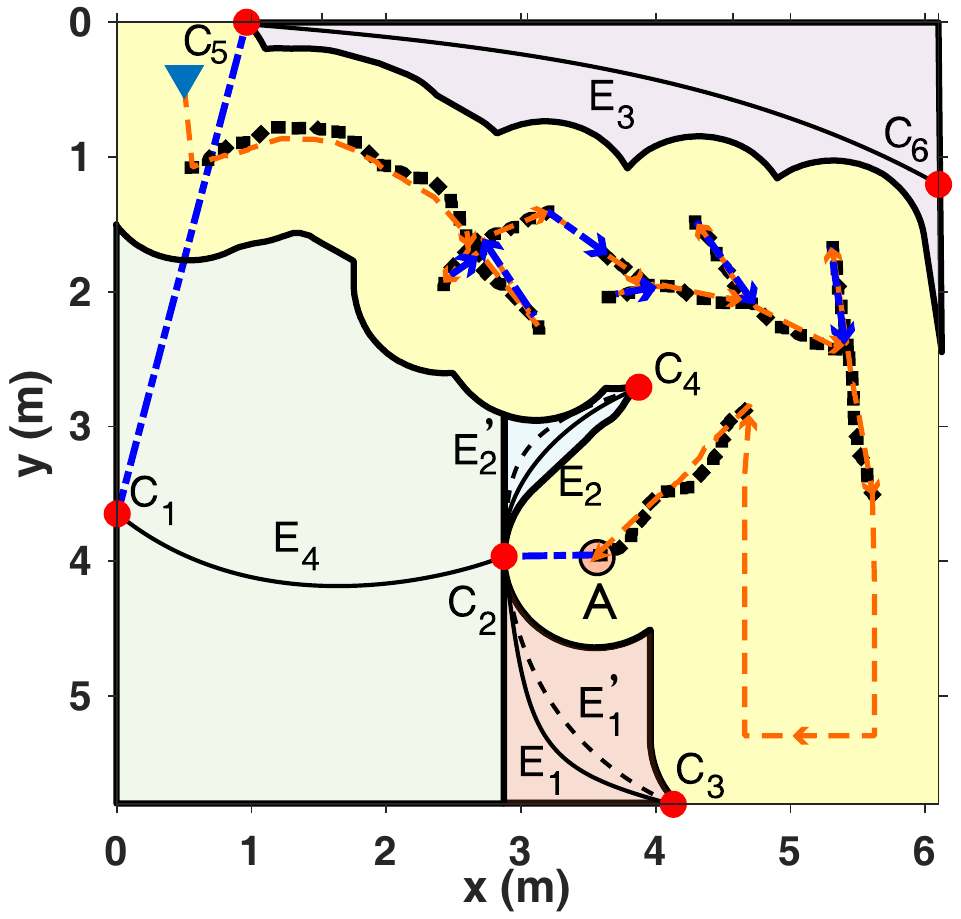}}
	\hspace{-4mm}
	\subfigure[]{
		\label{traj:d}
		\includegraphics[width=1.81in]{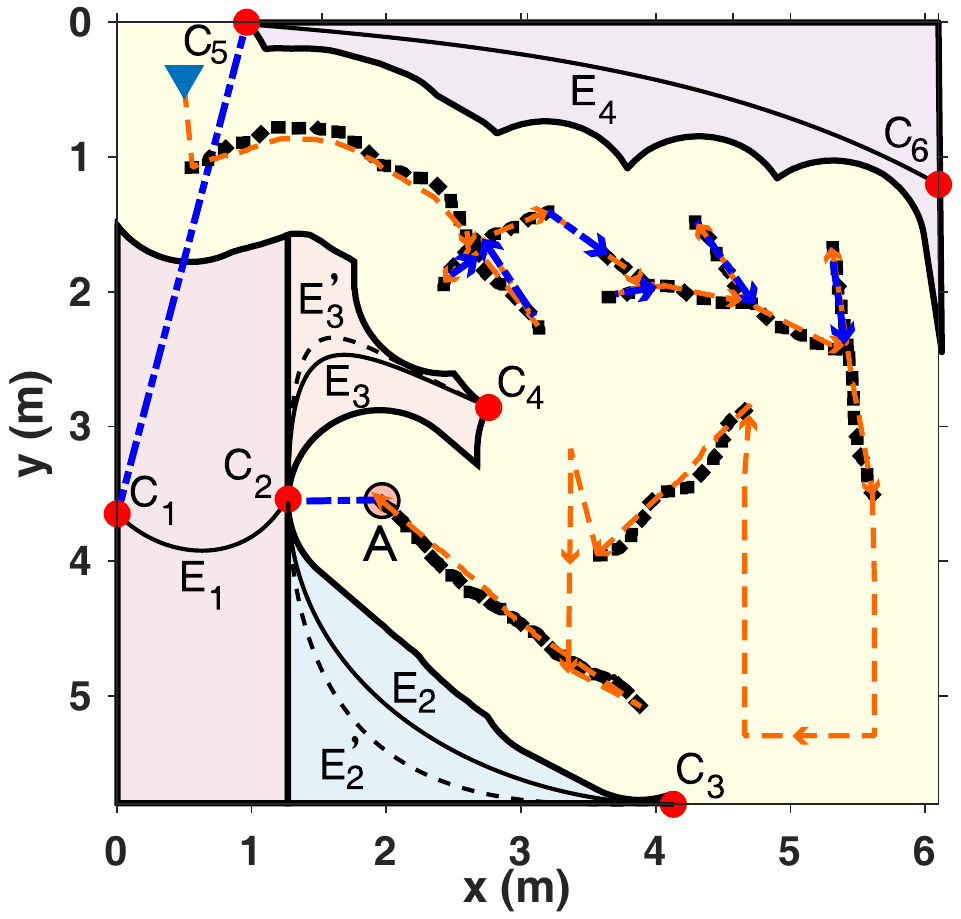}}
	\hspace{-4mm}
	\subfigure[]{
		\label{oSCCpath}
		\includegraphics[width=1.8in]{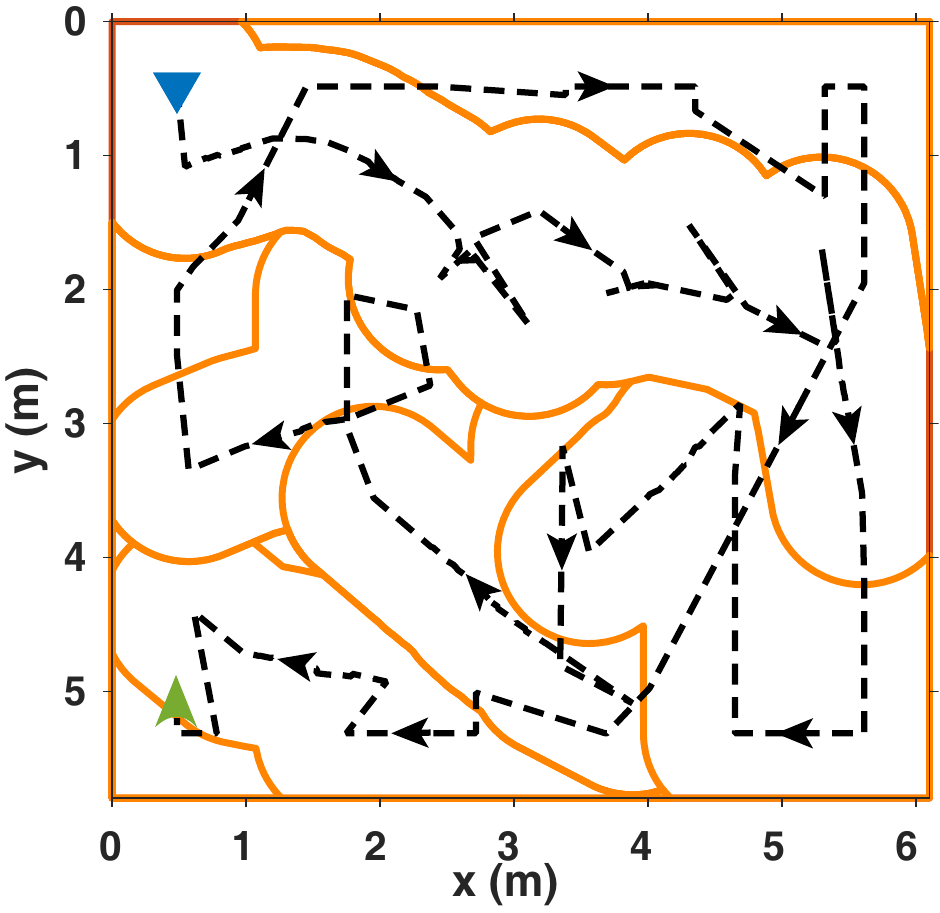}}
	\vspace{-2mm}
	\caption{An illustrative planning example by the ${\tt oSCC}$ algorithm. (a)-(c) The Reeb graphs as additional cracks are detected during the loop cycles of the ${\tt oSCC}$ algorithm. All covered cracks are marked by thick, dotted black curves. The robot starts from the top-left corner. Each Reeb graph is updated when the robot is at point $A$ (highlighted in a circle). Those Reeb graphs are constructed after removing already-covered areas (yellow shaded areas). The red dots ($C_i$) denote the nodes of the Reeb graph. The solid curves ($E_i$) are the edges. The doubled edges are indicated by the thin, dashed black curves ($E_i'$). The bold boundaries are the cell boundaries. The orange-dash lines are the robot paths. The blue-dotted dashed lines are the connections between two nodes. (d) The final path using the ${\tt oSCC}$ algorithm. The final MCD of the free space is plotted in the thick orange curves. The final trajectory is shown by the black dashed lines. The robot starts from the top-left corner and ends at the bottom left, marked by a triangle and an arrow, respectively. }
	\label{oSCC_traj}
	\vspace{-3mm}
\end{figure*}

Algorithm~\ref{alg_oSCC} shows the structure of the ${\tt oSCC}$ algorithm with three sections: initialization (lines 2 to 5), footprint coverage (lines 6 to 8), and sensing coverage (lines 9 to 11). The already footprint-covered area is denoted as $\mathcal{M}_\text{cov}$. In the initialization section, robot $\mathbf{R}$ computes Reeb graph $\mathbb{G}_\text{w}$ for the uncovered area $\mathcal{W} \backslash \mathcal{M}_\text{cov}$ (line 2) without any target information $\mathcal{T}$. Function ${\tt Reeb\_seq}$ computes the transverse sequence for $\mathbb{G}_\text{w}$ using a priority queue based on the {\it cell} location and area, as well as the number of successive edges (children). An optimal and complete coverage path $\mathcal{P}_R$ is generated for $\mathcal{W} \backslash \mathcal{M}_\text{cov}$ (line 5). In the coverage section, robot $\mathbf{R}$ follows the path calculated in the initialization section until a node of $\mathbb{G}_\text{c}$ is encountered (lines 9 to 11). Finally, the robot enters $\mathcal{M}_\text{c}$ to construct $\mathbb{G}_{c}$, computes the path using the ${\tt GCC}$ planner (line 6), and follows $\mathbb{G}_{c}$ until it ends (line 7). Line 8 shows the update of the covered area $\mathcal{M}_\text{cov}$. When no uncovered cells or edges remain in $\mathbb{G}_\text{w}$, the workspace is optimally and completely covered. An example of the robot path $\mathcal{P}_R$ generated by Algorithm~\ref{alg_oSCC} is illustrated in Fig.~\ref{GraphConf:c}.

\begin{algorithm}[t!]
	\DontPrintSemicolon
	\label{alg_oSCC}
	\caption {${\tt oSCC}$}
	\SetAlgoVlined
	\SetKwInOut{Input}{Input}
	\SetKwInOut{Output}{Output}
	\Input{$\mathcal{W}, \mathcal{I}, S$}
	\Output{$\mathcal{P}_R$}
	\nl $\mathcal{M}_\text{cov} \gets \O$, $\mathcal{T} \gets {\tt get\_topology}(\mathcal{I})$\;
	\While{$ \mathcal{M}_\text{cov} \neq \mathcal{W}$}
	{
		\nl $(\bs{C},{\bs{E}}) \gets {\tt MCD}(\mathcal{W\;\setminus\; M}_\text{cov})$\;
		\nl $\mathbb{G}_\text{w} \gets {\tt Reeb\_graph}(\bs{C},{\bs{E}})$, $\bs{\Pi}_w \gets {\tt Reeb\_seq}(\mathbb{G}_\text{w})$\;
		\nl $\textit{conn} \gets {\tt cell\_connection}(\bs{\Pi}_w,{\bs{E}})$\;
		\nl $\mathcal{P}_R \gets {\tt complete\_coverage}(\bs{\Pi}_w,{\bs{E}},\textit{conn},S)$\;
		\While{$true$}{	
			\If{A node of crack $\mathcal{T}$ is found}{
				\nl $\mathbb{G}_{c} \gets {\tt Crack\_Graph}(\mathcal{I})$, $\bs{\Pi}_c \gets {\tt GCC}(\mathbb{G}_{c})$\;
				\nl ${\tt follow\_path}(\bs{\Pi}_c(i).\textit{nodes})$\;
				\nl $\mathcal{M}_\text{cov} \gets \mathcal{M}_\text{cov} \cup (\mathbb{G}_{c} \oplus S)$\;
				\textbf{break}
			}
			\nl $P_\text{n} \gets $ the next step of $\mathcal{P}_R$, $\tt follow\_path$($P_\text{n}$) \;
			\nl $\mathcal{M}_\text{cov}\gets \mathcal{M}_\text{cov}\cup (P_\text{n} \oplus S),\; \mathcal{I} \gets {\tt get\_image} $\; 
			\nl $\mathcal{T} \gets {\tt get\_topology}(\mathcal{I})$\;
		}
	}
	
\end{algorithm}

In Algorithm~\ref{alg_oSCC}, after traversing one target edge, we remove the covered areas, and update $\mathbb{G}_\text{w}$ of the remaining space $\mathcal{W} \backslash \mathcal{M}_\text{cov}$ to avoid passing the same target twice to reach another uncovered cell. Taking the example shown in Fig.~\ref{traj:b}, the robot is currently on node $A$ and $\mathbb{G}_\text{w}$ is updated after removing the already covered areas (yellow shaded areas). The odd nodes in the resulting $\mathbb{G}_\text{w}$ include $C_1$, $C_2$, $C_3$, and $C_4$. Therefore, $AC_2$ and $C_1C_3$ are connected to form the least-cost Euler tour, and $C_4$ represents the path's ending node. When the resulting Euler tour needs to double edges in $\mathbb{G}_\text{w}$, the corresponding cell is split into two components. The first part is covered by a wall-following motion, where a wall is defined as the boundaries of $\mathcal{W}\backslash \mathcal{M}_\text{cov}$. The other part is covered by the zigzag motion of the leftover space in the cell. As shown in Fig.~\ref{traj:c}, edges $E_1$ and $E_2$ are doubled, and in Fig.~\ref{traj:d}, edges $E_2$ and $E_3$ are doubled. Note that the splitting of cells does not increase the cost of covering the whole cell. To minimize the cost, we select the coverage motion direction according to the next connected edge in the path.

We have the following property for the ${\tt oSCC}$ planner, with proof given in Appendix~\ref{proof3}.
\begin{prop}
	\label{lem3}
	{The ${\tt oSCC}$ algorithm guarantees completeness in coverage planning and results in the least-cost Euler tour for constructing the traversal ordering at the cell level, ensuring no redundancy in terms of individual cell coverage. By eliminating redundancy in individual cell coverage, it results in the most efficient path, minimizing the robot’s travel distance to locally cover all the individual cells when the connections of each covered cell are not considered.}
\end{prop}
 
The ${\tt oSCC}$ algorithm ensures that each individual cell in the free space is covered exactly once, thereby avoiding redundancy in terms of individual cell coverage when the connections of each covered cell are not considered. However, due to the unknown dimensions of the cells, the robot may not always have enough information to minimize the zigzag motion connecting adjacent cells according to the Euler tour sequence.
The optimality gap in our approach is primarily related to the connections between adjacent cells based on the Euler tour sequence. For worst-case analysis of the suboptimality bound, assuming that the connections between each pair of adjacent cells are small enough to be negligible in the optimal solution, the worst connection between each pair of adjacent cells resulting from the ${\tt oSCC}$ could be the number of free cells in $\mathbb{G}_\text{w}$, denoted as $N_\text{cell}$, minus one (i.e., the number of connections), multiplied by $\sqrt{l^2+(2S)^2}$ (i.e., the longest robot travel distance for one connection), where $l$ is the slice length (the side length of the free space). Therefore, in the worst case, the difference from the optimal solution is bounded by $(N_\text{cell}-1)\sqrt{l^2+(2S)^2}$.
The simplified $\mathbb{G}_\text{w}$ reduces the number of edges. The algorithm is solved in polynomial time because of the structure of $\mathbb{G}_\text{w}$ and is therefore used efficiently online.

\begin{figure}[h!]
	\vspace{-1mm}
	\hspace{-4mm}
	\subfigure[]{
		\includegraphics[width=1.75in]{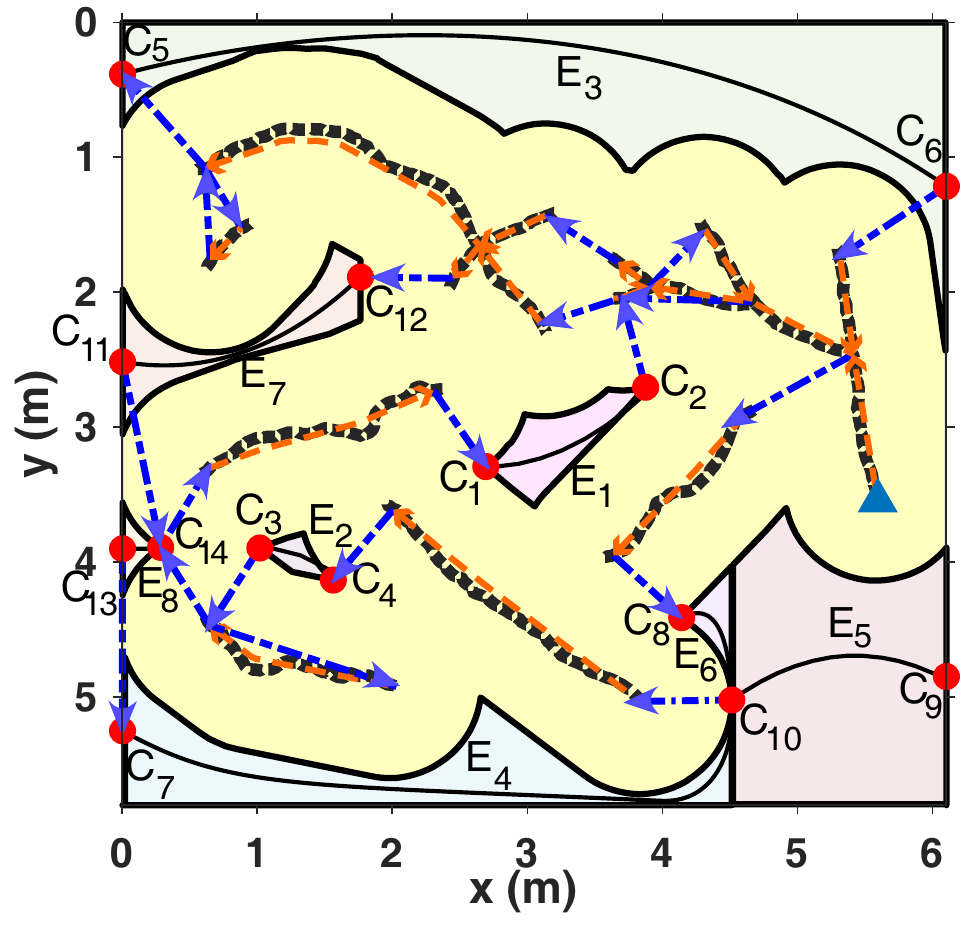}
		\label{SCCcell}}
	\hspace{-4mm}
	\subfigure[]{
		\includegraphics[width=1.75in]{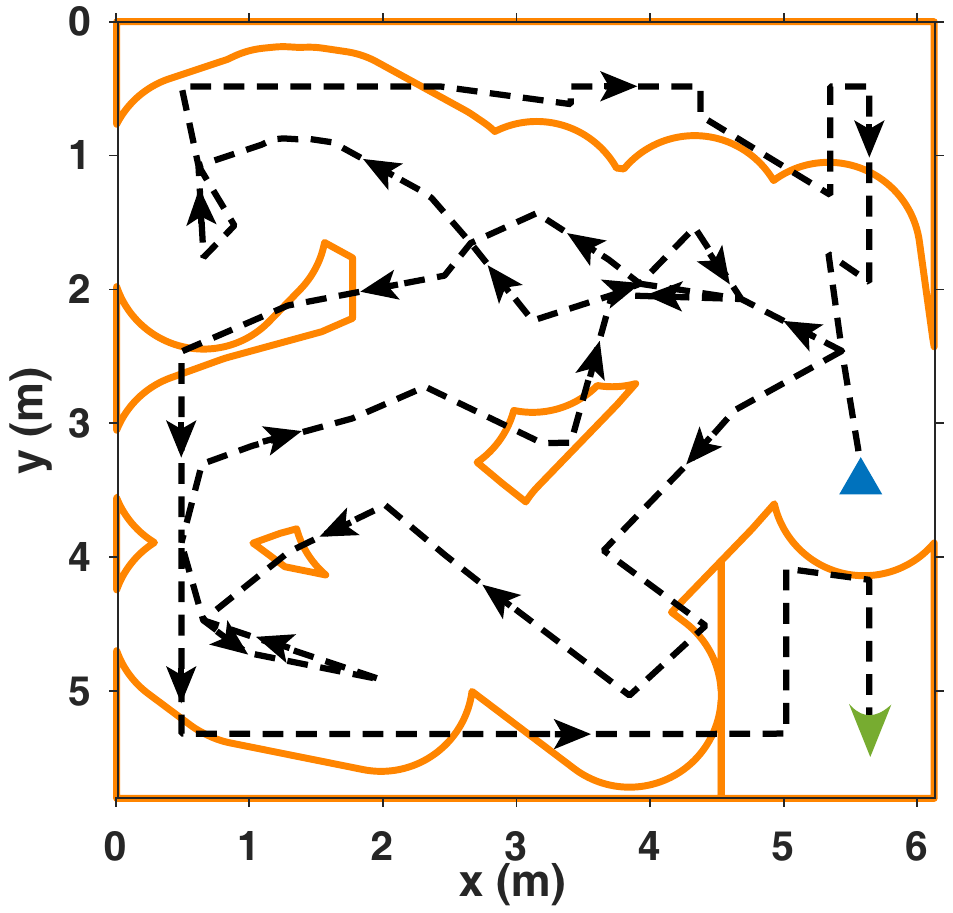}
		\label{SCCpath}}
	\vspace{-4mm}
	\caption{Planning result by the ${\tt SCC}$ algorithm with the cracks as in Fig.~\ref{oSCC_traj}. (a) The MCD and the corresponding $\mathbb{G}_\text{w}$ using ${\tt SCC}$. The orange dashed lines are the constructed $\mathbb{G}_\text{c}$. The blue-dotted dashed lines show the node connections. (b) The final route of the robot. The legends are the same as those in Fig.~\ref{oSCC_traj}.}
	\label{SCC_traj}
	\vspace{-2mm}
\end{figure}

We further illustrate the ${\tt oSCC}$ and ${\tt SCC}$ planners through an example. Fig.~\ref{oSCC_traj} shows the planning result under ${\tt oSCC}$ in $\mathcal{W}$, with dimensions $l=5.79$~m and $w=6.10$~m, and robot configuration, $S=0.69$~m and $a=8.9$~cm. The crack information (thick, black, dotted lines) is initially unknown. Under the ${\tt oSCC}$ planner, the robot first follows the initial path planned by Algorithm~\ref{alg_oSCC} until detecting a crack node. Then it follows the online updated $\mathbb{G}_\text{c}$ and uses the ${\tt GCC}$ algorithm to generate local paths for the footprint coverage of the scanned cracks. The covered regions are then removed from $\mathcal{W}$, and $\mathbb{G}_\text{c}$ and $\mathbb{G}_\text{w}$ are updated accordingly. Figs.~\ref{traj:b}-\ref{traj:d} show the updated graphs, and Fig.~\ref{oSCCpath} illustrates the final path. For comparison, Fig.~\ref{SCC_traj} presents the planning results under the ${\tt SCC}$ algorithm with known crack information. The constructed MCD of the free space is shown in Fig.~\ref{SCCcell}, along with the corresponding $\mathbb{G}_\text{w}$ (solid lines), $\mathbb{G}_\text{c}$ (dash lines), and $\mathcal{M}_\text{c}$ (yellow shaded areas). The least-cost Euler tour is constructed by the blue dotted dash lines. Fig.~\ref{SCCpath} demonstrates the final route of the robot. For further details, readers can refer to the companion video clip. Experiments and comparisons will be discussed in Section~\ref{results}.

\vspace{-2mm}
\section{Crack Filling Motion Planning and Control}
\label{control}

In this section, we present the robot motion control to follow the planned trajectory and also the coordinated motion of the crack-filling action. 

\vspace{-3mm}
\subsection{Robot Kinematic Models}

Fig.~\ref{robotplan:a} shows the bottom view of the crack-filling robot. The robot is equipped with four independently driving omni-directional wheels, and therefore, it can move in any direction with free rotation. Fig.~\ref{robotplan:b} illustrates the driving mechanism for the filling nozzle, denoted as $\mathbf{N}_z$. Two frames are used in robot modeling: a global frame $\mathcal{N}(x,y)$ and a body frame $\mathcal{B}(x_b,y_b)$. The robot footprint is assumed to be square, with its center $O$ equidistant from four wheels, denoted by $W_i$, $i=1, \cdots, 4$. The distance between $O$ and $W_i$ is denoted as $R_d$. The nozzle is driven by two step motors (denoted as $W_5$ and $W_6$) with two timing belts and moves along the $x_b$- and $y_b$-axis in $\mathcal{B}$.

\begin{figure}[t!]
	\hspace{-2mm}
	\subfigure[]{
		\label{robotplan:a}
		\includegraphics[width=1.67in]{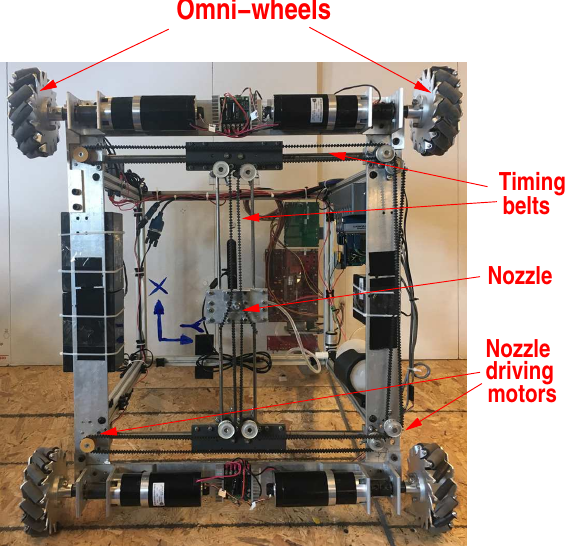}}
	\hspace{-2mm}
	\subfigure[]{
		\label{robotplan:b}
		\includegraphics[width=1.78in]{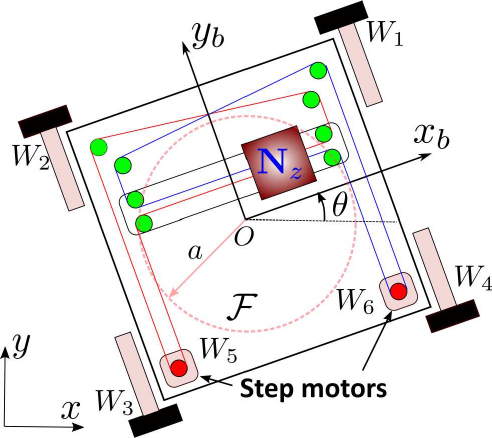}}
	\vspace{-2mm}
	\caption{(a) The bottom view of the crack-filling robot. (b) A schematic of the filling nozzle driving mechanism and robotic kinematic configuration.}
	\label{robotplan}
	\vspace{-0mm}
\end{figure}

The robot pose is captured by $\bs{q}_r=[x_r \; y_r \; \theta]^{\mathsf{T}}$ in $\mathcal{N}$, where $(x_r,y_r)$ and $\theta$ are the position of robot center $O$ and the robot orientation, respectively. The position of $\mathbf{N}_z$ in $\mathcal{B}$ is denoted as $\bs{q}_n=[x_n \; y_n]^{\mathsf{T}}$. We denote the angular velocity for $W_i$ as $\omega_i$, $i=1,\cdots,6$, and define  $\bs{\omega}_r=[\omega_1 \; \omega_2 \; \omega_3 \; \omega_4]^{\mathsf{T}}$ and $\bs{\omega}_n=[\omega_5 \; \omega_6]^{\mathsf{T}}$. Assuming no wheel slip and no deformation of the timing belts, the kinematic models for robot motion in $\mathcal{N}$ and nozzle relative motion in  $\mathcal{B}$ are obtained as
\begin{equation}
	\bs{\omega}_r=\bs{A}_r \dot{\bs{q}}_r, \; \bs{\omega}_n=\bs{A}_n {\dot{\bs{q}}}_n,
	\label{eq_kinematic_model}
\end{equation}
where
\begin{equation}
	\bs{A}_r=\frac{\sqrt{2}}{R_w}\begin{bmatrix}
		-\s_{\theta_1} & \c_{\theta_1} & R_d \\
		-\s_{\theta_2} & \c_{\theta_2} & R_d \\
		-\s_{\theta_3} & \c_{\theta_3} & R_d \\
		-\s_{\theta_4} & \c_{\theta_4} & R_d \\
	\end{bmatrix}, \; \bs{A}_n=\frac{1}{R_g}\begin{bmatrix} 1 & -1 \\ 1 & 1 \\
	\end{bmatrix},
	\label{eq30}
\end{equation}
$R_g$ is the radius of the driving pulley for $W_5$ and $W_6$, and $R_w$ is the radius of the robot wheel. In~(\ref{eq30}), we use notations $\s_{\theta}=\sin \theta$ and $\c_{\theta}=\cos \theta $ for $\theta$ and other angles. Angles $\theta_1=\theta + \pi/4$ and $\theta_{i+1}=\theta_{i} + \pi/2$, $i=1,2,3$.

\vspace{-3mm}
\subsection{Filling Nozzle Motion Planning}

The nozzle motion needs to be coordinated with the robot's motion to efficiently fill cracks within $\mathcal{F}$; see Fig.~\ref{robotplan:b}. Note that multiple cracks can be located within $\mathcal{F}$ at a time, and we need to determine how to move the nozzle to fill these cracks while the robot is in motion. Generally, the nozzle motion is much faster compared to the robot's movement velocity. Therefore, for simplicity, we neglect the time duration for the nozzle to move from one crack to another without performing filling action and only consider the movement time along cracks during the filling action.

Fig.~\ref{nozzleplan:a} illustrates the geometric relationship between the robot trajectory $\mathcal{P}_R$ and multiple cracks within $\mathcal{F}$. We use the arc length of the path $\mathcal{P}_R$, denoted as $s$, as the parameter to characterize any arbitrary point $p$ on the cracks. For a point $p$ on a crack within $\mathcal{F}$, we define a projection map $\pi(p): \;{p} \mapsto p_c(s)$, where $p_c(s)\in \mathcal{P}_R$, as the minimum distance to $p$. Considering that there are $n_c$ cracks located within $\mathcal{F}$, where $n_c \in \mathbb{N}$, we denote the mapping $\pi_i(p)$ for the $i$th crack by the above definition, where $i=1,\cdots,n_c$. We assume that $\pi_i(p)$ is bijective so that its inverse $\pi^{-1}_i(p)$ exists. By using $\pi_i(p)$, all points on the $n_c$ cracks inside $\mathcal{F}$ are mapped onto $\mathcal{P}_R$.

\begin{figure}[t!]
	\hspace{-1mm}
	\subfigure[]{
		\label{nozzleplan:a}
		\includegraphics[width=1.7in]{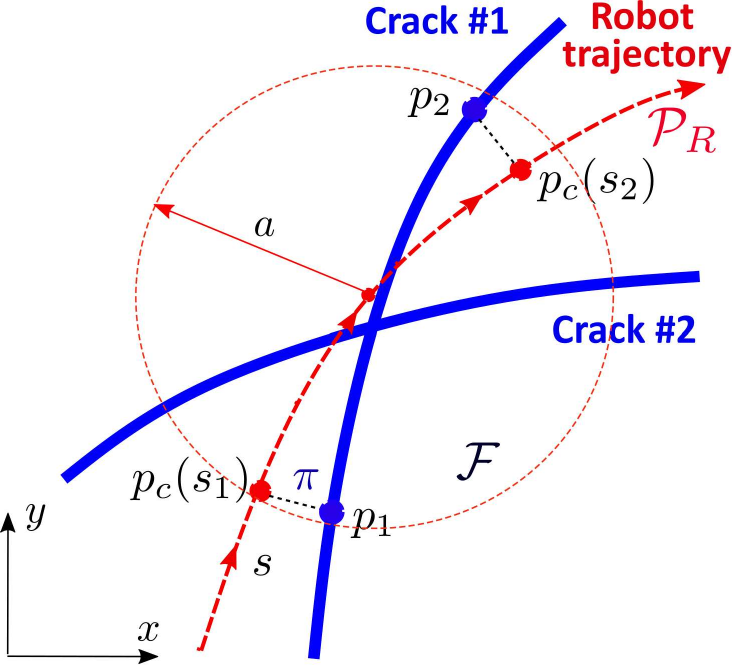}}
	\hspace{-0mm}
	\subfigure[]{
		\label{nozzleplan:b}
		\includegraphics[width=1.5in]{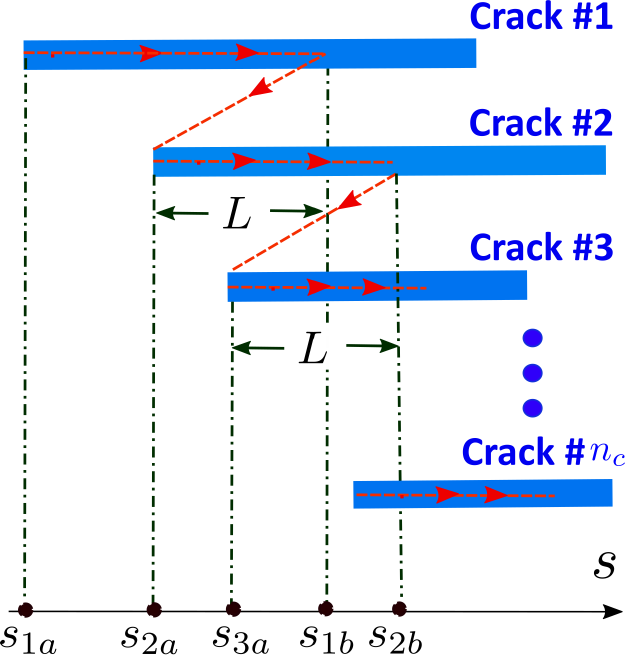}}
	\caption{(a) Schematic of the projection mapping $\pi_i$ from multiple cracks to the robot path $\mathcal{P}_R$. (b) An illustrative example of the filling nozzle planning across multiple cracks in $\mathcal{F}$.}
	\label{nozzleplan}
	\vspace{-0mm}
\end{figure}

Value $s$ increases along the robot $\mathbf{R}$'s moving direction, and points on $n_c$ cracks within $\mathcal{F}$ with the same $s$ value have the same priority for filling. To determine the filling sequences, we consider the nozzle $\mathbf{N}_z$ to move and switch the filling action among these $n_c$ cracks after staying along one crack for a threshold distance $L$. Fig.~\ref{nozzleplan:b} illustrates the nozzle travel sequence among all cracks within $\mathcal{F}$. Algorithm~\ref{alg_nozzle_planning} describes the nozzle motion sequence planning. Input $\{\bs{p}_i\}_{1}^{n_c}$ is the point sequence sets of all cracks; that is, $\{\bs{p}_i\}$ contains the point sequence of the $i$th crack, $i=1,\cdots,n_c$. ${\tt startpoint}$ (line 1) is to find the first point of each crack. The function ${\tt findmin}$ is to find the minimum value and the corresponding crack index. The notation $h(i)/h(I_1)$ indicates removing $h(I_1)$ from all $h(i)$ (line 3). The function ${\tt addpoint}$ (line 4) is to add an interval point of the crack $I_1$ to $^s\bs{p}_m$, and the interval is from $p_1$ to $p_2+L$. Finally, we obtain the output {$\{\bs{p}_n\}$ that represents the point sequence of the nozzle in $\mathcal{B}$.
	
\begin{algorithm}[t!]
		\DontPrintSemicolon
		\label{alg_nozzle_planning}
		\caption {Nozzle Motion Planning}
		\SetAlgoVlined
		\SetKwInOut{Input}{Input}
		\SetKwInOut{Output}{Output}
		\Input{$\{\bs{p}_i\}_{1}^{n_c}$}
		\Output{$\{\bs{p}_n\}$}
		\nl $\bs{p}_c^i \leftarrow \pi_i(\bs{p}_i)$, $ h(i) \leftarrow {\tt startpoint} (\bs{p}_c^i)$, $i=1,\cdots,n_c$\;
		\nl $(p_1,I_1) \leftarrow {\tt findmin} (h(i))$, 
		$m \leftarrow 0$ \;
		\While{$m<n_c$}{
			\nl $(p_2,I_2) \leftarrow {\tt findmin} [h(i)/h(I_1)]$ \;
			\nl $ ^s\bs{p}_m \leftarrow {\tt addpoint} (\bs{p}_c(I_1), p_1, p_2+L) $ \;
			\nl \lIf{${\tt isfinish}(I_1)$}{$m \leftarrow m+1$,  $h(I_1) \leftarrow inf$
			}
			\nl		\lElse{ $h(I_1) \leftarrow p_2+L$
			}
			\nl $(p_1,I_1) \leftarrow (p_2, I_2)$\;
		}
		\nl $\{\bs{p}_n\} \leftarrow \{\pi^{-1}_i (^s\bs{p}_i)\}$, $i=1,\cdots,n_c$
	\end{algorithm}
	
As depicted in Fig.~\ref{nozzleplan:b}, according to Algorithm~\ref{alg_nozzle_planning}, the first points for each crack of the first three cracks are $h(1)=s_{1a}$, $h(2)=s_{2a}$, and $h(3)=s_{3a}$. In the algorithm, $p_1=s_{1a}$ and $I_1=1$ since $s_{1a}$ is the minimum point. Then, we find the minimum from $h(2)$ and $h(3)$, so that $p_2=s_{2a}$, $I_2=2$ (line 3). We also obtain $s_{1b}=s_{2a}+L$ (line 4), and therefore, all points on crack 1 from $s_{1a}$ to $s_{1b}$ are added to $^s\bs{p}_1$. The next iteration then adds the points on crack 2 and continues until all $n_c$ cracks are covered. The value of $L$ is chosen to provide a trade-off between the nozzle switching frequency among cracks and the dedicating time duration to a single crack. When $L$ is small, the switching between different cracks becomes frequent, and when $L$ is large, it might miss filling some cracks within $\mathcal{F}$. We consider using $L\leq 2a$ for the switching between cracks within $\mathcal{F}$ to have good performance.

\begin{figure*}[htb!]
	\centering
	\subfigure[]{
		\label{test:a}  
		\includegraphics[width=2.25in]{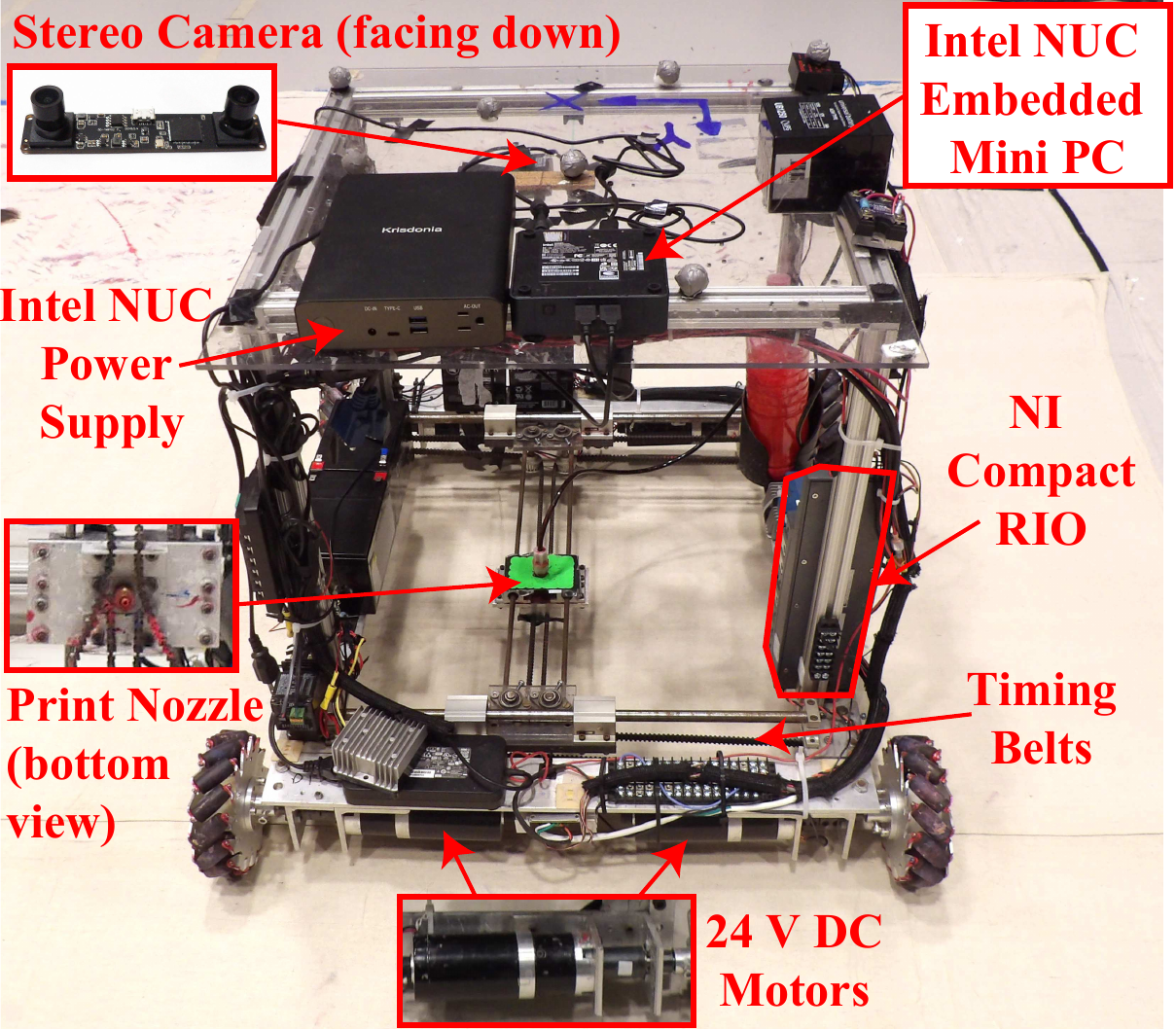}}
	\hspace{0mm}
	\subfigure[]{
		\label{test:c}  
		\includegraphics[width=3.1in]{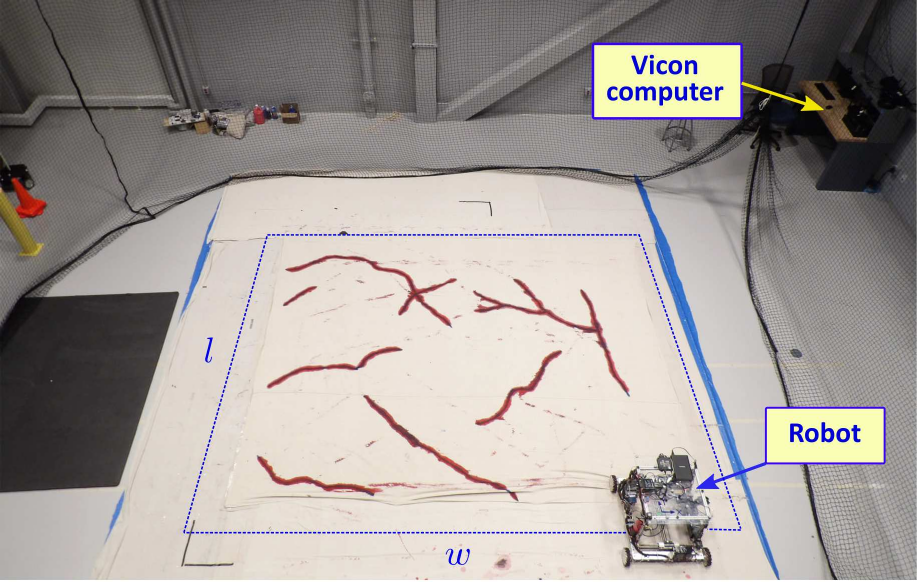}}
	\hspace{0mm}
	\subfigure[]{
		\label{test:d}  
		\includegraphics[width=1.35in]{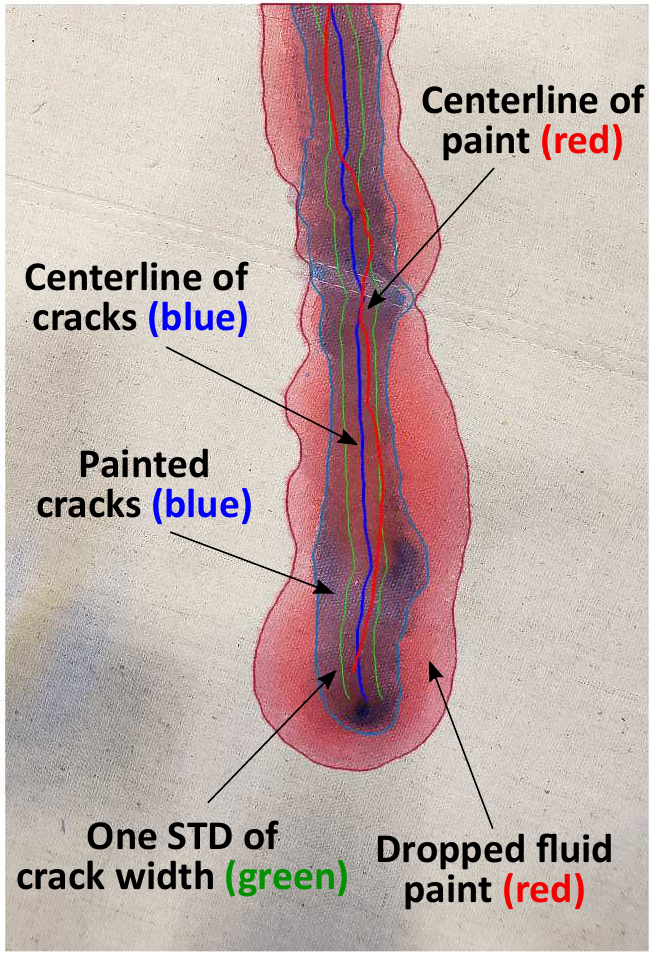}}
	\vspace{-2mm}
	\caption{(a) The omni-directional crack-filling robot with various sensors and actuators. (b) Indoor experimental setup with an optical motion capture system. (c) An illustrative example of the filling error calculation. The crack filling errors are calculated by the center-line differences between the cracks (blue) and the delivered paint (red).}
	\label{test}
	\vspace{-3mm}
\end{figure*}

\subsection{Robot and Nozzle Motion Control}

The robot's orientation is independently controlled with linear velocity due to the use of omni-directional wheels. We first discuss the choice of orientation control. The robot's driving energy expense can be represented as $J_\theta=\frac{1}{2}\sum_{i=1}^4 \omega_i^2$. Taking the derivative of $J_\theta$ with respect to $\theta$, we obtain
\begin{eqnarray}
\frac{\partial J_\theta}{\partial \theta}
&=&\frac{2}{R_w^2} \sum_{i=1}^4
[(\dot{x}_r^2-\dot{y}_r^2)\s_{\theta_i} \c_ {\theta_i}
+ \dot{x}_r \dot{y}_r (\s^2_{\theta_i} - \c^2_ {\theta_i}) \nonumber \\
&&- (\dot{x}_r+\dot{y}_r) R_d \dot{\theta} \s_{\theta_i}
+ (\dot{y}_r-\dot{x}_r) R_d \dot{\theta} \c_{\theta_i}].
\label{eq_partial_theta}
\end{eqnarray}
Using~(\ref{eq_kinematic_model}), it is straightforward to show that $\frac{\partial J_\theta}{\partial \theta}=0$, indicating that the energy expense is independent of the robot's orientation. Similarly, we obtain 
\begin{equation}
\frac{\partial J_\theta}{\partial \dot{\theta}}
=\frac{2 R_d}{R_w^2}
\sum_{i=1}^{4}
\left[-\dot{x}_r \s_{\theta _i}+\dot{y}_r \c_ {\theta _i}+R_d \dot{\theta}\right]
=\frac{8 R_d^2 \dot{\theta}}{R_w^2}.
\label{eq_partial_omega}
\end{equation}
The above result implies that changing the orientation increases the energy cost. Therefore, from~(\ref{eq_partial_theta}) and~(\ref{eq_partial_omega}), we set $\dot{\theta}=0$ to minimize the energy expense $J_\theta$, and in implementation, we further maintain $\theta=0$ as the desired body orientation during robot movement for simplicity.

For the robot and nozzle motion control, the objective is to reach all cracks within $\mathcal{F}$ and complete the crack-filling action while robot $\mathbf{R}$ is in motion. The desired nozzle path is given by Algorithm~\ref{alg_nozzle_planning} in $\mathcal{B}$ as $\{\bs{p}_n\}$, while the desired robot path $\mathcal{P}_R$ is denoted as $\bs{p}_r$ in $\mathcal{N}$. $\mathcal{P}_R$ is computed from the coverage path planning algorithms, specifically Algorithms 3, 4, and 6. Defining $\bs{\xi}=[\bs{q}_r^{\mathsf{T}}\; \bs{q}_n^{\mathsf{T}}]^{\mathsf{T}}$ and $\bs{u}=[\bs{\omega}_r^{\mathsf{T}}\; \bs{\omega}_n^{\mathsf{T}}]^{\mathsf{T}}$, from~(\ref{eq_kinematic_model}), a discrete-time state-space model is used to represent the robot and nozzle motions at the $k$th step
\begin{equation}
\bs{\xi}(k+1)=\bs{\xi}(k)+\bs{B}\bs{u}(k),
	\label{eq_plant}
\end{equation}
where $k \in \mathbb{N}$, $\bs{B}=\Delta T [(\bs{A}^{\mathsf{T}}_r \bs{A}_r)^{-1}\bs{A}_r^{\mathsf{T}} \; \bs{A}_n^{-1}]^{\mathsf{T}}$ and $\Delta T$ is the sampling period. 

The robot velocity is slow compared to the nozzle motion. We denote $\mathcal{X}_r \subset \mathbb{R}^3$ and $\mathcal{X}_n \subset \mathbb{R}^2$ as the allowable robot and filling nozzle velocity sets in $\mathcal{N}$, respectively. We then have the velocity constraints as $\|\bs{v}_r(k)\|_2  \le \|\bs{v}_n(k)\|_2$, where $\bs{v}_r(k)=\dot{\bs{q}}_r(k) \in \mathcal{X}_r$ and $\bs{v}_n(k)=\dot{\bs{q}}_n(k) \in \mathcal{X}_n,$ are the robot velocity and the nozzle relative velocity in $\mathcal{N}$, respectively. Using~(\ref{eq_kinematic_model}), $\bs{v}_r$ and $\bs{v}_n$ are calculated as
\begin{equation}
\bs{v}_r=(\bs{A}^{\mathsf{T}}_r \bs{A}_r)^{-1}\bs{A}_r^{\mathsf{T}} \bs{\omega}_r,\; \bs{v}_n=\begin{bmatrix}
v_{n1} & v_{n2}\end{bmatrix}^{\mathsf{T}}
\label{eq_function_kine}
\end{equation}
where
\begin{eqnarray*}
	v_{n1}&=&\frac{R_g}{2}(\omega_5+\omega_6) \c_\theta - {x_n} \dot{\theta} \s_\theta +\frac{R_g}{2}(\omega_5-\omega_6) \s_\theta-\\
	&&{y_n} \dot{\theta} \c_\theta + \dot{x}_r, \\ v_{n2}&=&\frac{R_g}{2}(\omega_5+\omega_6) \s_\theta + {x_n} \dot{\theta} \c_\theta-\frac{R_g}{2}(\omega_5-\omega_6) \c_\theta-\\
	&&{y_n} \dot{\theta} \s_\theta + \dot{y}_r.
\end{eqnarray*}
Note that~\eqref{eq_function_kine} establishes the relationship between state $\bs{\xi}$ and the control input $\bs{u}$. 

The objective function at the $k$th step is to minimize the position errors and the inputs as 
\begin{equation}
J(k) =\sum_{i=0}^{H} [\bs{e}_{\xi}^{\mathsf{T}}(k+i)\bs{e}_{\xi}(k+i) + \bs{u}(k+i)^{\mathsf{T}}\bs{u}(k+i)],
\label{eq_index}
\end{equation}
where $H \in \mathbb{N}$ is the predictive horizon, error $\bs{e}_{\xi}(j)=\bs{\xi}(j)-\bs{\xi}_d(j)$, $j\in \mathbb{N}$, desired trajectories $\bs{\xi}_d=[\bs{q}^{\mathsf{T}}_{rd} \; \bs{p}_n^{\mathsf{T}}]^{\mathsf{T}}$ and $\bs{q}_{rd}=[\bs{p}^{\mathsf{T}}_r \; 0]^{\mathsf{T}}$. We apply the physical constraints to keep the nozzle motion inside $\mathcal{F}$ and motor speeds below their limits, which leads to the following MPC problem:
\begin{subequations}
	\begin{align}
		\min_{\bs{u}_k} \; & J(k) \\
		\text{subj. to} \; &  \bs{\xi}(k+i+1)=\bs{\xi}(k+i)+\bs{B}\bs{u}(k+i), \\
		& \|\bs{q}_{n} \|_2 \leq a, \; \|\bs{\omega}_{r} \| \leq \omega_r^{\max}, \;\|\bs{\omega}_{n} \| \leq \omega_n^{\max},\\
		& \|\bs{v}_r(k)\|_2  \le \|\bs{v}_n(k)\|_2, \; \|\bs{v}_n(k) \|_2 \leq v_m, 
	\end{align}
	\label{opt}
\end{subequations}
\hspace{-1.6mm}where $i=0,\cdots,H$, $\bs{u}_k=\{\bs{u}(k),\cdots,\bs{u}(k+H)\}$, $\omega_r^{\max}$, $\omega_n^{\max}$, and $v_m$ are the maximum velocity limits of the robot and nozzle driving motors and nozzle motion, respectively. We use YALMIP~\cite{Lofberg2004} to solve the MPC design in~(\ref{opt}). 

\vspace{-0mm}
\section{Experimental Setup and Evaluation Metrics}
\label{experiment}

\subsection{Experimental Setup}

Fig.~\ref{test:a} depicts the crack filling robot prototype, while Fig.~\ref{test:c} showcases the indoor experimental setup. Instead of creating actual cracks on the floor surface, we simulated crack maps on drop cloths using blue paint. This approach allowed us to primarily test and validate the motion planner and robot control design. To emulate the crack-filling action, the robot dispensed red paint to cover the drawn cracks. The paint chosen for the experiments was dense, minimizing dispersion or enlargement after application on the cloth. This setup enables us to assess the proposed motion planners and compare their performance with other benchmark algorithms across various crack characteristics, such as distribution and density, at a relatively low cost.

Using optical markers positioned on the top surface of the robot and the motion capture system (consisting of 8 Vantage cameras, Vicon Ltd.), we captured the robot's position and orientation at a frequency of $100$~Hz. A hydraulic pump and a solenoid valve were used for fluid paint delivery through the nozzle. With the known robot location, the local crack images within the region centered around the robot with a range of $S$ were fed to the planner to emulate the onboard crack detection sensor. The local position of the nozzle in the robot frame was obtained from a stereo camera mounted at the center of the robot pointing downward; see Fig.~\ref{test:a}. In addition to the nozzle position, the stereo camera provided the real-time location of both unfilled and filled cracks within $\mathcal{F}$.

The onboard control implementation consisted of two platforms: the low-level controller was deployed on a real-time embedded system (Compact RIO NI-cRIO-9074, National Instruments Inc.), while the upper-level controller was a portable high-performance microprocessor (Intel NUC7i7DNK, Intel Corp.). The low-level controller primarily handled robot motion control and crack detection and planning. Imaging processing to identify crack positions and stepper motor control for nozzle motion were performed on the upper-level controller. Imaging data collection and motion control were executed at a rate of $10$ Hz. Synchronization between the motion capture system and the onboard computers was achieved through a WiFi wireless connection.

As shown in Fig.~\ref{test:c}, the dimensions of the indoor testing site are $l=5.79$~m and $w=6.10$~m. The physical and model parameters for the robot are as follows: $R_w=7.6$~cm, $R_g=1.3$~cm, $R_d=49$~cm, $S=69$~cm, and $a=8.9$~cm. For the crack-filling planning and MPC design, the parameters are set as follows: $L=3/5 a$, $\Delta=0.1$~s, $H=10$, $\omega_r^{\max}=1.31$~rad/s, $\omega_n^{\max}=7.87$~rad/s, and $v_m=0.1$~m/s. To create crack maps with varying densities and distributions, we used a crack image database from~\cite{shi2016automatic}. The density of a crack represents the Minkowski sum area of each topology (calculated by $\mathcal{T}\oplus S$) over the total workspace area. Random coordinate points and angles were generated using uniform and Gaussian distributions for the location and orientation of each topology. Four sets of crack maps were selected with different crack distributions and densities, that is, uniformly distributed cracks with a 100\% density (denoted as U100) and an 80\% density (U80), and Gaussian distribution cracks with a 100\% density (G100) and a 20\% density (G20).  

\begin{figure*}[htb!]
	\hspace{-2mm}
	\subfigure[]{
		\label{global_robot_path}
		\includegraphics[width=2.05in]{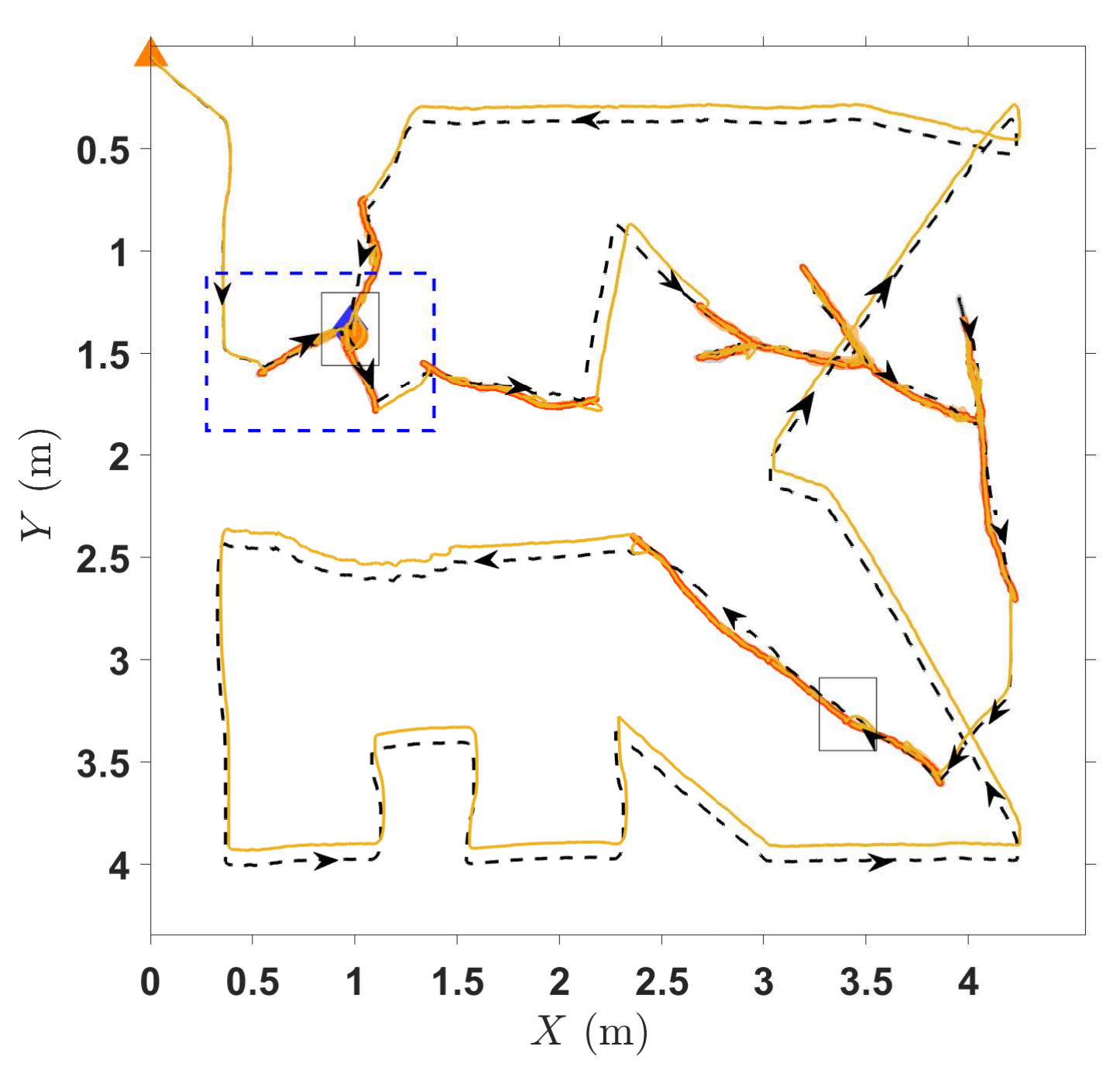}}
	\hspace{-3mm}
	\subfigure[]{
		\label{global_snapshot_path}
		\includegraphics[width=3.4in]{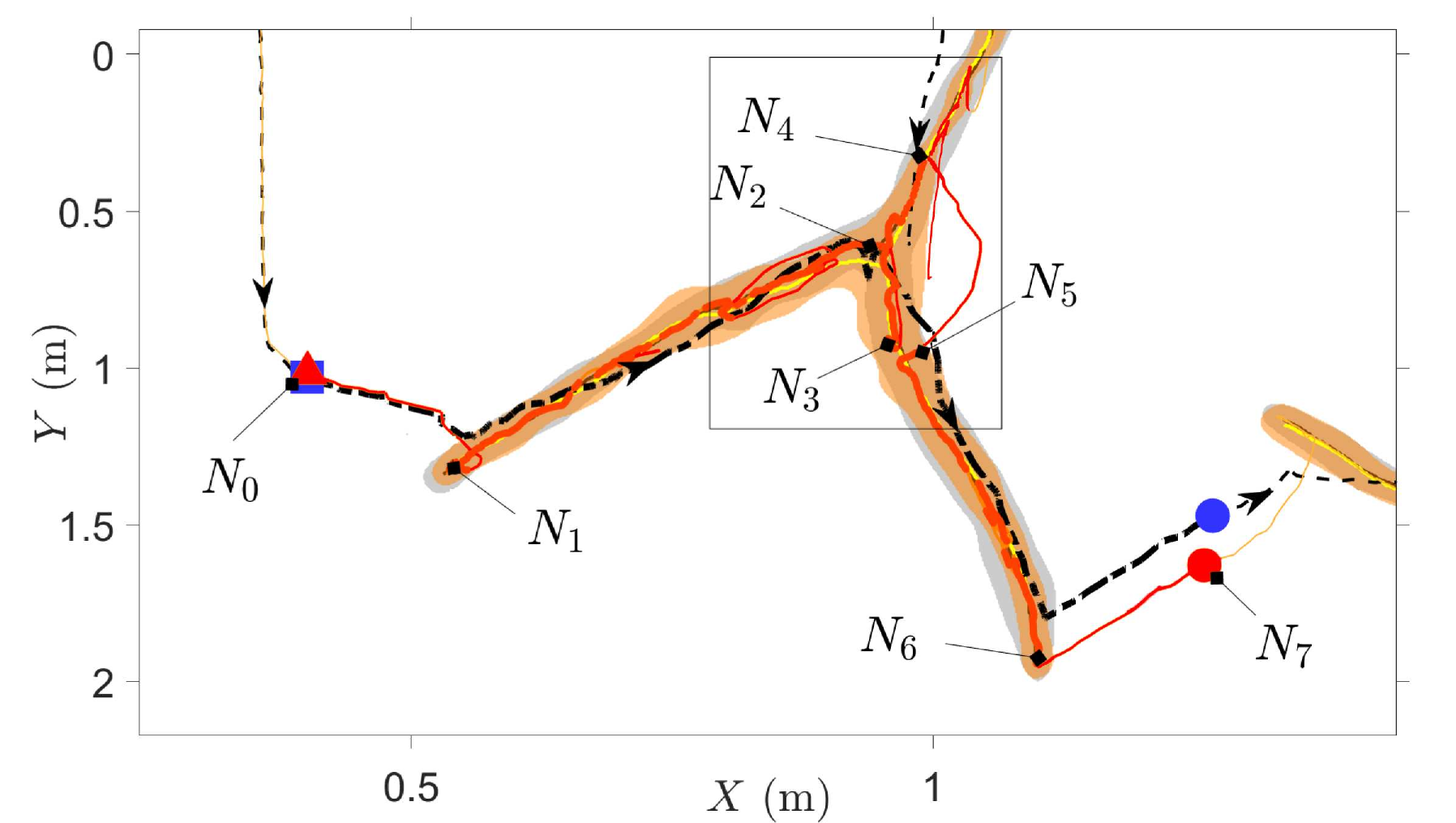}}
	\hspace{-5mm}
	\subfigure[]{
		\label{local_print_image}
		\includegraphics[width=1.73in]{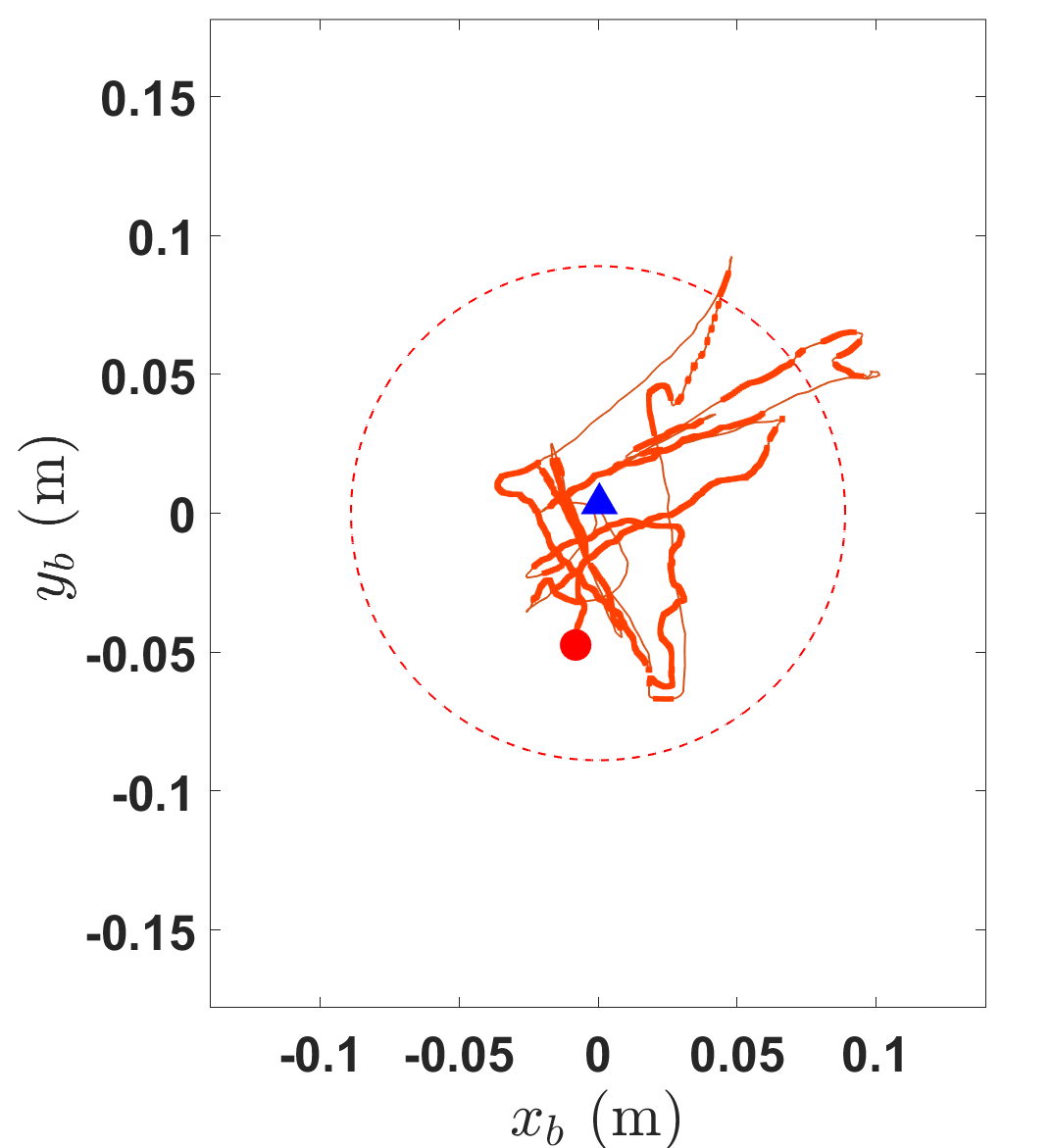}}
	\vspace{-4mm}
	\caption{Experimental results for the robot and nozzle planning and control to fill a crack map that was generated by 80\% density with uniform distribution (U80). (a) Trajectories of the robot and the nozzle. The robot's center path is represented by the dashed black curve, and the nozzle path is marked by the solid orange line. The thick orange lines indicate the painted areas. The arrows dictate the robot's traveling direction. The orange marks ``{\color{orange} $\blacktriangle$}'' and ``{\color{orange} \large $\bullet$}'' indicate the robot's starting and ending locations. (b) A zoomed-in image of the robot and nozzle path for the portion marked in the blue rectangular box in (a). The thick gray and orange regions represent the cracks and painted areas, respectively. The thick and thin red line segments indicate the active (i.e., delivering paint) and inactive filling (i.e., no paint delivery) actions, respectively. The red marks ``{\color{red} $\blacktriangle$}'' and ``{\color{red} \large $\bullet$}'' indicate the nozzle's starting and ending locations for the zoomed-in path, respectively. Similarly, the blue marks ``{\color{blue} $\blacksquare$}'' and ``{\color{blue} \large $\bullet$}'' indicate the robot's starting and ending locations for the zoomed-in path, respectively. (c) A plot of the nozzle trajectory with respect to the robot frame for the highlighted trajectory (from points $N_0$ to $N_7$) in (b). The red dotted circular is region $\mathcal{F}$. The thick and thin line segments indicate the active and inactive filling actions. The blue mark ``{\color{blue} $\blacktriangle$}'' is the starting location of the nozzle, while red the ``{\color{red} \large $\bullet$}'' indicates the nozzle's ending location.}
	\label{fig_algorithm}
	\vspace{-1mm}
\end{figure*}

\vspace{-3mm}
\subsection{Evaluation Metrics}

For comparison purposes, we also implemented two heuristic benchmark coverage planning algorithms. The first algorithm, ${\tt ZigZag}$, solves the complete footprint coverage problem~\cite{LaValle2006} by generating zigzag paths with a slice width of $2a$ (i.e., the diameter of $\mathcal{F}$). The robot then follows this path, filling only the cracks within its footprint area along the trajectory. The ${\tt ZigZag}$ algorithm provides exhaustive coverage using the robot's footprint, but the coverage contains large overlaps. A greedy algorithm, denoted as ${\tt Greedy}$, is used to generate zigzag waypoints that cover the free space using the onboard detection sensor. If any targets are detected, the robot follows and covers them by footprint in the explored slices, then returns to the next unexplored waypoint to continue scanning for the targets. Algorithm~\ref{greedy} illustrates the implementation of the ${\tt Greedy}$ planner.

\begin{algorithm}[ht!]
	\DontPrintSemicolon
	\label{greedy}
	\caption {${\tt Greedy}$}
	\SetAlgoVlined
	\SetKwInOut{Input}{Input}
	\SetKwInOut{Output}{Output}
	\Input{$\mathcal{W}, \mathcal{I}, S$}
	\Output{$\mathcal{P}_R$}
	\nl $\mathcal{P}_R  \gets {\tt  complete\_coverage} (\mathcal{W},S)$\;
	\For{\text{each unexplored waypoint of $\mathcal{P}_R$} $ \rightarrow {P_\text{n}}(i) $}{
		\nl ${\tt follow\_path}( P_\text{n}(i))$, $\mathcal{T} \gets {\tt get\_topology} (\mathcal{I})$\;
		\lIf{crack is found}{follow $\mathcal{T}$}
	}
\end{algorithm}

We used various metrics to assess the performance of motion planning and crack filling: (1) {\em filling time}: the total time for the nozzle to deliver paint for all cracks; (2) {\em robot traveling time}: the overall time taken by the robot to scan and fill the cracks; (3) {\em robot path length} and {\em nozzle path length}: the combined arc lengths of the trajectories traveled by the robot and the nozzle, respectively; (4) {\em sensor coverage}: the percentage of the total sensor-covered area over the entire workspace. Sensor coverage values can exceed 100\%, indicating overlapping coverage ratios; and (5) {\em filling accuracy}: the percentage fraction of the total length of cracks with filling error that is greater than a threshold. As shown in Fig.~\ref{test:d}, we calculated the crack filling error as the difference between the extracted center lines of the cracks (blue line) and the filling paint (red line). The threshold value was taken as the variation in the width of painted marks along the crack. In this study, we calculated and used $5$~mm as the threshold value.

\vspace{-1mm}
\section{Results}
\label{results}

\renewcommand{\arraystretch}{1.3}
\begin{table*}[htb!]
	\begin{center}
		\setlength{\tabcolsep}{0.022in}
		\caption{Experiment performance comparison on four crack maps under five planning algorithms}	
		\label{ExpTable}
		\begin{tabular}{|c|cccc|cccc|cccc|cccc|cccc|cccc|}
			\hline\hline
			& \multicolumn{4}{c|}{Filling time (s)} & \multicolumn{4}{c|}{Robot travel time (s)} & \multicolumn{4}{c|}{Robot path length (m)} & \multicolumn{4}{c|}{Nozzle path (m)} & \multicolumn{4}{c|}{Sensor coverage (\%)} & \multicolumn{4}{c|}{Filling accuracy (\%)} \\ \hline
			Crack dist.  & U100   & U80 & G100 & G20  & U100   & U80 & G100 & G20  & U100   & U80 & G100 & G20  & U100   & U80 & G100 & G20  & U100   & U80 & G100 & G20  & U100   & U80 & G100 & G20\\ \hline
			\multirow{2}{*}{${\tt oSCC}$}  & 731 & {\bf 401} & {\bf 356} & 346  &1398 & {\bf 937}  & 952  & {\bf 889}  & {\bf 51}  & {\bf 42}  & 52 & {\bf 46} & 26 & {\bf 15} & 15 & 14 & {\bf 131} & {\bf 109} & 133 & {\bf 119}  &98.9 & 98.9 &98.1 & 98.1 \\ 
			& & & & & & & & & (50) & (40) & (45) & (44) & & & & & (129) & (103) & (117) & (114) & & & &           \\ \hline
			\multirow{2}{*}{${\tt SCC}$}  & {\bf 654} & 430 & 366 & {\bf 345} & {\bf 1328} & 975& {\bf 919} & {\bf 889} & {\bf 51} &49 & {\bf 49} & 52& {\bf 23} &16 & {\bf 14} & {\bf 12} & 132 & 126 & {\bf 127} & 134 & 99.1 & 99.8 & 98.4 & 98.4  \\
			& & & & & & & & & (48) & (43) &(46) & (45)& & & & & (123) & (111) & (118) & (117) &  &  &  &  \\ \hline
			\multirow{2}{*}{${\tt GCC}$} &559&364& 273 & 276 &914 & 577 & 455 & 455 & 30 & 19 & 15 &13 & 30 & 20 & 14 & 12 & {58} & {34} &24 &{20} & 99.1 & 99.8 & 98.4 & 98.4 \\
			&&& & &&&&& (29) &(17) &(13)&(12) & & & & & (60) &(34) & (24)&(20) & & & & \\ \hline
			\multirow{2}{*}{${\tt Greedy}$} & 746& 446& 414 &372& 1714 &1228 &1104 &1060 &73&63&58&57&29& 19&17&16&187&163&150&147&99.6&99.5&99.1& 99.1 \\ 
			&&&&&&&&&(65)&(53)&(53)&(48)&&&&&(166)&(137)&(137)&(123)&&&&\\ \hline
			\multirow{2}{*}{${\tt ZigZag}$} &752&480&448&415&2611&2284&2149&2166&196&195&195&195&29&19&16&18&504&502&501&501&99.1&99.8&98.4& 98.4 \\ 
			&&&&&&&&&(201)&(201)&(201)&(201)&&&&&(517)&(517)&(517)&(517)&&&&\\ \hline\hline
		\end{tabular}
	\end{center}
	\vspace{-3mm}
\end{table*}

\subsection{Experimental Results}

We first present the experimental results under the ${\tt oSCC}$ planner and nozzle motion control. Fig.~\ref{global_robot_path} shows the robot and nozzle motion trajectory to fill a crack map that was generated by an 80\% density with a uniform distribution (i.e., U80). Under the ${\tt oSCC}$ planner, the robot started from the upper-left corner and covered the entire workspace (as indicated by the black dashed lines). The nozzle trajectory, depicted by the orange lines, effectively filled all the cracks. Fig.~\ref{global_snapshot_path} provides a close look at the robot and nozzle motion trajectories upon encountering multiple cracks within $\mathcal{F}$, offering a zoomed-in view of the area outlined in blue in Fig.~\ref{global_robot_path}. The cracks and filled paint are represented by the gray and orange areas, respectively. At position $N_0$, the nozzle was aligned with the robot's center and proceeded towards position $N_1$ to commence crack filling. It then moved on to complete the segment between points $N_1$ and $N_2$. Upon reaching $N_2$, the nozzle motion planner facilitated a switch between two cracks: it filled the segment between $N_2$ and $N_3$ before transitioning to complete the segment between $N_2$ and $N_4$. Filling ceased at $N_4$, prompting the nozzle to proceed to position $N_5$, where it filled the segment connecting to $N_6$. At $N_6$, the nozzle finalized the filling process and, through the motion of the robot, moved to point $N_7$. Remaining portions of the cracks in Fig.~\ref{global_snapshot_path}, not covered by the nozzle's motion from $N_0$ to $N_7$, were filled during the robot's return trajectory (as illustrated in Fig.\ref{global_robot_path}).

Fig.~\ref{local_print_image} illustrates the nozzle motion trajectory viewed in the robot body frame $\mathcal{B}$ from $N_0$ to $N_7$. Most of the trajectories are contained within $\mathcal{F}$, marked as the red dotted circle. It's worth noting that we consider the inner circle of the rectangular robot as the footprint $\mathcal{F}$ in the planners. The experimental nozzle has the capability to extend beyond $\mathcal{F}$ to fill the cracks. Fig.~\ref{velocity_constaint} illustrates the velocity magnitude profiles for both the robot and nozzle during the experiment shown in Fig.~\ref{fig_algorithm}. The plots include the robot's traveling velocity magnitude represented by the black line, the nozzle's relative velocity magnitude profiles during crack filling (depicted by thick orange dots), and during non-filling periods (shown by orange dashed lines). These results validate the velocity constraint specified in~(\ref{opt}d), which indicates that the nozzle's traveling velocity during filling actions was significantly higher than the robot's travel velocity. Conversely, the robot moved faster when the nozzle was not engaged in filling actions. These findings underscore the collaborative operation between the nozzle planner and the robot motion.

\begin{figure}[t!]
	\hspace{-2mm}
	\includegraphics[width=3.45in]{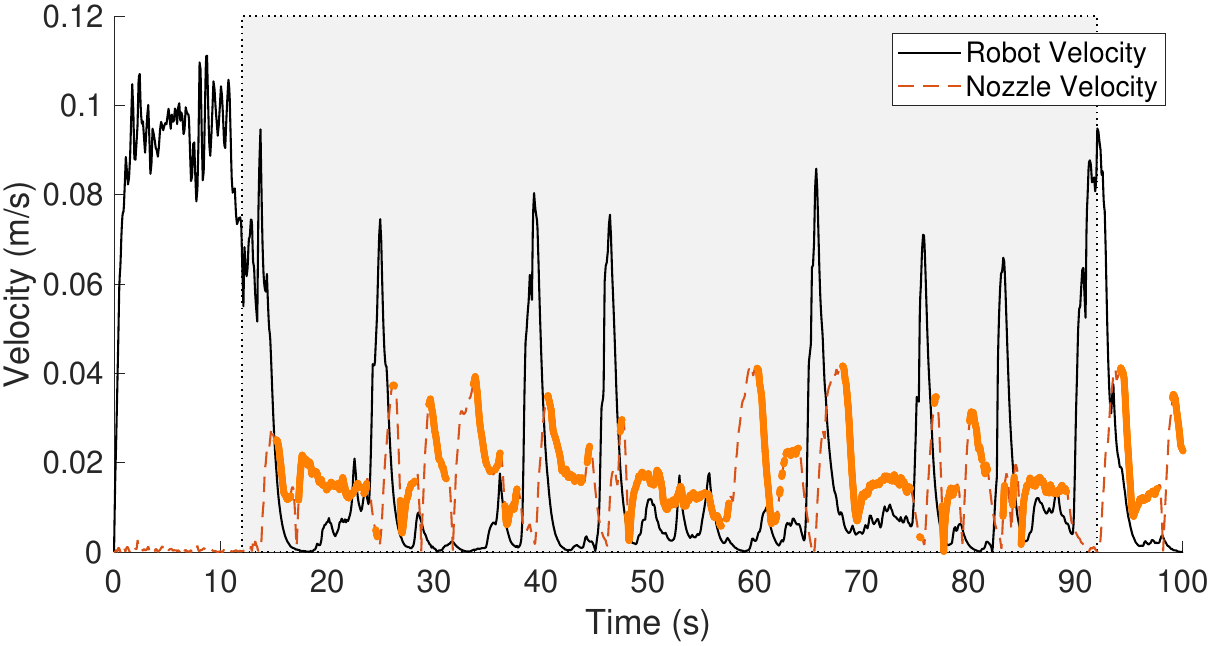}
	\vspace{-2mm}
	\caption{Experimental results for the robot and nozzle velocities in Fig.~\ref{fig_algorithm}. The shaded area shows the active filling region for the portion marked in the blue rectangular box in Fig.~\ref{fig_algorithm}(a). The highlighted orange-thick segments indicate the nozzle speed when filling the cracks, and the thin orange-dash line segments indicate the inactive filling actions.}
	\label{velocity_constaint}
\end{figure}

\begin{figure*}[htb!]
	\hspace{-4mm}
	\subfigure[]{
		\label{oSCCexp}
		\includegraphics[width=1.9in]{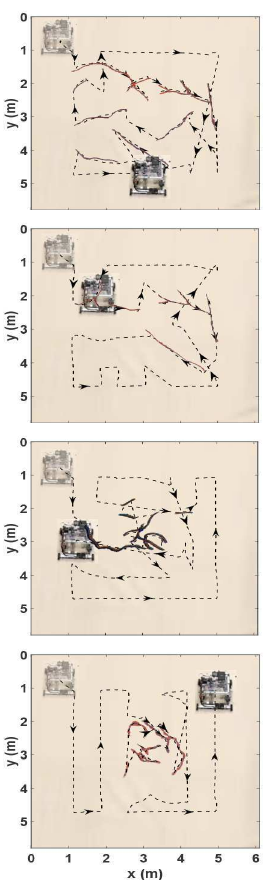}
	}
	\hspace{-7.5mm}
	\subfigure[]{
		\label{SCCexp}
		\includegraphics[width=1.9in]{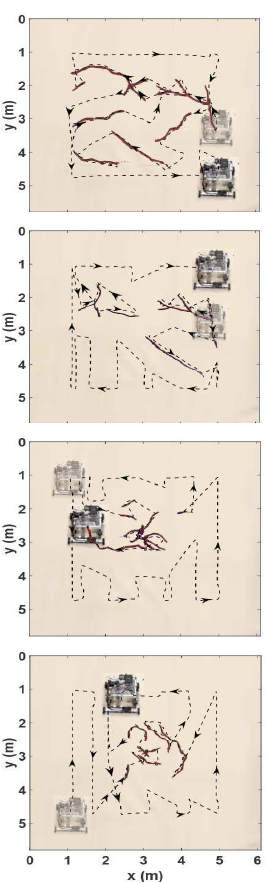}
	}
	\hspace{-7.1mm}
	\subfigure[]{
		\label{Greedyexp}
		\includegraphics[width=1.9in]{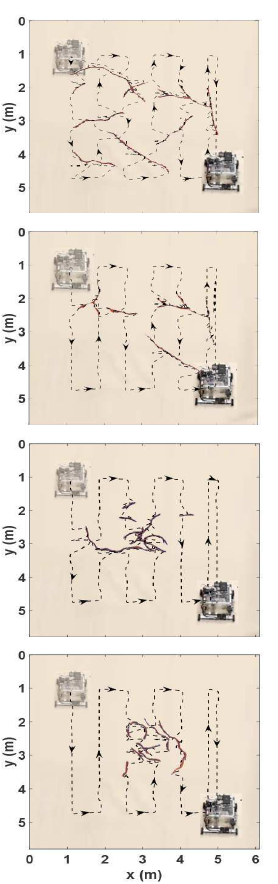}
	}
	\hspace{-7.4mm}
	\subfigure[]{
		\label{Zigzagexp}
		\includegraphics[width=1.9in]{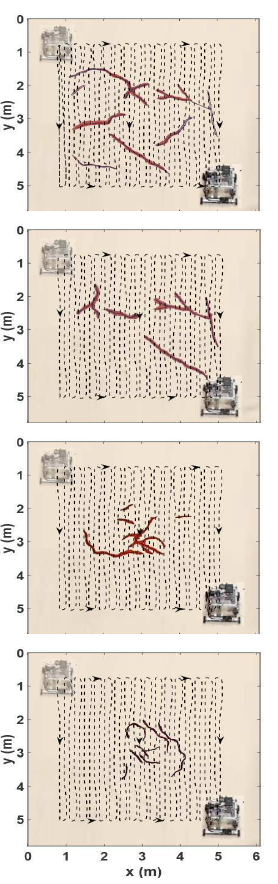}}
	\vspace{-2mm}	
	\caption{The experimental comparison of crack filling outcomes with four crack density and distribution profiles. From the top to the bottom rows, the crack maps are U100, U80, G100, and G20, respectively. Each column represents the experimental results under one motion planning algorithm. The robot's starting and ending locations are marked by a ``shadowed'' and an actual robot image, respectively. (a) Results under the ${\tt oSCC}$ planner, (b) the ${\tt SCC}$ planner, (c) the ${\tt Greedy}$ planner, and (d) the ${\tt ZigZag}$ planner. The blue and red areas represent the cracks and red paint that were dropped by the robot to cover the cracks. The black dashed lines represent the robot center's traveling trajectories, and the arrows indicate the motion directions. More details can be found in the companion video clip.}
	\label{ExpResults}
\end{figure*}

\begin{figure*}[t!]
	\hspace{-2.5mm}
	\subfigure[]{
		\includegraphics[width=1.75in]{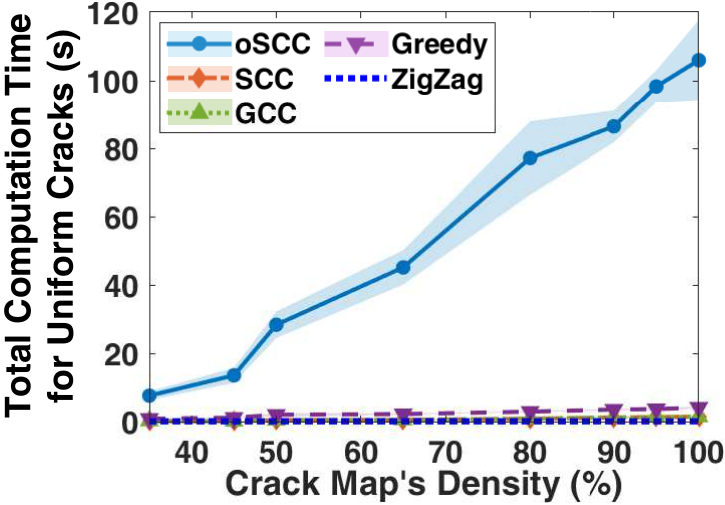}
		\label{CT_comp:a}}
	\hspace{-4mm}
	\subfigure[]{
		\includegraphics[width=1.75in]{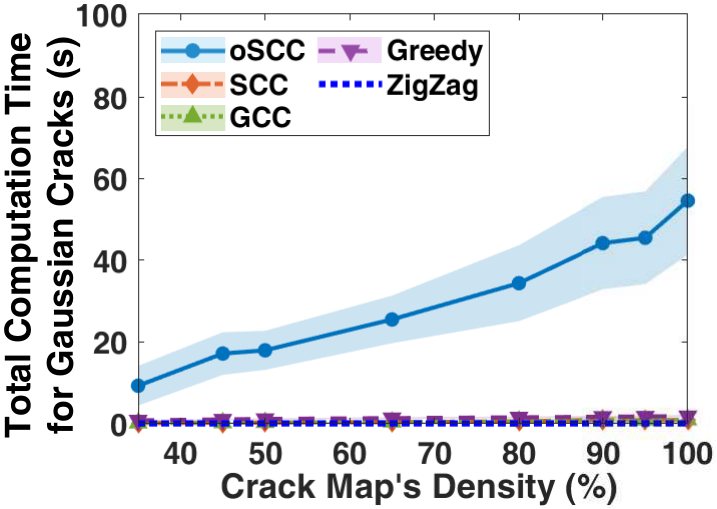}
		\label{CT_comp:b}}
	\hspace{-4mm}
	\subfigure[]{
		\includegraphics[width=1.79in]{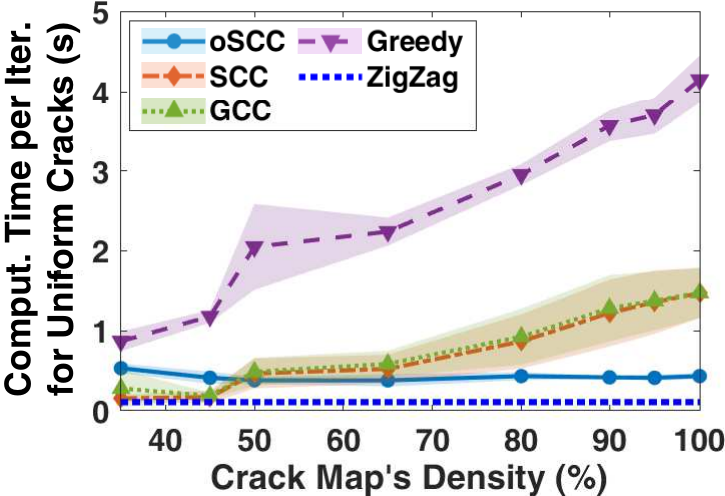}
		\label{CT_comp:c}}
	\hspace{-4mm}
	\subfigure[]{
		\includegraphics[width=1.79in]{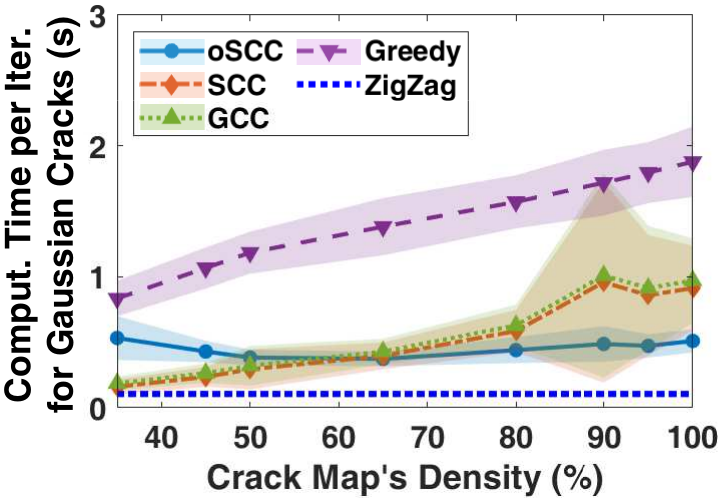}
		\label{CT_comp:d}}
	\vspace{-2mm}
	\caption{Computational time comparisons of the ${\tt oSCC}$, ${\tt SCC}$, ${\tt GCC}$, ${\tt Greedy}$, and ${\tt ZigZag}$ planning algorithms for uniformly distributed cracks and Gaussian-based distributed cracks with different densities in the free space. (a) Total computation time to generate paths for uniformly distributed cracks. (b) Total computation time to generate paths for cracks with Gaussian distributions. (c) Average computation time per iteration for cracks with uniform distribution. (d) Average computation time per iteration for Gaussian-based distributed cracks. The lines represent the mean values, while the shaded areas represent one standard deviation.} 
	\label{CT_comp}
	\vspace{-4mm}
\end{figure*}

We next present the comparison results among the various coverage planning algorithms. The experiments were conducted on four crack patterns: U100, U80, G100, and G20, using four planners: ${\tt oSCC}$, ${\tt SCC}$, ${\tt Greedy}$, and ${\tt ZigZag}$. Fig.~\ref{ExpResults} shows the experimental comparison of actual crack filling outcomes. Readers can further refer to the companion video clip for experimental comparisons under the four algorithms. By experimental results, all of the planners completely covered the free space and filled the cracks. The planned trajectories under ${\tt oSCC}$ and ${\tt SCC}$ were different due to the online feature of the former algorithm. The trajectories under ${\tt Greedy}$ and ${\tt ZigZag}$ shared a similar zigzag scanning pattern, but the latter generated a much denser pattern because of the smaller size of the footprint $\mathcal{F}$ than the sensor coverage range (i.e., $a < S$).

We further conducted image processing to compute the evaluation metrics and compare the results under these planners. Table~\ref{ExpTable} lists the performance comparison. The ${\tt GCC}$ algorithm generates the optimal robot path for the footprint coverage problem. When the sensor range exceeds half the length of the rectangular workspace, the sensor can detect all cracks within the free space in a single scan, effectively reducing the SIFC problem to a footprint coverage problem. Therefore, the path lengths obtained from the ${\tt GCC}$ planner represent the lower bounds on the path length. For comparison purposes, the ${\tt GCC}$ planner was also implemented and included in the table as the benchmark. In Table~\ref{ExpTable}, the bolded values indicate the best performance among all five planners, and the numbers in parentheses show the simulation results under the corresponding planner. We included these simulation results to validate the computational approach. We will present additional computational results later in this section. From the evaluation metrics in Table~\ref{ExpTable}, we observe several facts. First, the ${\tt oSCC}$ and ${\tt SCC}$ outperformed the ${\tt Greedy}$ and ${\tt ZigZag}$ planners in terms of filling time, robot traveling time, robot path length, nozzle path length, and sensor coverage. The filling accuracy values under all planners are similar and close to 100\%. Between ${\tt oSCC}$ and ${\tt SCC}$, the performance is similar in terms of all evaluation metrics, and both are near-optimal compared with the ${\tt GCC}$ optimal planner. The sensor coverage values under all planners except ${\tt GCC}$ are more than 100\%, which indicates the overlapped coverage under most planners. These comparison results confirm the efficiency and effectiveness of the ${\tt oSCC}$ planner in achieving complete and near-optimal sensor and footprint coverage of the free space.

To further evaluate the algorithms, we conducted simulations to assess the statistical performance of the motion planners with different crack densities and distributions. The dimensions of the workspace and all parameters in the simulation setup were kept consistent with the experimental setup. By randomly selecting topologies and setting centroids and the orientation of the crack points and branch angles, five sets of uniformly distributed crack maps were generated. Each set consisted of eight maps with crack densities ranging from $35\%$ to $100\%$, resulting in a total of $5 \times 8=40$ uniformly distributed crack maps. Similarly, six crack maps per density were created by generating coordinate points with a Gaussian distribution. The center of the map was represented by the mean of the distribution, and the area in percentage was the standard deviation. To achieve different spreads of the cracks, the standard deviations were set to $\frac{1}{12}$, $\frac{1}{6}$, and $\frac{1}{3}$ of the total workspace areas. This generated a total of $6\times 3\times 8=144$ Gaussian-distributed crack maps. Therefore, a total of 184 crack maps were generated and used for testing.

We simulated and computed the performance of the ${\tt oSCC}$, ${\tt SCC}$, ${\tt GCC}$, ${\tt Greedy}$, and ${\tt ZigZag}$ planners with respect to different crack distributions (uniform and Gaussian) and densities. Figs.~\ref{CT_comp:a} and~\ref{CT_comp:b} show the total computation time comparison for the uniform and Gaussian distributions, respectively. The ${\tt oSCC}$ planner scanned and computed paths on an iterative basis, while others computed paths over the entire workspace. In each iteration of the ${\tt oSCC}$, the robot scanned the area, extracted the crack graph, and computed the shortest path. The computation time of the ${\tt oSCC}$ depended on the crack density in the workspace. From the figures, the total computation time of the ${\tt oSCC}$ planner increases linearly with crack density. The computation time difference between ${\tt oSCC}$ and ${\tt SCC}$ reflects the iterative nature of the ${\tt oSCC}$ algorithm. In terms of computation time per iteration, the ${\tt oSCC}$ matches the range of the ${\tt SCC}$ planner, as shown in Figs.~\ref{CT_comp:c} and~\ref{CT_comp:d}. The ${\tt oSCC}$ has a much lower computation time per iteration than that of the ${\tt Greedy}$ planner and outperforms both the ${\tt SCC}$ and the ${\tt GCC}$ at high crack densities.

We next compared the robot path lengths and sensor coverage against crack density. Fig.~\ref{SimuResult} illustrates the comparison of robot path length and sensor coverage for uniformly and Gaussian distributed cracks with varying densities. The path lengths under the ${\tt GCC}$ and ${\tt ZigZag}$ planners represent the lower and upper bounds, respectively. The statistical comparison confirms that the proposed ${\tt oSCC}$ achieves similar performance as the ${\tt SCC}$ algorithm, and both outperform the ${\tt Greedy}$ and ${\tt ZigZag}$ benchmarks. Regarding sensor coverage values, the ${\tt oSCC}$, ${\tt SCC}$, ${\tt Greedy}$, and ${\tt ZigZag}$ planners achieve $100\%$ workspace coverage for all the maps. The values greater than 100\% measure the overlapped sensor coverage against the entire free space. Sensor overlapping shares the same trend as the path length comparison for ${\tt oSCC}$, ${\tt SCC}$, ${\tt Greedy}$, and ${\tt ZigZag}$. The overlapping area and robot path length under the ${\tt Greedy}$ algorithm are always larger than those by the ${\tt oSCC}$. Compared to the ${\tt Greedy}$ algorithm, the ${\tt oSCC}$ planner reduces the sensor overlap by up to $62\%$ and shortens the robot's path by up to $24\%$ for densely and scatteredly distributed cracks. For uniformly distributed cracks, the ${\tt oSCC}$ and ${\tt SCC}$ planners achieve similar path lengths and sensor range coverage. In the case of Gaussian-based distributed cracks, the ${\tt oSCC}$ planner results in less sensor overlap and shorter path length than ${\tt SCC}$. Therefore, the ${\tt oSCC}$ outperforms the other planners in covering and scanning the cracks under both distributions.

\begin{figure}[t!]
	\vspace{2mm}
	\hspace{-3mm}
	\subfigure[]{
		\includegraphics[width=1.735in]{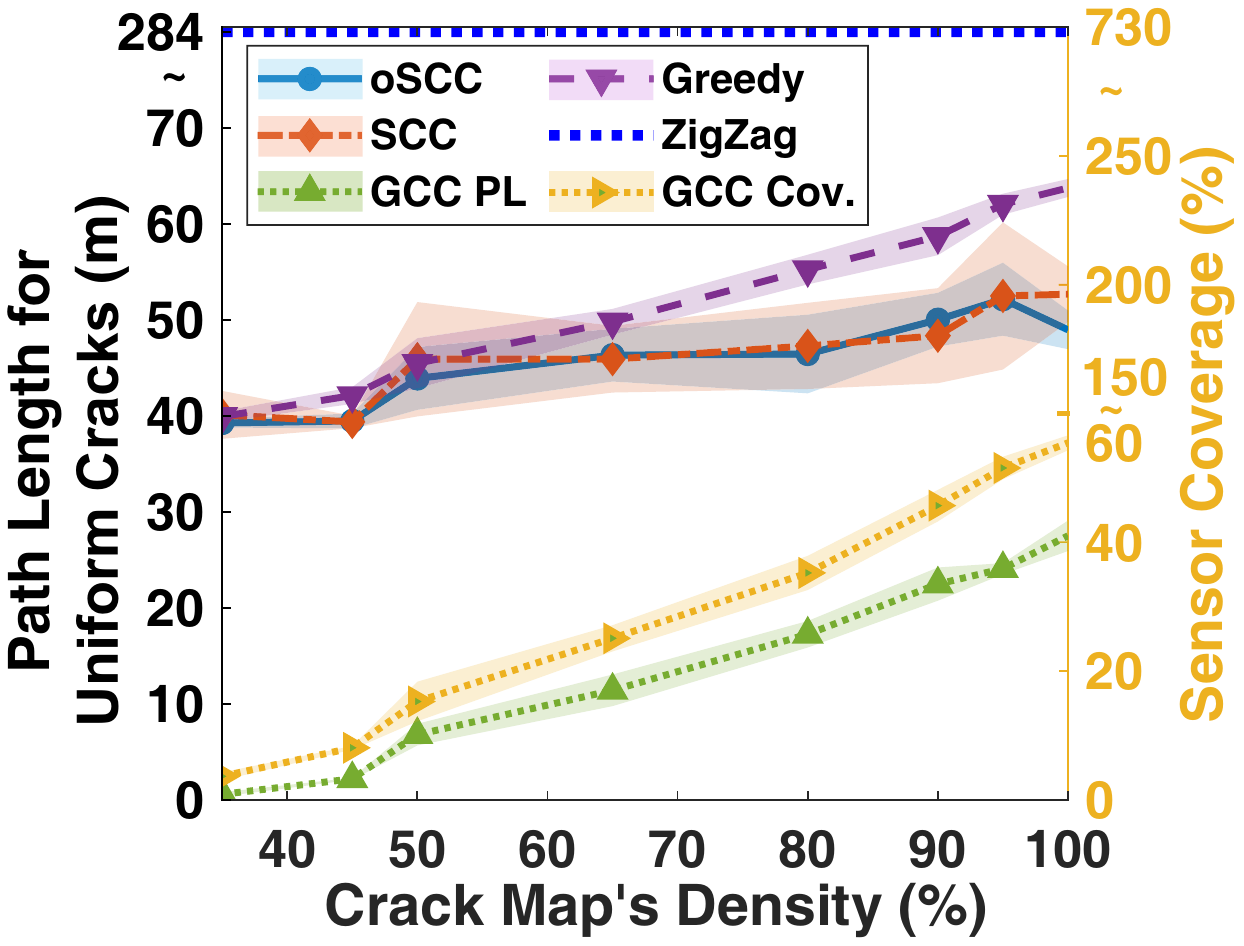}
		\label{SimuResult:c}}
	\hspace{-4mm}
	\subfigure[]{
		\includegraphics[width=1.735in]{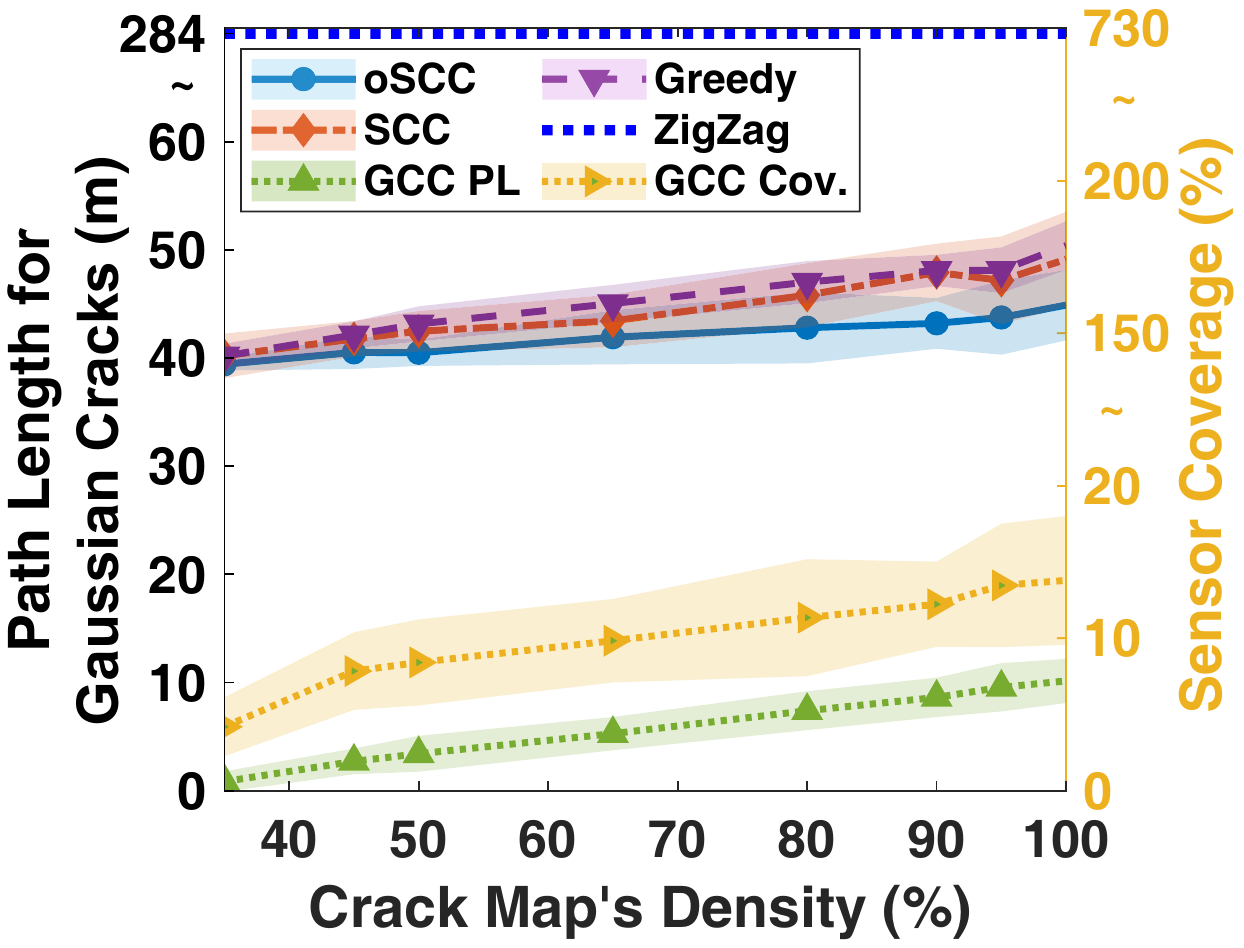}
		\label{SimuResult:d}}
	\vspace{-2mm}
	\caption{Robot path length and sensor coverage comparison under the ${\tt oSCC}$, ${\tt SCC}$, ${\tt GCC}$, ${\tt Greedy}$, and ${\tt ZigZag}$ planners for (a) uniformly distributed cracks and (b) Gaussian distributed cracks with different densities. The left and right $y$-axis labels represent the robot path length and sensor coverage, respectively. The orange and green dotted lines show the sensor coverage and path length for ${\tt GCC}$, respectively. All planners except ${\tt GCC}$ achieve 100\% sensor coverage. The lines represent the mean values, and the shaded areas indicate one standard deviation.} 
	\label{SimuResult}
	\vspace{-3mm}
\end{figure}

\vspace{-2mm}
\subsection{Discussion}

To further explain the sensor overlapping results under the ${\tt oSCC}$ and ${\tt SCC}$ planners shown in Fig.~\ref{SimuResult}, we look into the underlying differences in the planning algorithms. Overlapping occurs when adding minimum-cost connecting edges to create an Eulerian graph, specifically when adding the minimum-cost path between the exit of one cell and the entry of the subsequent cell. It can also be caused by traversing an edge in the crack graph $\mathbb{G}_\text{c}$ more than once to achieve an optimal Eulerian path. All connecting edges contribute to the overlapped sensing area and the increase in path length. The ${\tt oSCC}$ planner results in less sensor overlap and shorter path length than the ${\tt SCC}$ because ${\tt oSCC}$ reduces repeated connecting edges in the online planning process, especially when cracks are clustered. In addition, comparing Figs.~\ref{SimuResult:c} with \ref{SimuResult:d}, the sensor overlapping and the path length of the uniformly distributed cracks are larger than those of the Gaussian distribution for both the ${\tt oSCC}$ and ${\tt SCC}$ planners. These observations confirm that the connecting edges are the main contributor to sensor overlapping because the uniform distribution of cracks results in their widespread scattering, which subsequently leads to an increase in the length of the connecting edges during the LP matching process in the algorithm.

\begin{figure}[t!]
	\hspace*{-3mm}
	\subfigure[]{
		\includegraphics[width=3.5in]{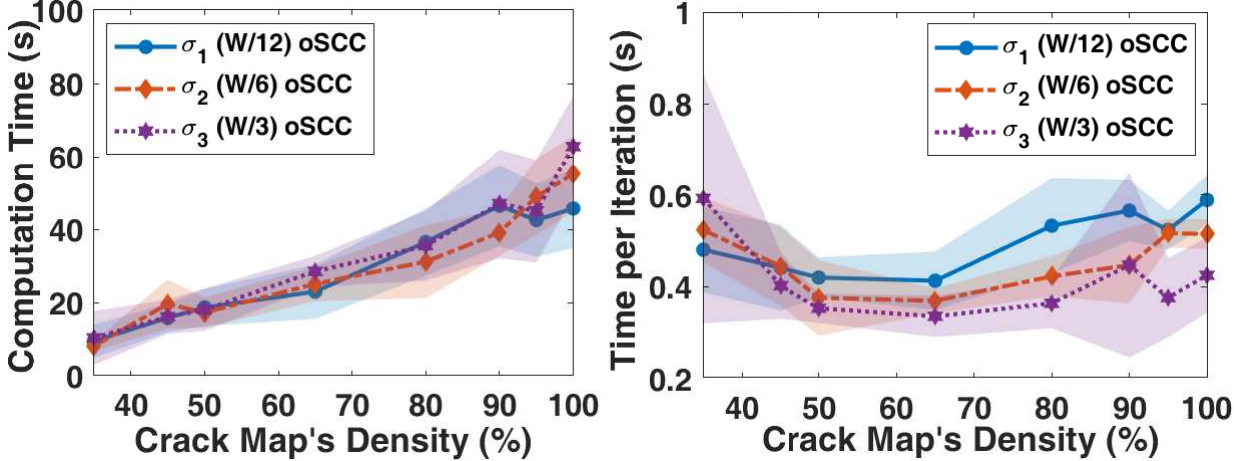}
		\label{Gau_Sig:a}}
	\hspace*{-3mm}
	\subfigure[]{
		\includegraphics[width=3.5in]{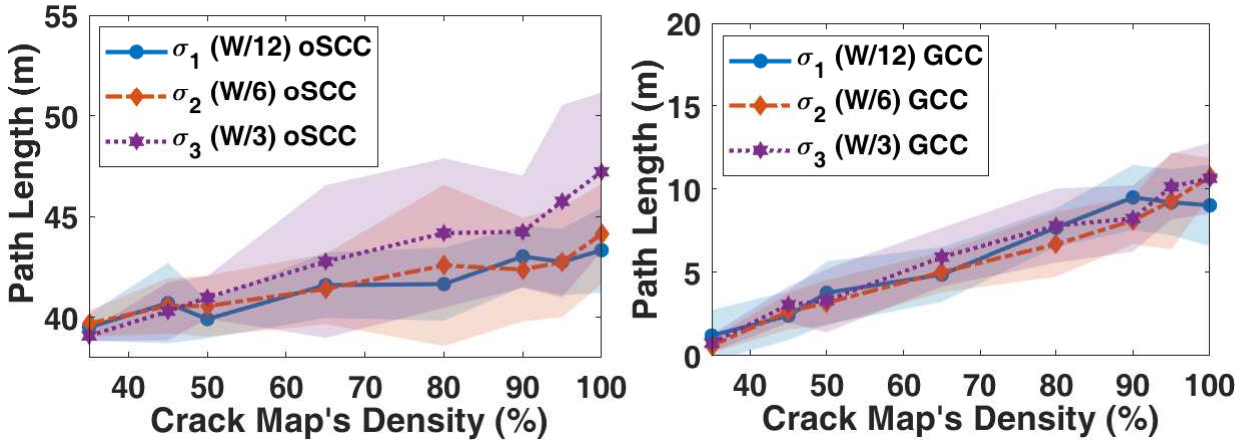}
		\label{Gau_Sig:c}}
	\vspace{-2mm}		
	\caption{Comparisons among the Gaussian maps varied by three dispersities. (a) Computation time (left) and average computation time per iteration (right) of the ${\tt oSCC}$ planner. (b) Robot path length comparison between the ${\tt oSCC}$ (left) and ${\tt GCC}$ (right) planners. The lines represent the mean values, while the shaded areas represent one standard deviation. }  
	\label{Gau_Sig}
	\vspace{-3mm}
\end{figure}

We further conducted a study to understand how crack dispersity impacts the ${\tt oSCC}$ planner's performance. We used Gaussian distribution crack maps to illustrate the results. The crack dispersity is defined by the normal distribution of the crack centroid locations. As explained previously, three standard variations of the total workspace areas were used, namely, $\sigma_1=\frac{1}{12}$, $\sigma_2=\frac{1}{6}$, and $\sigma_3=\frac{1}{3}$. Fig.~\ref{Gau_Sig} shows the comparison of the computation time and robot path length under the ${\tt oSCC}$ and ${\tt GCC}$ planners with varying crack densities and three dispersities. We chose the ${\tt GCC}$ planner as the benchmark for the reason discussed above. Fig.~\ref{Gau_Sig:a} displays the computation time (total and per iteration) of the ${\tt oSCC}$ planner. The results exhibit a similar trend for different crack dispersities. Fig.~\ref{Gau_Sig:c} presents a comparison of robot path length under three different levels of crack dispersity for both the ${\tt oSCC}$ and ${\tt GCC}$ planners. It is observed that the ${\tt oSCC}$ planner results in a reduction of robot path length when cracks are cluttered (i.e., for small values of $\sigma$). In contrast, when the crack maps are known a priori and the entire free space is not covered, the ${\tt GCC}$ planner does not exhibit a significant difference in robot path length. These observations align with the previously discussed relationship between crack dispersity and connecting edge length (i.e., the more scattered the cracks are, the longer the connecting edges), as well as the role of connecting edges between substantial cells in contributing to overlapping. The ${\tt GCC}$ planner only adds connecting edges between cracks and does not result in noticeable differences in path length for varying levels of crack dispersity. Therefore, connecting edges between substantial cells play a major role in increasing path length and overlapping.
While the robot's ability to precisely optimize the path between adjacent cells is limited due to unknown cell dimensions, extensive experimentation and comparison with ${\tt GCC}$ and ${\tt SCC}$ planners show that the final path of the ${\tt oSCC}$ planner achieves a high level of efficiency in terms of travel distance. Statistical simulations and extensive experimental results in Table I and Figs. 14-16 confirm this efficiency.

The robot's sensor range $S$ and footprint radius $a$ might vary, and those two parameters affect the Euler tour generation. We conducted simulations to analyze the robot's traveling distance under various ranges of $S$ and $a$ values. In experiments, the ratio between $S$ and $a$ is $S/a=7.7$, and the ratio of the workspace width $w$ to $S$ is $w/S=8.8$. Because the footprint size mostly influences the local path planner in the given sensor range, we therefore study the effect of varying $S$ when $w$ and $a$ are fixed. Fig.~\ref{SWA} shows the comparison of the robot path length over various $S/a$ ratios under the ${\tt oSCC}$, ${\tt SCC}$, ${\tt GCC}$, ${\tt Greedy}$, and ${\tt ZigZag}$ planners for uniformly and Gaussian distributed crack maps. The path lengths under the ${\tt GCC}$ and ${\tt ZigZag}$ planners are the lower and upper bounds, respectively. When $S/a=1$, the robot must scan the entire workspace with its footprint. Then the problem degrades into the full coverage problem, and the final path is the same as that under the ${\tt ZigZag}$ planner. When $w/S=2$, the sensor detects all the cracks in the free space in one scan. The path lengths of ${\tt oSCC}$ and ${\tt SCC}$ converge to the results under the ${\tt GCC}$ planner. From the results in the figures, the ${\tt oSCC}$ planner outperforms the ${\tt Greedy}$ planner with different sensor ranges, particularly at large sensor ranges. When $S/a<7$ (or $w/S>8$), the ${\tt oSCC}$ achieves a shorter path length than ${\tt SCC}$ for uniform cracks; see Fig.~\ref{SWA:a}. The results imply that the proposed ${\tt oSCC}$ planner yields shorter paths as the sensor range increases. This comparison provides valuable insights for the appropriate selection of planning algorithms for different applications.

\begin{figure}[t!]
	\hspace*{-3mm}
	\subfigure[]{
		\includegraphics[width=1.73in]{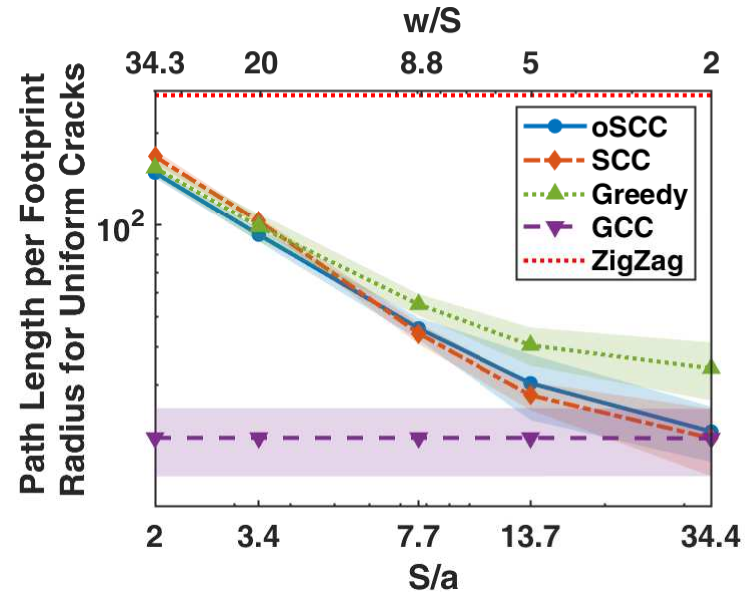}
		\label{SWA:a}}
	\hspace*{-3.7mm}
	\subfigure[]{
		\includegraphics[width=1.73in]{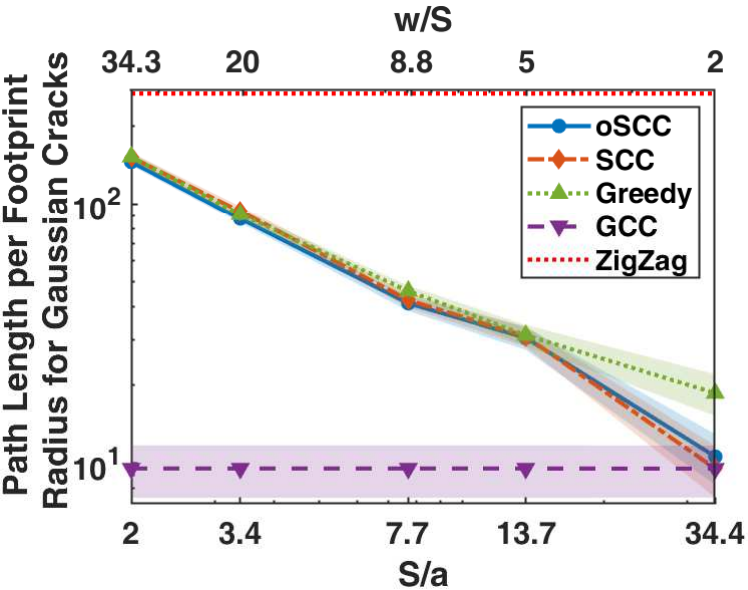}
		\label{SWA:b}}
	\vspace{-2mm}		
	\caption{Robot path length with various crack detection sensor radii ($S$) under the ${\tt oSCC}$, ${\tt SCC}$, ${\tt GCC}$, ${\tt Greedy}$, and ${\tt ZigZag}$ planners for (a) uniformly distributed cracks and (b) Gaussian-based distributed cracks. The workspace width $w=6.1$~m and the footprint radius $a=8.9$~cm are fixed, and the lines and shaded areas represent the mean and one standard deviation, respectively.} 
	\label{SWA}
	\vspace{-3mm}
\end{figure}

In this paper, we assume a known free space where the robot can move freely in any direction without obstacles. We used classic cell decomposition methods, specifically MCD, for both coverage tasks. The work in~\cite{acar2002morse,acar2002sensor} demonstrates the applicability of MCD to workspaces with various shapes and a finite number of obstacles, covering both known and unknown environments. Although obstacle handling is not the main focus of this paper, we acknowledge that in real-world situations, robot sensors can detect obstacles, and these obstacles can be geometrically represented in the environment using polygonal approximations. MCD can then incorporate the detected obstacles into the determination of cell boundaries to ensure that cells do not overlap with obstacles or cross their boundaries. 
The robot's motion planning algorithm, such as the ${\tt oSCC}$ planner, can generate a path to navigate through the cracks and cells while avoiding obstacles, ensuring complete coverage of the workspace. The work in~\cite{acar2002morse} extended MCD into three dimensions (3D), enabling coverage of closed, orientable, and connected surfaces in 3D. While we showcase results in a known rectangular free space, the method's adaptability permits its extension to free spaces with various shapes and even to uneven or vertical 3D surfaces, provided the number of obstacles remains finite. This characteristic renders the method suitable for a wide range of real-world scenarios.

Several limitations of this work can be further explored and improved in the future. The planning algorithms assumed that the nozzle motion was fast enough such that all targets within the robot footprint could be covered in time. As a result, minimizing the robot's traveling distance was considered as the objective, and this simplified treatment decoupled the planning of robot motion from nozzle movement. The experiments were conducted in a simulated indoor environment with drop cloths and paint to create cracks and filling action, as the primary focus of the work was on motion planning algorithm development. The performance metrics were estimated using the paint width, and even though the selected cloth did not absorb paint heavily, the results were not perfectly accurate compared to real crack-filling. We only used one mobile robot to conduct sensing and footprint coverage in this study, and it would be interesting to extend the SIFC problem with multiple collaborative robots to increase efficiency. 
In this work, we made assumptions about known robot locations and constant crack width to emphasize the core planning concept and motion planning and control algorithmic development. For future work, our focus will be on addressing uncertainties in localization by employing advanced techniques in simultaneous localization and mapping and achieving real-time estimation of the crack characteristics. Additionally, we plan to integrate crack width considerations into the planning algorithm's cost and explore adaptive control algorithms to adjust the robot's footprint coverage strategy based on encountered crack characteristics. These enhancements will significantly improve the applicability of our approach, allowing it to handle further realistic scenarios.

\section{Conclusion}
\label{sec_concl}


We have presented a motion planning and control design for simultaneous robotic sensor-based inspection and footprint coverage, with applications to crack detection and repair in civil infrastructure. To address the challenging task of simultaneously performing two complete coverage tasks in SIFC, we first proposed a graph-based target coverage algorithm for the mobile robot. Subsequently, we introduced a novel algorithm to solve the SIFC problem with unknown target information. This algorithm ensured the complete sensor scan of the workspace and the full footprint coverage of all targets, with near-optimal performance in terms of traveling distance. With the planned robot trajectory, the nozzle motion was coordinated to efficiently fill all cracks underneath the robot footprint. Extensive experimental results confirmed the effectiveness of the proposed motion planning and control algorithms under various target distributions. Furthermore, we discussed and demonstrated comparisons with other benchmark planning algorithms. The presented near-optimal and complete coverage planning algorithm has the potential to be used to other robotic SIFC applications.
\begin{appendices}
\vspace{-0mm}

%
%

\section{Sketch proof of Lemma~\ref{lem1}}
\label{proof1}

According to the assumption of MCD, no two critical points change the slice connectivity at the same time. Thus, critical points collinear with the slice direction require special consideration. Let the endpoint node be a node with only one connected edge. When using a horizontal sweep direction, vertical crack edges connecting with endpoint nodes are treated as cell boundaries, and those endpoint nodes are considered critical points, denoted as vertical critical points. For example, as illustrated in Fig.~\ref{CriticalPoint}, $N_1$, $N_2$, and $N_3$ represent vertical critical points. Since the vertical edge of the crack graph divides one cell into two adjacent cells, vertical critical points have two edges in the Reeb graph. One edge originates from the crack graph, and the other from one of the two adjacent cells. This extension ensures that the slice connectivity remains constant within each cell.
	
\begin{figure}[htb!]
		\centering
		\subfigure[]{
			\includegraphics[height=0.93in]{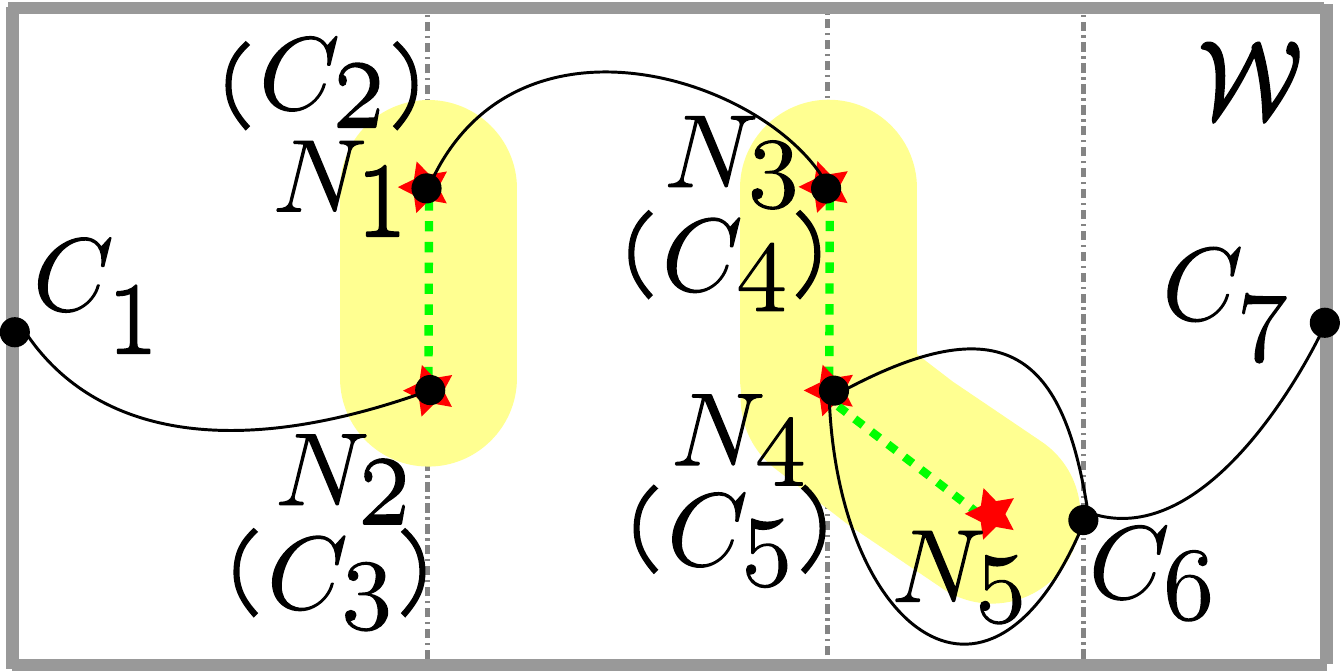}
			\label{CriticalPoint}}
		\subfigure[]{
			\includegraphics[height=0.93in]{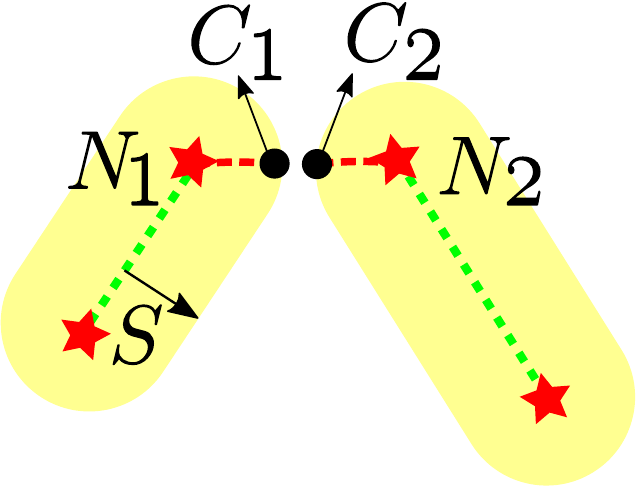}
			\label{closePoint}}
		\vspace{-2mm}
		\caption{ (a) Special consideration is given to the vertical critical points, i.e., the endpoint nodes $N_1$, $N_2$, and $N_3$. (b) If the distance between two critical points $C_1$ and $C_2$ is less than $S$, then the distance between the corresponding nodes $N_1$ and $N_2$ in the crack graph is greater than $2S$. }
		\vspace{-0mm}
\end{figure}
	
Except for the vertical critical points, the remaining critical points of the target region are generated by the surface normal perpendicular to the sweep direction. Note that the target region is the ``dilated'' crack graph by a circular disk with a radius of $S$. If the boundary of the target region is convex, we always find a node in the crack graph within an $S$-distance of the critical point. As shown in Fig.~\ref{GraphConf:a}, critical points $C_2$, $C_4$, and $C_5$ correspond to nodes $N_1$, $N_5$, and $N_4$ in the crack graph, respectively. If the boundary of the target region is concave, then the corresponding node in the crack graph is the intersection point of edges (intersection node), e.g., critical point $C_3$ is associated with node $N_3$ in Fig.~\ref{GraphConf:a}. Thus, every critical point on the boundary of the target region is associated with one node in the crack graph. This proves the lemma.

\vspace{-2mm}
\section{Sketch proof of Proposition~\ref{lem2}}
\label{proof2}

Because of the property of MCD, all the critical points that are generated by the convex boundaries (i.e., convex critical points) are connected to three cells. Similarly, all the critical points that are generated by the concave boundaries (i.e., concave critical points) are connected to one cell. In the Reeb graph, the convex and concave critical points correspond to nodes of degree three and one, respectively. Let us denote the convex and concave critical points of the target region as $C_\text{vex}$ and $C_\text{cav}$ and their corresponding crack nodes as $N_\text{vex}$ and $N_\text{cav}$, respectively, by Lemma~\ref{lem1}.

If $N_\text{vex}$ is an endpoint node, then it has degree one. If $N_\text{vex}$ is an intersection node, then its degree plus the number of concave critical points associated with $N_\text{vex}$ is odd. To form the Euler tour, the nodes with odd degrees need to be paired up with the least cost. Notice that the distance between $C_\text{vex}$ and $N_\text{vex}$ is the sensing range $S$. Because the distance between other nodes in the crack graph and $C_\text{vex}$ is greater or equal to $S$, connecting each pair of $C_\text{vex}$ to $N_\text{vex}$ results in parts of the minimum-cost Euler tour. For the case where the distance between two critical points $C_1$ and $C_2$ is less than $S$, as shown in Fig.~\ref{closePoint}, the distance between the corresponding nodes $N_1$ and $N_2$ in the crack graph is greater than $2S$. Therefore, the optimal approach is to combine the $C_\text{vex}$ (node of degree three) with their corresponding $N_\text{vex}$ (odd degree node) in the Euler tour. These combined nodes have an even degree, guaranteeing that the robot is not stuck at such nodes.

For $C_\text{cav}$, $N_\text{cav}$ must be an intersection node. The parity of the intersection node of the crack graph is the same as the parity of the number of critical points associated with it. Therefore, pairing up these odd nodes with the least cost ensures the optimal Euler tour. The vertical critical points are defined on the graph nodes (by Lemma~\ref{lem1}) and have a degree of two. The vertical critical points do not affect the optimality of the Euler tour. Thus, to find the minimum cost of the Euler tour, the edge of the crack graph never gets doubled, as all the critical points of the target region result in even degrees by connecting them to their corresponding nodes in the crack graph. To pair up other odd nodes in the Reeb graph, only edges corresponding to the cells (free space) get doubled. Doubling the selected edges means splitting the corresponding cells into two portions, which does not increase the cost of covering the whole area. Thus, as all parts of the free space and crack graph are covered exactly once, optimality is preserved. This proves the proposition.

\vspace{-2mm}
\section{Sketch proof of Proposition~\ref{lem3}}
\label{proof3}
The completeness of the ${\tt oSCC}$ planner follows directly from the properties of the Euler tour used to solve the route. By definition, the construction of $\mathbb{G}_\text{w}$ does not stop until all the areas in $\mathcal{W}$ are covered. The Reeb graph $\mathbb{G}_\text{w}$ provides a complete representation of the free space. Because each edge of the graph is traversed (i.e., each cell is covered), it guarantees that all available free spaces are covered. Therefore, the algorithm is complete.

The ${\tt oSCC}$ planner reinforces the connections of the critical points to their corresponding nodes in $\mathbb{G}_\text{c}$ until reaching its end. The resulting least-cost Euler tour, obtained from the doubling of selected edges, establishes an order in which the cells of the Reeb graph should be visited. The Euler tour ensures that no area is covered twice by not traversing any edge twice. Note that the covered spaces are removed from $\mathcal{W}$, and the cell coverage does not duplicate any covered area. Doubling the edges of cells does not increase any cost because the two cell components do not overlap. By definition, no edge of the Euler tour is traversed twice, and this implies no area is covered twice. 
Therefore, the algorithm ensures no redundancy in terms of individual cell coverage. All free space is covered exactly once from the least-cost Euler tour. This results in the most efficient path that minimizes the robot's travel distance to locally cover all the individual cells when the connections of each covered cell are not considered.

\end{appendices}

\bibliography{KaiyanYuRef,YiRefrev,Kaiyan_planning}

\begin{thebibliography}{10}
\providecommand{\url}[1]{#1}
\csname url@samestyle\endcsname
\providecommand{\newblock}{\relax}
\providecommand{\bibinfo}[2]{#2}
\providecommand{\BIBentrySTDinterwordspacing}{\spaceskip=0pt\relax}
\providecommand{\BIBentryALTinterwordstretchfactor}{4}
\providecommand{\BIBentryALTinterwordspacing}{\spaceskip=\fontdimen2\font plus
\BIBentryALTinterwordstretchfactor\fontdimen3\font minus
  \fontdimen4\font\relax}
\providecommand{\BIBforeignlanguage}[2]{{%
\expandafter\ifx\csname l@#1\endcsname\relax
\typeout{** WARNING: IEEEtran.bst: No hyphenation pattern has been}%
\typeout{** loaded for the language `#1'. Using the pattern for}%
\typeout{** the default language instead.}%
\else
\language=\csname l@#1\endcsname
\fi
#2}}
\providecommand{\BIBdecl}{\relax}
\BIBdecl

\bibitem{zhang2023automated}
J.~Zhang, X.~Yang, W.~Wang, J.~Guan, L.~Ding, and V.~C. Lee, ``Automated guided
  vehicles and autonomous mobile robots for recognition and tracking in civil
  engineering,'' \emph{Automat. Constr.}, vol. 146, p. 104699, 2023.

\bibitem{eskandari2020robotic}
M.~Eskandari~Torbaghan, B.~Kaddouh, M.~Abdellatif, N.~Metje, J.~Liu,
  R.~Jackson, C.~D. Rogers, D.~N. Chapman, R.~Fuentes, M.~Miodownik
  \emph{et~al.}, ``Robotic and autonomous systems for road asset management: a
  position paper,'' \emph{Proc. Inst. Civ. Eng., Smart Infrastruct. Constr.},
  vol. 172, no.~2, pp. 83--93, 2020.

\bibitem{LaTMech2013}
H.~M. La, R.~S. Lim, B.~B. Basily, N.~Gucunski, J.~Yi, A.~Maher, F.~A. Romero,
  and H.~Parvardeh, ``{Mechatronic systems design for an autonomous robotic
  system for high-efficiency bridge deck inspection and evaluation},''
  \emph{{IEEE/ASME} Trans. Mechatronics}, vol.~18, no.~6, pp. 1655--1664, 2013.

\bibitem{LaCASE2013}
H.~La, R.~Lim, B.~Basily, N.~Gucunski, J.~Yi, A.~Maher, F.~Romero, and
  H.~Parvardeh, ``{Autonomous robotic system for high-efficiency
  non-destructive bridge deck inspection and evaluation},'' in \emph{Proc. IEEE
  Conf. Automat. Sci. Eng.}, Madison, WI, 2013, pp. 1065--1070.

\bibitem{GucunskiIJIRA2017}
N.~Gucunski, B.~Basily, J.~Kim, J.~Yi, T.~Duong, K.~Dinh, S.-H. Kee, and
  A.~Maher, ``{RABIT: implementation, performance validation and integration
  with other robotic platforms for improved management of bridge decks},''
  \emph{Int. J. Intell. Robot. Appl.}, vol.~1, no.~3, pp. 271--286, 2017.

\bibitem{Bormann2018}
R.~Bormann, F.~Jordan, J.~Hampp, and M.~H{\"a}gele, ``Indoor coverage path
  planning: Survey, implementation, analysis,'' in \emph{Proc. IEEE Int. Conf.
  Robot. Autom.}, Brisbane, Australia, 2018, pp. 1718--1725.

\bibitem{kan2020online}
X.~Kan, H.~Teng, and K.~Karydis, ``Online exploration and coverage planning in
  unknown obstacle-cluttered environments,'' \emph{{IEEE} Robot. Autom. Lett.},
  vol.~5, no.~4, pp. 5969--5976, 2020.

\bibitem{wu2019energy}
C.~Wu, C.~Dai, X.~Gong, Y.-J. Liu, J.~Wang, X.~D. Gu, and C.~C. Wang,
  ``Energy-efficient coverage path planning for general terrain surfaces,''
  \emph{{IEEE} Robot. Autom. Lett.}, vol.~4, no.~3, pp. 2584--2591, 2019.

\bibitem{Palacios2017}
A.~T. Palacios, A.~S{\'a}nchez~L, and J.~M.~E. Bedolla~Cordero, ``The random
  exploration graph for optimal exploration of unknown environments,''
  \emph{Int. J. Adv. Robot. Syst.}, vol.~14, no.~1, 2017, article
  1729881416687110.

\bibitem{galceran2013survey}
E.~Galceran and M.~Carreras, ``A survey on coverage path planning for
  robotics,'' \emph{Robot. Auton. Syst.}, vol.~61, no.~12, pp. 1258--1276,
  2013.

\bibitem{tan2021comprehensive}
C.~S. Tan, R.~Mohd-Mokhtar, and M.~R. Arshad, ``A comprehensive review of
  coverage path planning in robotics using classical and heuristic
  algorithms,'' \emph{IEEE Access}, vol.~9, pp. 119\,310--119\,342, 2021.

\bibitem{mannadiar2010optimal}
R.~Mannadiar and I.~Rekleitis, ``Optimal coverage of a known arbitrary
  environment,'' in \emph{Proc. IEEE Int. Conf. Robot. Autom.}, Anchorage, AK,
  2010, pp. 5525--5530.

\bibitem{rekleitis2008efficient}
I.~Rekleitis, A.~P. New, E.~S. Rankin, and H.~Choset, ``Efficient boustrophedon
  multi-robot coverage: an algorithmic approach,'' \emph{Ann. Math. Artif.
  Intell.}, vol.~52, no.~2, pp. 109--142, 2008.

\bibitem{Gupta2018}
J.~Song and S.~Gupta, ``$\varepsilon ^{\star }$: An online coverage path
  planning algorithm,'' \emph{{IEEE} Trans. Robotics}, vol.~34, no.~2, pp.
  526--533, 2018.

\bibitem{song2020care}
J.~Song and S.~Gupta, ``Care: Cooperative autonomy for resilience and
  efficiency of robot teams for complete coverage of unknown environments under
  robot failures,'' \emph{Auton. Robots}, vol.~44, no.~3, pp. 647--671, 2020.

\bibitem{elgibreen2019dynamic}
H.~ElGibreen and K.~Youcef-Toumi, ``Dynamic task allocation in an uncertain
  environment with heterogeneous multi-agents,'' \emph{Auton. Robots}, vol.~43,
  no.~7, pp. 1639--1664, 2019.

\bibitem{Bustacara2005}
E.~Gonzalez, O.~Alvarez, Y.~Diaz, C.~Parra, and C.~Bustacara, ``Bsa: A complete
  coverage algorithm,'' in \emph{Proc. IEEE Int. Conf. Robot. Autom.},
  Barcelona, Spain, 2005, pp. 2040--2044.

\bibitem{Zhang2021ICRA}
J.~Tang, C.~Sun, and X.~Zhang, ``Mstc*:multi-robot coverage path planning under
  physical constrain,'' in \emph{Proc. IEEE Int. Conf. Robot. Autom.}, Xi'an,
  China, 2021, pp. 2518--2524.

\bibitem{acar2002morse}
E.~U. Acar, H.~Choset, A.~A. Rizzi, P.~N. Atkar, and D.~Hull, ``Morse
  decompositions for coverage tasks,'' \emph{Int. J. Robot. Res.}, vol.~21,
  no.~4, pp. 331--344, 2002.

\bibitem{choset1998coverage}
H.~Choset and P.~Pignon, ``Coverage path planning: The boustrophedon cellular
  decomposition,'' in \emph{Proc. Int. Conf. on Field and Service Robotics},
  Canberra, Australia, 1998, pp. 203--209.

\bibitem{acar2002sensor}
E.~U. Acar and H.~Choset, ``Sensor-based coverage of unknown environments:
  Incremental construction of morse decompositions,'' \emph{Int. J. Robot.
  Res.}, vol.~21, no.~4, pp. 345--366, 2002.

\bibitem{acar2006sensor}
E.~U. Acar, H.~Choset, and J.~Y. Lee, ``Sensor-based coverage with extended
  range detectors,'' \emph{{IEEE} Trans. Robotics}, vol.~22, no.~1, pp.
  189--198, 2006.

\bibitem{Acar}
E.~Acar, H.~Choset, Y.~Zhang, and M.~Schervish, ``{Path planning for robotic
  demining: robust sensor-based coverage of unstructured environments and
  probabilistic methods},'' \emph{Int. J. Robot. Res.}, vol.~22, no. 7-8, pp.
  441--466, 2003.

\bibitem{gabriely2002spiral}
Y.~Gabriely and E.~Rimon, ``Spiral-stc: An on-line coverage algorithm of grid
  environments by a mobile robot,'' in \emph{Proc. IEEE Int. Conf. Robot.
  Autom.}, Philadelphia, PA, 2002, pp. 954--960.

\bibitem{choset2001coverage}
H.~Choset, ``Coverage for robotics--a survey of recent results,'' \emph{Ann.
  Math. Artif. Intell.}, vol.~31, no. 1-4, pp. 113--126, 2001.

\bibitem{LaValle2006}
S.~{LaValle}, \emph{{Planning Algorithms}}.\hskip 1em plus 0.5em minus
  0.4em\relax New York, NY: Cambridge University Press, 2006.

\bibitem{ersson2001path}
T.~Ersson and X.~Hu, ``Path planning and navigation of mobile robots in unknown
  environments,'' in \emph{Proc. IEEE/RSJ Int. Conf. Intell. Robot. Syst.},
  Maui, HI, 2001, pp. 858--864.

\bibitem{likhachev2005anytime}
M.~Likhachev, D.~I. Ferguson, G.~J. Gordon, A.~Stentz, and S.~Thrun, ``Anytime
  dynamic \uppercase{A}*: An anytime, replanning algorithm.'' in \emph{Proc.
  Int. Conf. Automated Plan. Sched.}, Monterey, CA, 2005, pp. 262--271.

\bibitem{paull2013sensor}
L.~Paull, S.~Saeedi, M.~Seto, and H.~Li, ``Sensor-driven online coverage
  planning for autonomous underwater vehicles,'' \emph{{IEEE/ASME} Trans.
  Mechatronics}, vol.~18, no.~6, pp. 1827--1838, 2013.

\bibitem{hess2013poisson}
J.~Hess, M.~Beinhofer, D.~Kuhner, P.~Ruchti, and W.~Burgard, ``Poisson-driven
  dirt maps for efficient robot cleaning,'' in \emph{Proc. IEEE Int. Conf.
  Robot. Autom.}, Karlsruhe, Germany, 2013, pp. 2245--2250.

\bibitem{hess2014probabilistic}
J.~Hess, M.~Beinhofer, and W.~Burgard, ``A probabilistic approach to
  high-confidence cleaning guarantees for low-cost cleaning robots,'' in
  \emph{Proc. IEEE Int. Conf. Robot. Autom.}, Hong Kong, China, 2014, pp.
  5600--5605.

\bibitem{rottmann2021probabilistic}
N.~Rottmann, R.~Denz, R.~Bruder, and E.~Rueckert, ``A probabilistic approach
  for complete coverage path planning with low-cost systems,'' in \emph{"Proc.
  2021 Europ. Conf. Mobile Robots}, Bonn, Germany, 2021, pp. 1--8.

\bibitem{liu2008omni}
Y.~Liu, J.~J. Zhu, R.~L. Williams, and J.~Wu, ``Omni-directional mobile robot
  controller based on trajectory linearization,'' \emph{Robot. Auton. Syst.},
  vol.~56, no.~5, pp. 461--479, 2008.

\bibitem{indiveri2009swedish}
G.~Indiveri, ``Swedish wheeled omnidirectional mobile robots: Kinematics
  analysis and control,'' \emph{{IEEE} Trans. Robotics}, vol.~25, no.~1, pp.
  164--171, 2009.

\bibitem{yu2019icra}
K.~{Yu}, C.~{Guo}, and J.~{Yi}, ``Complete and near-optimal path planning for
  simultaneous sensor-based inspection and footprint coverage in robotic crack
  filling,'' in \emph{Proc. IEEE Int. Conf. Robot. Autom.}, Montreal, Canada,
  2019, pp. 8812--8818.

\bibitem{guo2017optimal}
C.~Guo, K.~Yu, Y.~Gong, and J.~Yi, ``Optimal motion planning and control of a
  crack filling robot for civil infrastructure automation,'' in \emph{Proc.
  IEEE Conf. Automat. Sci. Eng.}, Xi'an, China, 2017, pp. 1463--1468.

\bibitem{kwan1962graphic}
M.-K. Kwan, ``Graphic programming using odd or even points,'' \emph{Chinese
  Math}, vol.~1, pp. 273--277, 1962.

\bibitem{christofides1981algorithm}
N.~Christofides, V.~Campos, A.~Corber{\'a}n, and E.~Mota, ``An algorithm for
  the rural postman problem,'' Imperial College London, London, UK, {Report
  IC.O.R.81.5}, 1981.

\bibitem{pearn1995algorithms}
W.~L. Pearn and T.~Wu, ``Algorithms for the rural postman problem,''
  \emph{Comput. Oper. Res.}, vol.~22, no.~8, pp. 819--828, 1995.

\bibitem{edmonds1973matching}
J.~Edmonds and E.~L. Johnson, ``Matching, euler tours and the chinese
  postman,'' \emph{Math. Program.}, vol.~5, no.~1, pp. 88--124, 1973.

\bibitem{fleury}
M.~Fleury, ``Deux problemes de geometrie de situation,'' \emph{J. de
  Mathematiques Element. 2nd ser. (in French)}, vol.~2, pp. 257--261, 1883.

\bibitem{xu2014efficient}
A.~Xu, C.~Viriyasuthee, and I.~Rekleitis, ``Efficient complete coverage of a
  known arbitrary environment with applications to aerial operations,''
  \emph{Auton. Robots}, vol.~36, pp. 365--381, 2014.

\bibitem{Lofberg2004}
J.~L{\"{o}}fberg, ``{YALMIP} : A toolbox for modeling and optimization in
  {MATLAB},'' in \emph{Proc. 2004 IEEE Int. Symp. Comp. Aided Control Syst.
  Design}, Taipei, Taiwan, 2004, pp. 284--289.

\bibitem{shi2016automatic}
Y.~Shi, L.~Cui, Z.~Qi, F.~Meng, and Z.~Chen, ``Automatic road crack detection
  using random structured forests,'' \emph{{IEEE} Trans. Intell. Transport.
  Syst.}, vol.~17, no.~12, pp. 3434--3445, 2016.

\end{thebibliography}
\begin{biography}[{\includegraphics[width=1in,height=1.25in,clip,keepaspectratio]{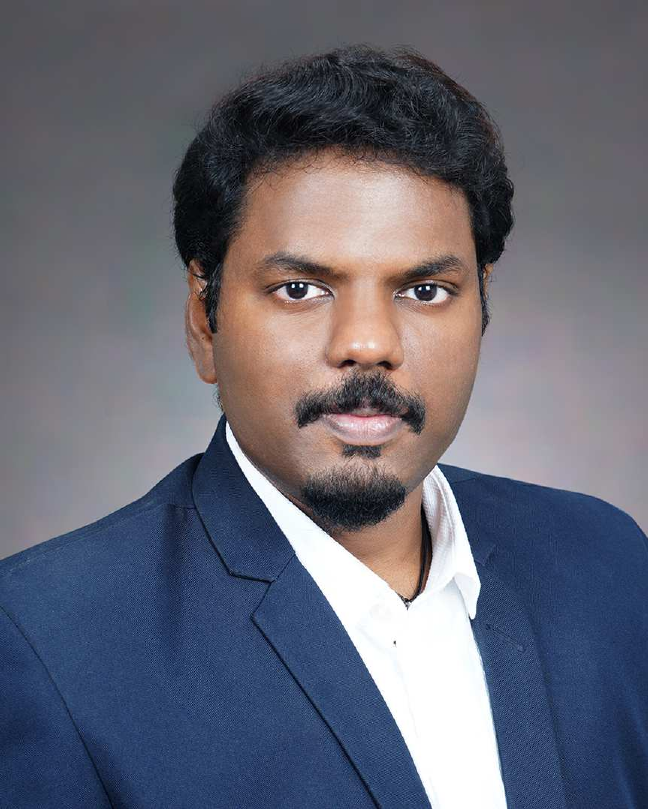}}]{Vishnu Veeraraghavan} (Graduate Student Member, IEEE) received the B.S. degree in Mechanical Engineering from Anna University, Tamil Nadu, India, in 2013, and the M.S. degree in Mechanical Engineering from Binghamton University, Binghamton, NY, USA, in 2016. He is currently working towards a Ph.D. degree in Mechanical Engineering at Binghamton University. His current research focuses on construction robotics, field robots, sensor-based planning, deep learning and artificial intelligence. 
\end{biography}
\vskip 0pt plus -1fil
\vspace{-3\baselineskip}
\begin{biography}[{\includegraphics[width=1in,height=1.25in,clip,keepaspectratio]{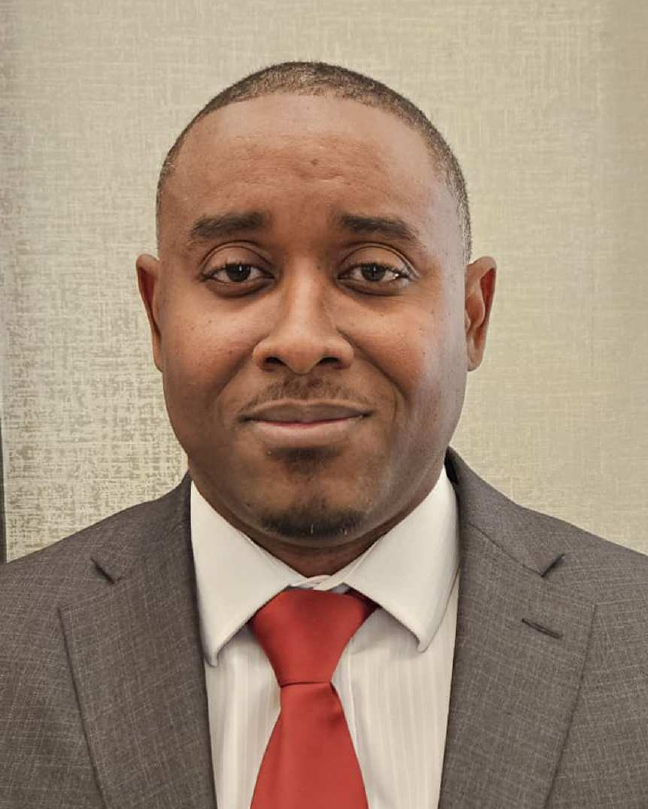}}]{Kyle Hunte} (Member, IEEE) received the B.S. degree in Electrical and Computer Engineering, as well as the M.S. degree in Electrical and Computer Engineering, from the University of the West Indies, St. Augustine, Trinidad and Tobago, in 2010 and 2013, respectively. He also obtained an M.S. degree in Mechanical and Aerospace Engineering from Rutgers University, Piscataway, NJ, USA, in 2023. Currently, he is pursuing a Ph.D. degree in Electrical and Computer Engineering at Rutgers University, Piscataway, NJ, USA. His research interests encompass nonlinear control systems, autonomous robot systems, collaborative robots, deformable object manipulation, motion planning, and AI integration into robotics.
\end{biography}
\vskip 0pt plus -1fil
\vspace{-3\baselineskip}
\begin{biography}[{\includegraphics[width=1in,height=1.25in,clip,keepaspectratio]{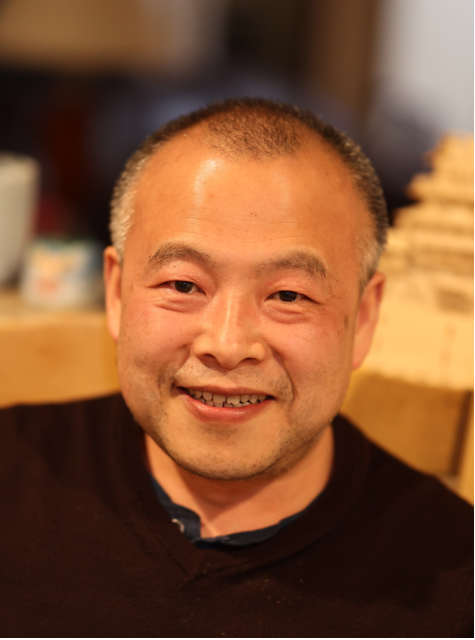}}]{Jingang Yi} (Senior Member, IEEE) received the B.S. degree in electrical engineering from Zhejiang University, Hangzhou, China, in 1993, the M.Eng. degree in precision instruments from Tsinghua University, Beijing, China, in 1996, and the M.A. degree in mathematics and the Ph.D. degree in mechanical engineering from the University of California, Berkeley, CA, USA, in 2001 and 2002, respectively. He is currently a Professor of Mechanical Engineering at Rutgers University. His research interests include autonomous robotic systems, dynamic systems and control, mechatronics, automation science and engineering, with applications to biomedical systems, civil infrastructure and transportation systems.

Dr. Yi is a Fellow of the American Society of Mechanical Engineers (ASME). He was a recipient of the 2010 US NSF CAREER Award. He currently serves as a Senior Editor of the {\em IEEE Transactions on Automation Science and Engineering} and as an Associate Editor for {\em International Journal of Intelligent Robotics and Applications}. He served as a Senior Editor of the {\em IEEE Robotics and Automation Letters} and an Associate Editor for {\em IEEE Transactions on Automation Science and Engineering}, {\em IEEE/ASME Transactions on Mechatronics}, {\em IEEE Robotics and Automation Letters}, {\em IFAC Journal Mechatronics}, {\em Control Engineering Practice}, and {\em ASME Journal of Dynamic Systems, Measurement and Control}.

\end{biography}
\vskip 0pt plus -1fil

\vspace{-3\baselineskip}
\begin{biography}[{\includegraphics[width=1in,height=1.25in,clip,keepaspectratio]{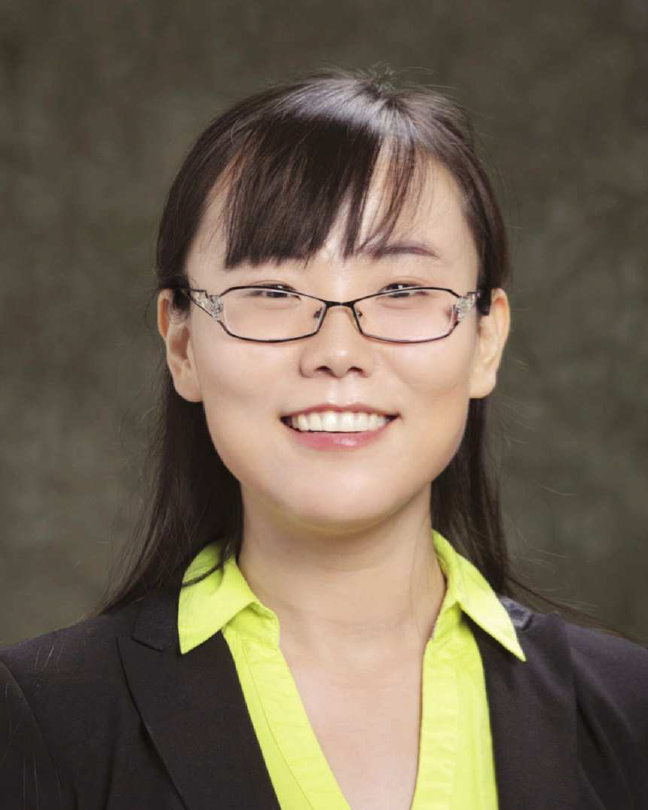}}]{Kaiyan Yu} (Member, IEEE) received the B.S. degree in intelligent science and technology from Nankai University, Tianjin, China, in 2010, and the Ph.D. degree in mechanical and aerospace engineering from Rutgers University, Piscataway, NJ, USA, in 2017. She joined the Department of Mechanical Engineering, Binghamton University, Binghamton, NY, USA, in 2018, where she is currently an Assistant Professor. Her current research interests include autonomous robotic systems, motion planning and control, mechatronics, automation science and engineering with applications to nano/micro particles control and manipulation, Lab-on-a-chip, and biomedical systems. 
	
Dr. Yu is a member of the American Society of Mechanical Engineers (ASME). She is a recipient of the 2022 US NSF CAREER Award. She currently serves as an Associate Vice President of the {\em IEEE Robotics and Automation Society (RAS) Media Services Board} (since 2019) and an Associate Editor of the {\sc IEEE Transactions on Automation Science and Engineering}, {\em IEEE Robotics and Automation Letters}, {\em IFAC Mechatronics}, {\em Frontiers in Robotics and AI}, and the IEEE Robotics and Automation Society Conference Editorial Board and the ASME Dynamic Systems and Control Division Conference Editorial Board  (since 2018).		
\end{biography}

\end{document}